\documentclass[a4paper,fleqn]{cas-sc}

\usepackage[authoryear,longnamesfirst]{natbib}
\usepackage{amsmath}
\usepackage{booktabs}
\usepackage{soul}

\newcommand{\cmark}{\checkmark} 
\newcommand{\xmark}{\text{\sffamily X}}

\def\tsc#1{\csdef{#1}{\textsc{\lowercase{#1}}\xspace}}
\tsc{WGM}
\tsc{QE}
\tsc{EP}
\tsc{PMS}
\tsc{BEC}
\tsc{DE}

\begin{document}
\let\WriteBookmarks\relax


\shorttitle{Detecting AI-Generated Images}

\shortauthors{Mahara and Rishe}

\title [mode=title]{Methods and Trends in Detecting AI-Generated Images: A Comprehensive Review}
\tnotemark[1]

\tnotetext[1]{This work is based in part upon work supported by the National Science Foundation under Grant Nos. MRI20 CNS-2018611 and MRI CNS-1920182.}

\author[1]{Arpan Mahara
} [orcid=0009-0003-5831-3552]
\cormark[1]
\ead{amaha038@fiu.edu}

\author[1]{Naphtali Rishe
} [orcid=0000-0002-1611-4067]
\ead{rishen@cs.fiu.edu}

\affiliation[1]{organization={Knight Foundation School of Computing and Information Sciences, Florida International University},
    city={Miami},
    state={Florida},
    country={USA}}

\cortext[cor1]{Corresponding author}



\begin{abstract}
The proliferation of generative models, such as Generative Adversarial Networks (GANs), Diffusion Models, and Variational Autoencoders (VAEs), has enabled the synthesis of high-quality multimedia data. However, these advancements have also raised significant concerns regarding adversarial attacks, unethical usage, and societal harm. Recognizing these challenges, researchers have increasingly focused on developing methodologies to detect synthesized data effectively, aiming to mitigate potential risks. Prior reviews have predominantly focused on deepfake detection and often overlook recent advancements in synthetic image forensics, particularly approaches that incorporate multimodal frameworks, reasoning-based detection, and training-free methodologies. To bridge this gap, this survey provides a comprehensive and up-to-date review of state-of-the-art techniques for detecting and classifying synthetic images generated by advanced generative AI models. The review systematically examines core detection paradigms, categorizes them into spatial-domain, frequency-domain, fingerprint-based, patch-based, training-free, and multimodal reasoning-based frameworks, and offers concise descriptions of their underlying principles. We further provide detailed comparative analyses of these methods on publicly available datasets to assess their generalizability, robustness, and interpretability. Finally, the survey highlights open challenges and future directions, emphasizing the potential of hybrid frameworks that combine the efficiency of training-free approaches with the semantic reasoning of multimodal models to advance trustworthy and explainable synthetic image forensics.
\end{abstract}


\maketitle

\section{Introduction}
\begin{figure}[!t]
    \centering
    \includegraphics[width=\linewidth]{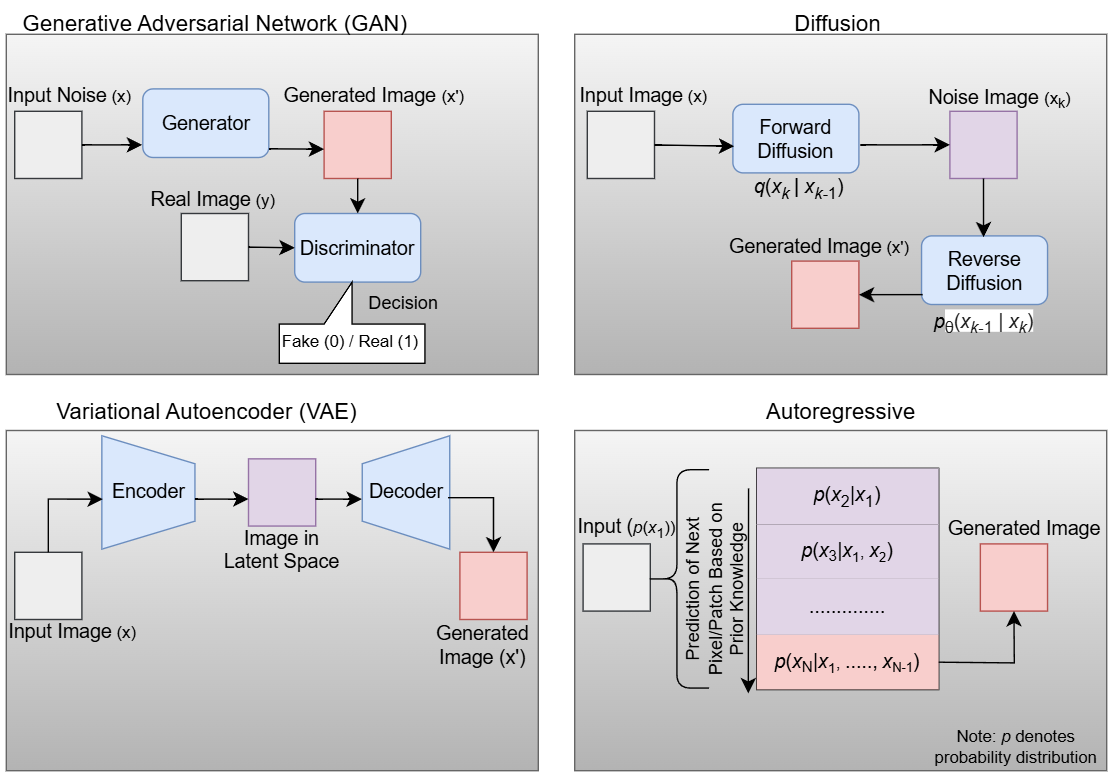}
    \caption{Illustration of Simplified Architecture of GAN, Diffusion, VAE and Autoregressive Generative Models.}
    \label{fig:generative_categories}
\end{figure}

The advent of advanced generative models has enabled the creation of highly realistic synthetic images. These images are synthesized through various approaches, including conditional methods such as image-to-image translation and text-to-image translation, as well as unconditional generation. Generative models based on Generative Adversarial Networks (GANs)~\cite{goodfellow2014generative}, Diffusion Models~\cite{sohl2015deep, ho2020denoising}, Variational Autoencoders (VAEs)~\cite{kingma2013auto}, and Autoregressive Models~\cite{van2008visualizing, van2016pixel} dominate the current literature on image generation. Figure~\ref{fig:generative_categories} provides a simple illustration of the architectures of these model families. While these four generative model families—GANs, Diffusion Models, VAEs, and Autoregressive—have dominated the landscape of image generation, other methods such as Normalizing Flows~\cite{rezende2015variational} have also made significant contributions and deserve attention. With these advancements, several state-of-the-art generative models have been made publicly available alongside commercially accessible tools such as Adobe Firefly~\cite{adobe_firefly}, MidJourney~\cite{midjourney}, DALL·E 3~\cite{dalle3}, and Imagen 3~\cite{imagen3}.

Although generative models have enabled advancements in image synthesis, their accessibility introduces critical concerns related to misinformation, privacy, and security. The ability to generate highly realistic AI-synthesized images raises ethical and security risks, including deceptive media~\cite{sandrini2023generative}, identity fraud~\cite{zhang2025ai}, and geopolitical manipulation~\cite{mylrea2025generative}. Consequently, forensic detection has become an essential area of research, aiming to reliably distinguish AI-generated imagery from real-world photographs. Despite significant progress, existing detection techniques face challenges in generalizability, robustness, scalability, and reasoning across evolving generative architectures. The detection of images generated by both conditional and unconditional generative methods has emerged as a rapidly growing research area. Despite this increasing interest, a comprehensive review that consolidates and systematically categorizes existing detection methodologies is still lacking. While prior surveys have presented reviews on the detection of image manipulations achieved through traditional computer-based techniques~\cite{sencar2013digital}, image editing tools~\cite{zheng2019survey}, and deepfake methodologies~\cite{wang2024deepfake}, our focus diverges from these areas and addresses the limitations in this domain. This survey fills this gap by providing a structured analysis of detection approaches, classifying them based on their underlying techniques and contributions, and focusing on recent advances that incorporate multimodal techniques, reasoning with large language models, and training-free detectors for enhanced forensic accuracy. Given the increasing relevance of such methods in forensic image analysis, it is crucial to explore their long-term applicability and effectiveness in generative image detection.

To establish a unified framework, this survey introduces a core taxonomy that organizes artificial intelligence-generated image (AIGI) detection methodologies into seven categories: spatial-domain analysis, frequency-domain analysis, fingerprint analysis, patch-based analysis, training-free methods, multimodal and reasoning-based models, and commercial detection frameworks. Each category represents a distinct perspective on identifying AI-generated imagery and serves as the foundation for the subsequent sections of this paper. By examining the evolving landscape of adversarial threats in generative AI, this study highlights key challenges, methodological trends, and emerging opportunities, reinforcing the need for continued research in forensic detection of AIGI.

\section{Reviews}
In the present survey of detection methods for AI-generated images, we categorize them into seven distinct groups, with the final category comprising commercial methods. For each group, we discuss the core proposal of each method in ascending order based on publication date. At the end of each subsection, we present a table summarizing whether the experimental evaluations of the methods satisfy three key criteria, described as follows:

1. Cross-Family Generators: This criterion evaluates whether a detection method, trained on images from one type of generative model (e.g., GAN), was tested on and demonstrated effectiveness in detecting images generated by a different type of generative model (e.g., Diffusion models). A method that evaluates images from multiple generative model types satisfies the cross-family criterion.

2. Cross-Category: This criterion examines whether a method was trained and tested on images belonging to different classes. For example, a detector trained on human face images and evaluated on its ability to detect generated images of animals satisfies the cross-category requirement.

3. Cross-Scene: This criterion assesses whether a method’s performance was tested across datasets with distinct data distributions. For instance, a detector trained on images from the ImageNet \cite{russakovsky2015imagenet} dataset and evaluated on images from the LSUN-Bedroom \cite{yu2015lsun} dataset satisfies the cross-scene requirement. Importantly, methods that meet the cross-scene criterion typically also satisfy the cross-category requirement, although the converse is not necessarily true.

It is worth mentioning that these criteria are not extreme cases and do not imply that such evaluations are impossible for any given method. Instead, they are based on the descriptions and experimental setups reported in the original papers. The goal is to assist future work in considering these criteria to demonstrate generalizability and provide real-world evidence.

\begin{figure}[!t]
    \centering
    \includegraphics[width=\linewidth]{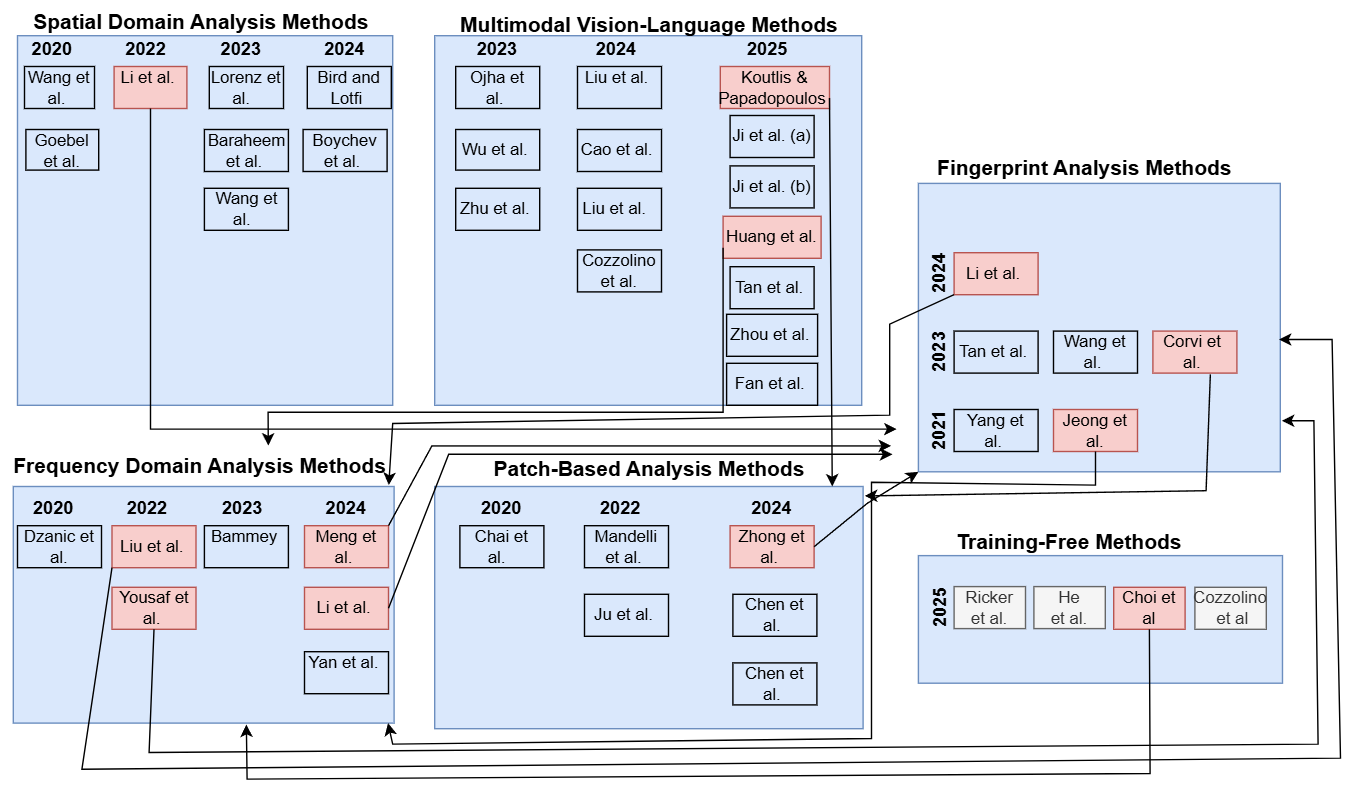}
    \caption{Categorization of Detection Methods Based on Core Architecture and Methodological Proposals. The figure illustrates the division of detection methods into categories. Each category is highlighted using blue-colored rectangles. Some methods are connected to multiple categories, shown using light-red highlights and arrows pointing to the respective sub-categories.}
    \label{fig:detection_categories}
\end{figure}

\subsection{Spatial Domain Analysis / Spatial Feature Analysis Methods}
Spatial domain analysis methods focus on detecting generated images by analyzing intrinsic features within the pixel-level spatial representation. These methods leverage spatial patterns such as texture irregularities, unnatural edge formations, and color inconsistencies, which are common artifacts introduced during the image synthesis process. By examining spatial features like intensity gradients, local pixel dependencies, and edge sharpness, these approaches can effectively uncover subtle distortions. Techniques often employed include convolutional neural networks (CNNs) for automated feature extraction, statistical analysis of pixel intensity distributions, and handcrafted feature-based classifiers. Spatial domain methods excel in capturing localized anomalies and abrupt visual transitions, providing a robust and interpretable approach for image forgery detection.

\subsubsection{CNN-Generated Images Are Surprisingly Easy to Spot: Wang et al. (2020)}
Wang et al. \cite{wang2020cnn} conducted a pivotal study demonstrating that a detection method trained on images generated by a single generative model, particularly a GAN, can generalize to detect synthetic images from a variety of unseen CNN-based generative models. This finding challenges the previously established view that cross-model generalization is inherently difficult for forensic classifiers. To evaluate this, the authors trained ProGAN \cite{karras2017progressive} on the LSUN dataset \cite{yu2015lsun}, producing a dataset of 720K images for training and 4K for validation, with an equal split between real and generated images. They employed a ResNet-50 \cite{he2016deep}, pretrained on ImageNet, as a binary classifier. Robust feature learning was facilitated through various data augmentation strategies, including: (a) no augmentation, (b) Gaussian blur, (c) JPEG compression, and (d) combinations of both with varying probabilities (50\% and 10\%).
The trained classifier demonstrated strong generalization capabilities, successfully detecting images synthesized by other prominent generative models such as StyleGAN \cite{karras2019style}, BigGAN \cite{brock2018big}, CycleGAN \cite{zhu2017unpaired}, StarGAN \cite{choi2018stargan}, and GauGAN \cite{park2019semantic}. This work highlights the existence of common artifacts across CNN-generated images, suggesting that classifiers can leverage these shared patterns for generalizable detection across different architectures and tasks.

\subsubsection{Detection and Attribution of GAN Images: Goebel et al. (2020)}
Goebel et al. \cite{goebel2020detection} introduced a comprehensive framework to detect, attribute, and localize GAN-generated images through co-occurrence matrix analysis and deep learning. This method builds on insights from steganalysis, leveraging pixel-level co-occurrence features to identify artifacts introduced during image synthesis. The approach begins by computing co-occurrence matrices from the RGB channels of the input image in four orientations: horizontal, vertical, diagonal, and anti-diagonal. Each co-occurrence matrix captures a 256×256 histogram of pixel value pairs, normalized and stacked into a tensor of size \(256 \times 256 \times 12\). The matrices are defined as:
\[
C_{i,j} = \sum_{m,n} 
\begin{cases} 
1, & \text{if } I[m,n] = i \text{ and } I[m+1,n] = j \\ 
0, & \text{otherwise}.
\end{cases}
\]

These features are then processed using a modified XceptionNet \cite{chollet2017xception}, designed for three key tasks:
1. Binary Detection: Classifying images as real or GAN-generated.
2. Multi-Class Attribution: Identifying the GAN architecture (e.g., ProGAN, CycleGAN).
3. Localization: Generating heatmaps to identify manipulated regions through patch-based analysis.

Extensive experiments conducted on over 2.76 million images demonstrated the effectiveness of this method across various GAN models, including ProGAN, StyleGAN, CycleGAN, StarGAN, and SPADE/GauGAN. Additionally, t-SNE \cite{van2008visualizing} visualizations showed a clear separation between real and GAN-generated images, reinforcing the model's interpretability and robustness against varying JPEG compression levels and patch sizes. This framework advances GAN forensics by integrating detection, attribution, and localization into a unified pipeline.

\subsubsection{Estimating Artifact Similarity with Representation Learning: Li et al. (2022), GASE-Net}
Li et al. \cite{li2022ArtifactSimilarity} introduced GASE-Net, a framework designed to detect GAN-generated images by estimating artifact similarity. This method addresses challenges in cross-domain generalization and robustness against post-processing, using a two-stage approach: representation learning and representation comparison. In the representation learning stage, a ResNet-50 backbone is modified with instance normalization (IN) applied to the shallow layers to enhance feature extraction by filtering out instance-specific biases. This ensures that the learned representations remain invariant across different domains while retaining category-level distinctions. Feature maps from reference images are aggregated element-wise to form domain prototypes, which serve as robust representations of GAN or pristine image domains.

In the representation comparison stage, the feature map of a consult (suspicious) image is concatenated with the domain prototypes along the channel dimension. A shallow CNN processes the concatenated tensor to output similarity scores. The network is optimized using a Category and Domain-Aware (CDA) loss, which maximizes inter-class separation and minimizes intra-class variation by leveraging both domain and category information. The ground truth similarity scores \( v_{\text{true}} \) for optimization are defined as:
\[
\hat{s}_n =
\begin{cases}
1, & \text{if } y^* = y_n, \\
0, & \text{if } y^* \neq y_n,
\end{cases}
\]
where \( y^* \) and \( y_n \) denote the category labels of the consult image and the \( n \)-th domain prototype, respectively. The predicted similarity scores \( v_{\text{pred}} \) are optimized against \( v_{\text{true}} \) using Mean Square Error (MSE) loss.

During inference, similarity scores are averaged across GAN-generated and pristine domains. An image is classified as GAN-generated if the average fake score exceeds the pristine score. Extensive experiments demonstrated that GASE-Net outperforms state-of-the-art methods in cross-domain scenarios, preserving resilience against various post-processing techniques, including JPEG compression, Gaussian blur, and resizing.

\subsubsection{Local Intrinsic Dimensionality Analysis: Lorenz et al. (2023), AdaptedMultiLID}
Lorenz et al. \cite{lorenz2023detecting} proposed a framework utilizing the Multi-Local Intrinsic Dimensionality (multiLID) method for detecting diffusion-generated images. This approach builds upon the earlier work \cite{lorenz2022unfolding} and demonstrates strong performance in distinguishing both diffusion and GAN-generated images. The method starts by extracting low-dimensional feature maps using an untrained ResNet18 model. MultiLID is then computed to capture local growth rates of feature densities within the latent space. The multiLID feature vector for each sample \( x \) is defined as:
\[
\text{multiLID}_d(x)[i] = -\log\left(\frac{d_i(x)}{d_k(x)}\right),
\]
where \( d_i(x) \) and \( d_k(x) \) represent the Euclidean distances to the \( i^{\text{th}} \) and \( k^{\text{th}} \) nearest neighbors, respectively. A random forest classifier is trained on the labeled multiLID scores to perform image classification. Extensive experiments validate the effectiveness of this method, with high detection accuracy achieved for both diffusion and GAN-generated images across multiple datasets. The framework is also resilient to post-processing operations such as JPEG compression and Gaussian blur, making it suitable for real-world detection scenarios.

\subsubsection{AI-Generated Image Detection: Baraheem et al. (2023)}
Baraheem et al. \cite{baraheem2023ai} proposed a framework to detect GAN-generated images using transfer learning on pretrained classifiers. The authors compiled a diverse dataset called Real and Synthetic Images (RSI), consisting of 48,000 images synthesized by 12 GAN architectures across tasks such as image-to-image, sketch-to-image, and text-to-image generation. EfficientNetB4 \cite{tan2019efficientnet} achieved the best detection performance after fine-tuning on the RSI dataset. The model's architecture was modified by replacing the classifier head with layers for global average pooling, batch normalization, ReLU activation, dropout, and a sigmoid output. The training utilized the Adam optimizer with a batch size of 64, an initial learning rate of 0.001, and data augmentation techniques such as horizontal flips. To facilitate model explainability, the authors incorporated four Class Activation Map (CAM) methods, GradCAM \cite{selvaraju2017gradcam}, AblationCAM \cite{ramaswamy2020ablationcam}, LayerCAM \cite{jiang2021layercam}, and Faster ScoreCAM \cite{wang2020scorecam}, to visualize the discriminative regions influencing classification decisions. 

\subsubsection{Diffusion Reconstruction Error (DIRE): Wang et al. (2023)}
Wang et al. \cite{wang2023dire} proposed DIffusion Reconstruction Error (DIRE), a novel method to detect diffusion-generated images by leveraging reconstruction errors from pre-trained diffusion models. The method addresses the limitations of previous detectors, which struggled to generalize across different diffusion models. The key idea is that diffusion-generated images can be reconstructed more accurately by the pre-trained DDIM model \cite{song2020denoising} than real images. DIRE is defined as the \(L_1\)-norm of the difference between the input image \( x_0 \) and its reconstructed counterpart \( x'_0 \):
\[
\text{DIRE}(x_0) = \|x_0 - x'_0\|_1.
\]

The process involves applying forward noise to the input image and then performing a reverse denoising process to generate the reconstruction. A ResNet-50 classifier is trained using binary cross-entropy loss on these DIRE representations to distinguish between real and generated images. The authors demonstrated that DIRE achieves state-of-the-art performance on their proposed DiffusionForensics dataset, which includes images generated by various diffusion models across multiple domains (e.g., LSUN-Bedroom \cite{yu2015lsun}, ImageNet \cite{russakovsky2015imagenet}, and CelebA-HQ \cite{karras2017progressive}). Extensive experiments showed that DIRE not only excels in detecting diffusion-generated images but also generalizes well to unseen models and maintains robustness under perturbations like Gaussian blur and JPEG compression.

\subsubsection{Classification and Explainable Identification: Bird and Lotfi (2024)}
Bird and Lotfi \cite{bird2024cifake} introduced a framework for detecting AI-generated images using a large-scale dataset, CIFAKE, which is detailed in the datasets section. Their classification approach employs a Convolutional Neural Network (CNN) that processes images through stacked convolutional and pooling layers, followed by fully connected layers with a final sigmoid activation for binary classification. As seen in \cite{baraheem2023ai}, the study emphasizes explainability by implementing Gradient Class Activation Mapping (Grad-CAM) \cite{selvaraju2017gradcam}, which generates heatmaps to highlight regions influencing the model’s decisions. Grad-CAM computes importance weights \(\alpha_k\) for each feature map \(A_k\), producing a visual explanation as:
\[
L_c^{\text{Grad-CAM}} = \text{ReLU} \left( \sum_k \alpha_k A_k \right), \quad \alpha_k = \frac{1}{Z} \sum_{i,j} \frac{\partial y_c}{\partial A_k^{i,j}},
\]
where \(Z\) is the spatial size of the feature map. The heatmaps reveal that the model focuses on subtle imperfections, often in the image background, to differentiate between real and synthetic images. This approach improves trust in AI-generated image detection by combining high classification performance with visual interpretability, making it a valuable contribution to computer vision and data authenticity research.

\subsubsection{Self-Contrastive Learning on ImagiNet: Boychev et al. (2024)}
Boychev et al. \cite{boychev2024imaginet} introduce a large-scale dataset, ImagiNet, consisting of 200,000 real and synthetic images, as detailed in the datasets section. The detection framework consists of two stages: pretraining with Self-Contrastive Learning (SelfCon) and a calibration step. During the pretraining phase, the ResNet-50 \cite{he2016deep} backbone, initialized with ImageNet weights, is paired with a sub-network that projects intermediate feature maps into a shared latent space. This setup produces two output embeddings per input image, facilitating contrastive learning. The SelfCon loss is defined as:
\[
L_{\text{SelfCon}} = \sum_{i \in A, \omega \in \Omega} -\frac{1}{|P(i)||\Omega|} \sum_{p \in P(i), \omega' \in \Omega} \log \frac{\exp(\omega(x_i) \cdot \omega'(x_p) / \tau)}{\sum_{l \in Q(i)} \exp(\omega(x_i) \cdot \omega'(x_l) / \tau)},
\]
where \(A\) represents the set of anchor images in a batch, \(P(i)\) denotes positive samples for anchor \(x_i\), \(Q(i)\) contains negative samples, and \(\omega(x)\) is the normalized embedding. The method balances feature similarities and differences using a temperature parameter \(\tau\).

In the calibration step, the sub-network and projection heads are removed, and a multilayer perceptron (MLP) classifier is trained using cross-entropy loss on an equal number of real and synthetic images. This stage enhances robustness by fine-tuning the learned features for both detection and model identification tasks. Experimental results indicate that the framework achieves up to 0.99 AUC and 95\% balanced accuracy, demonstrating robust zero-shot performance on various synthetic image benchmarks. 

\paragraph{Comparative Analysis of Spatial Domain Analysis Methods}
Spatial-domain forensic methods share the unified goal of identifying AI-generated imagery through pixel-level and localized artifact analysis, yet they differ in their architectural principles and scope of generalization. Early CNN-based detectors, such as Wang et al.~\cite{wang2020cnn}, revealed that discriminative spatial artifacts are shared across different CNN generators, demonstrating cross-model generalization using a standard ResNet-50 backbone trained solely on ProGAN data. However, these early detectors were limited in interpretability and reliance on dataset-specific features. Goebel et al.~\cite{goebel2020detection} addressed these issues by introducing co-occurrence matrix–based representations that capture pixel dependencies across multiple orientations, enabling both detection and attribution within a unified framework. To overcome the poor cross-domain transferability of such handcrafted representations, Li et al.~\cite{li2022ArtifactSimilarity} proposed GASE-Net, emphasizing representation learning and artifact similarity estimation between domain prototypes and consult images, which improved robustness to post-processing and domain shifts. Lorenz et al.~\cite{lorenz2023detecting} further expanded generalization beyond GANs by introducing AdaptedMultiLID, which models intrinsic feature-space dimensionality rather than explicit pixel statistics, addressing the prior limitation of model specificity. Baraheem et al.~\cite{baraheem2023ai} contributed the RSI benchmark and demonstrated that transfer learning on large-scale diverse GAN datasets with EfficientNet and CAM visualizations enhances explainability and model generalization. To address the gap in detecting diffusion-generated content, Wang et al.~\cite{wang2023dire} introduced DIRE, which integrates spatial-domain analysis with diffusion-based reconstruction, achieving higher performance on unseen diffusion architectures. More recently, Bird and Lotfi~\cite{bird2024cifake} and Boychev et al.~\cite{boychev2024imaginet} tackled the remaining limitations of explainability and training efficiency by incorporating Grad-CAM interpretability and self-contrastive learning (SelfCon), respectively, offering improved transparency and zero-shot generalization.  
Collectively, these studies illustrate a clear methodological evolution—from handcrafted co-occurrence features and CNN-based spatial detection, to representation and reconstruction learning, and finally to contrastive and explainable frameworks. The trajectory underscores the field’s movement from model-specific and dataset-dependent detection toward unified, interpretable, and generalizable spatial-domain analysis for AI-generated image forensics.

\begin{table*}
  \caption{Evaluation of Cross-Family Generators, Cross-Category, and Cross-Scene Generalization in Detection Models from Spatial Domain Analysis Category}
  \label{tab:cross_evaluation_spatial}
  \begin{tabular}{lccc}
    \toprule
    \textbf{Models} & \textbf{Cross-Family Generators} & \textbf{Cross-Category} & \textbf{Cross-Scene} \\
    \midrule
    Wang et al. \cite{wang2020cnn} & \xmark & \cmark & \xmark \\
    \midrule
    Goebel et al. \cite{goebel2020detection} & \xmark & \cmark & \xmark \\
    \midrule
    Li et al. \cite{li2022ArtifactSimilarity} & \xmark & \xmark & \xmark \\
    \midrule
    Lorenz et al. \cite{lorenz2023detecting} & \cmark & \cmark & \xmark \\
    \midrule
    Baraheem et al. \cite{baraheem2023ai} & \xmark & \cmark & \xmark \\
    \midrule
    Wang et al. \cite{wang2023dire} & \cmark & \cmark & \cmark \\ 
    \midrule
    Bird \& Lotfi \cite{bird2024cifake} & \xmark & \cmark & \xmark \\
    \midrule
    Boychev et al. \cite{boychev2024imaginet} & \cmark & \cmark & \xmark \\
    \bottomrule
  \end{tabular}
\end{table*}

\begin{figure}[!t]
    \centering
    \includegraphics[width=\linewidth]{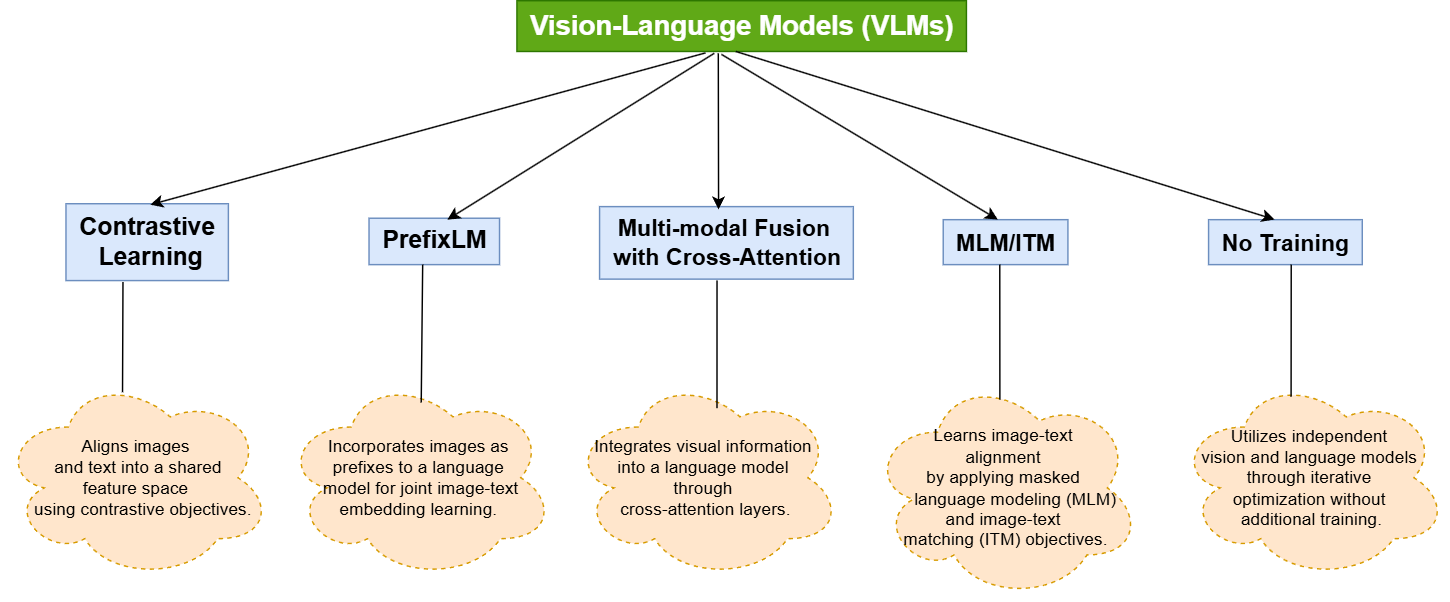}
    \caption{Taxonomy of Vision-Language Models (VLMs) categorized by learning paradigms.}
    \label{fig:vision_language_models}
\end{figure}

\subsection{Multi-modal Vision-Language Methods (and/or Multi-modal Large Language Models)}
This section encompasses two closely related but distinct sub-categories of forensic detection methods: Multimodal Vision-Language Models (VLMs) and Multimodal Large Language Models (MLLMs). The first category leverages vision-language models (VLMs) (see Fig.~\ref{fig:vision_language_models} for representative architectures) trained on large-scale image–text datasets to detect generated images by aligning visual and textual features. These approaches primarily depend on a visual encoder and typically lack explicit reasoning capabilities. They detect inconsistencies in synthetic images through cross-modal embeddings, enabling robust detection across a broad range of generative models. Representative examples include CLIP \cite{radford2021learning}, which belongs to the contrastive learning paradigm \cite{chen2020simplecontrastive} (see Fig.~\ref{fig:vision_language_models} for categorization), and its numerous transfer-learning adaptations that distinguish real images from generated ones. The second category, developed more recently, predominantly employs Multimodal Large Language Models (MLLMs) that jointly utilize visual and textual inputs to provide not only detection but also interpretive reasoning explaining the model’s decision. These methods commonly adopt recent VLM or MLLM backbones such as Qwen \cite{bai2025qwen2}, LLaMA, LLaVA \cite{liu2023visual}, and GPT \cite{achiam2023gpt}, achieving enhanced explainability and alignment with human-understandable reasoning.

\subsubsection{Universal Fake Image Detector by Ojha (2023)}
Ojha et al. \cite{ojha2023towards} identified that existing fake image detectors struggle with generalization, often misclassifying images from unseen generative models as real. This limitation arises from classifiers being asymmetrically tuned to detect artifacts specific to training data. To address this, the authors propose constructing a meaningful feature space using CLIP:ViT \cite{dosovitskiy2020image, radford2021learning}, trained on 400 million image-text pairs. They employ two approaches: (a) mapping training images to CLIP's final layer to create feature representations, then during inference, classifying images based on cosine distance to the nearest neighbor in real and fake feature spaces, and (b) augmenting CLIP with a linear layer for binary classification. Both methods demonstrated generalization, effectively detecting synthetic images from state-of-the-art generative models.

\subsubsection{Language-Guided Synthetic Image Detection by Wu (2023)}
Wu et al. \cite{wu2023generalizable} propose a language-guided approach for detecting synthetic images by integrating image-text contrastive learning in a VLM. They reformulate the detection task as an identification problem, determining whether a query image aligns with an anchor set of text-labeled images. This method enhances generalization by aligning visual features with carefully designed textual labels such as "Real Photo," "Real Painting," "Synthetic Photo," and "Synthetic Painting." The authors found these labels more effective than simpler "real" or "fake" categories, as they account for differences in image types, such as camera-captured versus digitally created content. The proposed LASTED framework encodes images using a ResNet-50x64 vision encoder and text labels using a transformer-based text encoder. During training, both encoders generate visual and textual representations, \(e_v\) and \(e_t\), respectively. A contrastive loss aligns these features, ensuring that matched pairs have higher similarity scores than unmatched ones:
\[
L_I = -\frac{1}{N} \sum_{i=1}^N \log \frac{\exp(e_{v,i} \cdot e_{t,i} / \tau)}{\sum_{j=1}^N \exp(e_{v,i} \cdot e_{t,j} / \tau)},
\]
where \(\tau\) is a temperature parameter.

During testing, the text encoder is discarded. A mean representation vector is computed from an anchor set of known images. The input image's representation is compared to this anchor vector using cosine similarity, which determines its category based on a predefined threshold. This approach allows the model to generalize across diverse generative models and contexts. While LASTED model has achieved better performance on the selected dataset, as seen in the paper text-labels are influential factors, selecting only four labels with photo and paintings might limit towards generalizability when the images are from very different domain such as medical images and the satellite imagery. The researchers should design appropriate labels for their specific tasks related images.

\subsubsection{GenDet: Good Generalizations by Zhu (2023)}
Zhu et al. \cite{zhu2023gendet} address the challenge of detecting synthetic images from unseen generators, which existing methods struggle to classify due to limited generalization. The authors propose \textbf{GenDet}, a detection model composed of two key components: \textit{Teacher-Student Discrepancy-Aware Learning} and \textit{Generalized Feature Augmentation}. These components are trained through an adversarial framework to improve generalization to both seen and unseen generators. The model employs a feature extractor \(E\), based on CLIP:ViT \cite{radford2021learning}, to extract features from input images. The \textit{Teacher-Student Discrepancy-Aware Learning} is designed to: a) Reduce the difference in output between a teacher network (\(N_t\)) and a student network (\(N_s\)) for real images, b) Amplify this difference for fake images to enhance detection.
The discrepancy losses are defined as:
\[
\mathcal{L}_{\text{min}} = \frac{1}{b} \sum_{i=1}^b \|N_t(f_i) - N_s(f_i)\|_2^2, \quad \mathcal{L}_{\text{max}} = -\mathcal{L}_{\text{min}},
\]
where \(f_i\) represents features from real or fake images, and \(b\) is the batch size.

To further enhance generalization, the \textit{Generalized Feature Augmenter} adversarially generates augmented features. This augmenter facilitates the depletion of the difference between the teacher and student networks for augmented fake features, encouraging the model to detect unseen synthetic images by maintaining a large discrepancy during inference. Finally, a binary classifier \(N_c\) is trained on the variation between the teacher and student outputs to classify images as real or fake. Experiments demonstrate that GenDet achieves state-of-the-art performance on the UniversalFakeDetect and GenImage datasets, surpassing prior methods in both accuracy and mean average precision (mAP).

\subsubsection{Mixture of Low-Rank Experts: Liu 2024}
Liu et al. \cite{liu2024mixture} propose a transferable AI-generated image detection model utilizing CLIP-ViT as the backbone with parameter-efficient fine-tuning. The method modifies the MLP layers of the last three ViT-B/32 blocks through a Mixture of Low-Rank Experts (MoLE), integrating both shared and separate Low-Rank Adapters (LoRAs). Shared LoRAs capture common feature representations across datasets, while separate LoRAs specialize in diverse generative patterns. A trainable gating mechanism dynamically assigns input tokens to appropriate experts, with a load-balancing loss ensuring uniform expert utilization. The model freezes most CLIP parameters, adapting only LoRA modules and a new MLP classification head with a sigmoid activation. The forward operation in each MLP block is expressed as:
\[
\Delta Wx = \frac{\alpha}{r} B A x + \sum_{i=1}^{N} G_i(x) \frac{\alpha_i}{r_i} B_i A_i x,
\]
where $G_i(x)$ is the gating function, and $A$, $B$, $A_i$, and $B_i$ are low-rank matrices.

The approach achieves state-of-the-art generalization across unseen diffusion and autoregressive models, with superior robustness to post-processing perturbations like Gaussian blur and JPEG compression. Experimental results on benchmarks such as UnivFD and GenImage show that the proposed method surpasses existing detectors, including UnivFD and DIRE, by up to +12.72\% in classification accuracy.

\subsubsection{MERGING A MIXTURE OF HYPER LORAS: HYPERDET by Cao 2024}
Cao et al. \cite{cao2024hyperdet} propose a generalizable synthetic image detection framework, HyperDet, leveraging the large pretrained multimodal model, CLIP: ViT-L/14, as the backbone, similar to Liu et al. \cite{liu2024mixture}, but with few notable innovations. Unlike Liu et al., the authors introduce a novel grouping of Spatial Rich Model (SRM) filters into five distinct groups to generate multiple filtered views of input images, capturing varying levels of pixel artifacts and features. Along this, HyperDet employs Hyper LoRAs, a hypernetwork-based approach that generates Low-Rank Adaptation (LoRA) weights for fine-tuning the CLIP model. These LoRA weights are computed using three types of embeddings: task embeddings, layer embeddings, and position embeddings. The outputs of these LoRA experts are merged to form a unified representation for classification, effectively integrating shared and specific knowledge for generalizable feature extraction. During training, HyperDet fine-tunes the last eight fully connected layers of the CLIP: ViT-L/14 model, along with the newly introduced Hyper LoRAs modules. To address imbalanced optimization, the framework employs a composite binary cross-entropy loss function, incorporating both original and filtered views of the images. This design obtains robust performance in detecting synthetic images across diverse generative models and datasets.

\subsubsection{Forgery-Aware Adaptive Transformer (FatFormer) by Liu: 2024}
Liu et al. \cite{liu2024fatformer} introduce FatFormer, a generalizable synthetic image detection framework utilizing the pretrained CLIP model inspired by the work of Ojha et al. \cite{ojha2023towards}. The authors address the limitations of freezing CLIP's layers, which hinders the generalization of forgery detection. FatFormer integrates two modules: the Forgery-Aware Adapter (FAA) and Language-Guided Alignment (LGA), for effective adaptation of CLIP's features. The FAA module extracts forgery artifacts from both image and frequency domains. The Image Forgery Extractor applies lightweight convolution layers to capture low-level artifacts, while the Frequency Forgery Extractor employs Discrete Wavelet Transform (DWT) and grouped attention mechanisms to dynamically aggregate multi-band frequency clues. The final adapted feature representation at each ViT stage is defined as:
\[
g^{(j)} = g^{(j)}_{\text{img}} + \lambda \cdot g^{(j)}_{\text{freq}},
\]
where \(\lambda\) balances image and frequency contributions.

The LGA module enhances text prompts using a Patch-Based Enhancer (PBE) and aligns image patch tokens with text embeddings through the Text-Guided Interactor (TGI). Contrastive loss is applied on the cosine similarities between image and text embeddings:
\[
S^{(i)} = \cos(f^{(0)}_{\text{img}}, f^{(i)}_{\text{text}}), \quad S'^{(i)} = \frac{1}{N} \sum_{t=1}^N \cos(f^{(t)}_{\text{img}}, f^{(i)}_{\text{text}}),
\]
where \(N\) is the number of patches.

\subsubsection{Raising the Bar with CLIP: By Cozzolino 2024}
Cozzolino et al. \cite{cozzolino2024raising} implement the CLIP: ViT-L/14 pretrained VLM, similar to the approaches in \cite{cao2024hyperdet, liu2024mixture, koutlis2025leveraging}, for detecting synthetic images with a straightforward yet impactful adjustment. The authors propose generating synthetic images by feeding real-image captions to text-to-image models and then extracting feature vectors for both real and synthetic images using CLIP's image encoder. Specifically, feature vectors are obtained from the second-to-last layer of the ViT module:

\[
r_i = \text{CLIP}(R_i), \quad f_i = \text{CLIP}(F_i),
\]

where \( r_i \) and \( f_i \) represent feature vectors for real image \( R_i \) and synthetic image \( F_i \), respectively.

For classification, the authors employ a simple linear Support Vector Machine (SVM). Notably, the analysis presented suggests that the CLIP-based detector does not rely on the same low-level traces exploited by most existing detectors, making it potentially more robust against adversarial attacks that target low-level features. The authors report using 32,000 images generated by GANs, diffusion models, and commercial text-to-image models for training and evaluation, demonstrating the good performance of their proposed approach across diverse datasets.

\subsubsection{Representation from Encoder-Decoder for Image Detection by Koutlis: 2025}
Koutlis and Papadopoulos \cite{koutlis2025leveraging} propose Representations from Intermediate Encoder-Blocks (RINE) to improve synthetic image detection by extracting low-level features from multiple layers of CLIP’s Vision Transformer (ViT). The method captures both low- and high-level visual semantics by concatenating CLS tokens from each intermediate transformer block into a comprehensive feature representation. The extracted CLS tokens from each block \( Z_l \) are aggregated:
\[
K = \bigoplus_{l=1}^{n} Z_l^{[0]} \in \mathbb{R}^{b \times n \times d},
\]
where \( Z_l^{[0]} \) denotes the CLS token from the \( l \)-th transformer block. To improve feature selection, a Trainable Importance Estimator (TIE) dynamically assigns weights to these representations:
\[
\widetilde{K}_{ik} = \sum_{l=1}^{n} S(A_{lk}) \cdot K_{ilk},
\]
where \( S(A_{lk}) \) represents softmax-activated importance scores for each block.

The features are processed through a projection network, then passed to a classification head with ReLU-activated dense layers and a final sigmoid output for binary classification. The framework optimizes performance using Binary Cross-Entropy (BCE) and Supervised Contrastive Loss:
\[
L = L_{CE} + \xi L_{Cont.},
\]
where \( \xi \) balances the contributions of both objectives. The authors demonstrate that RINE surpasses state-of-the-art methods on 20 test datasets, achieving a +10.6\% accuracy improvement, with training requiring only one epoch (approximately 8 minutes). Additionally, the model is robust to image perturbations, maintaining strong performance across GAN, diffusion, and other synthetic image types.

\subsubsection{Towards Explainable Fake Image Detection with Multi-Modal Large Language Models by Ji et al., 2025}
Ji et al. \cite{ji2025towards} present a study on explainable AI-generated image detection leveraging Multi-Modal Large Language Models (MLLMs) and prompt engineering. The core idea is to enhance interpretability and robustness by designing six specialized prompt paradigms (P1–P6), each interrogating a distinct visual or logical aspect of an image. These include: (P1) \textit{Defect Query} for identifying artifacts such as abnormal lighting or unrealistic textures, (P2) \textit{Regional Analysis} focusing on region-of-interest cues extracted using DINOv2, (P3) \textit{Common Sense Reasoning} to detect physical or spatial inconsistencies, (P4) \textit{Few-Shot Prompting} with exemplar comparisons, (P5) \textit{Structural Analysis} assessing missing or misplaced components, and (P6) \textit{Stereotype Matching} examining exaggerated or overly uniform features. The outputs of these six paradigms are then fused—either sequentially or via majority voting—to yield a final decision supported by coherent reasoning. The authors constructed a diverse dataset of 2,000 images (1,000 real and 1,000 AI-generated) spanning diffusion, GAN-based, and other generative architectures, ensuring cross-model generalization. They benchmarked four major MLLMs—GPT-4o, GPT-4o-mini, Llama-3.2-Vision-Instruct, and LLaVA-CoT—against advanced but non-reasoning detectors such as AEROBLADE \cite{ricker2024aeroblade}, CNNSpot \cite{wang2020cnn}, and so on. The fusion strategy using GPT-4o achieved the highest accuracy (93.4\%), outperforming CNNSpot (91.8\%) and even the best human annotator (86.3\%). The study also revealed that replacing the term “fake” with “generated” in prompts improved model acceptance and reduced rejections, enhancing stability and interpretability.

\subsubsection{Interpretable and Reliable Detection via Grounded Reasoning in MLLMs by Ji et al: 2025}
Ji et al. \cite{ji2025interpretable} propose an interpretable and reliable detection framework that advances AI-generated image forensics through grounded multimodal reasoning. The study introduces the FakeXplained dataset, which contains 8,772 high-quality AI-generated images annotated with bounding boxes and descriptive captions highlighting synthesis artifacts such as logical inconsistencies, unnatural textures, and structural errors. Using this dataset, the authors fine-tune the Qwen2.5-VL-32B \cite{bai2025qwen2} model through a two-stage process involving supervised fine-tuning (SFT) followed by Reinforcement Learning from Human Feedback (RLHF) with progressive Group Relative Policy Optimization (GRPO). This progressive training employs structured rewards for classification accuracy, grounding precision, and output format validity, guiding the model to produce coherent and human-aligned reasoning. The resulting model classifies images as real or synthetic as well as localizes flawed regions and provides textual justifications aligned with visual evidence. 
Furthermore, it demonstrates strong robustness under compression, cropping, and downsampling, as well as generalization to unseen datasets such as FaceForensics++ \cite{rossler2019faceforensics++} and GPT-4o-generated images. The approach establishes a significant step toward explainable and human-trustworthy AI-generated image detection through grounded reasoning in large multimodal models.

\subsubsection{ThinkFake: Reasoning in Multimodal Large Language Models by Huang et al: 2025}
The most important aspect of ThinkFake introduced by Huang et al. \cite{huang2025thinkfake} is the integration of reasoning into AI-generated image detection. The ThinkFake framework introduces a reasoning-based approach to identifying synthetic imagery using Multimodal Large Language Models (MLLMs/VLMs). Built upon the Qwen2.5-VL-7B model \cite{bai2025qwen2}, ThinkFake performs deliberate and interpretable detection through a Forgery Reasoning Prompt that follows a structured chain-of-thought pipeline—quick scan, detailed observation, technical tracing, auxiliary inspection, and final decision—to produce human-readable explanations. The framework incorporates Group Relative Policy Optimization (GRPO) reinforcement learning, where four reward functions—reasoning format, JSON structure, accuracy, and agentic reward—guide structured reasoning and enhance generalization. The agentic reward leverages auxiliary expert agents (UnivFD \cite{ojha2023towards}, AIDE \cite{yan2024sanity}, and so on) to provide semantic, frequency, and dual-stream feedback for more robust detection. Training follows the GenImage protocol, using 6,000 balanced samples of real and AI-generated images (from ImageNet and Stable Diffusion v1.4, respectively). A small supervised fine-tuning (SFT) set of 638 samples is created using Gemini-1.4-pro for reasoning annotations, while a larger reinforcement learning (RL) set of 5,000 samples is used for GRPO-based training. ThinkFake achieves 84.0\% mean accuracy on GenImage and 75.4\% on LOKI, surpassing all baseline detectors while maintaining strong interpretability and generalization capability.

\subsubsection{ForenX: Towards Explainable AI-Generated Image Detection with MLLMs by Tan et al., 2025}
ForenX \cite{tan2025forenx} proposes an explainable AIGI detector that augments an MLLM (LLaVA-8B \cite{liu2023visual} with Llama-3) using a forensic prompt constructed from CLIP-ViT \cite{dosovitskiy2020image, radford2021learning} features and a lightweight forensics projector; the projector creates a forensics embedding supervised by an auxiliary detection loss and mapped into the LLM’s token space, so the LLM consumes three inputs—text tokens, visual content tokens, and the forensic prompt—to produce a yes/no decision with human-readable reasons. The dataset pipeline has two parts: (i) large-scale machine-generated conversations from LLaVA over GenImage and ForenSynths (content Q\&A + detection labels), and (ii) ForgReason, a human-aligned set composed of 2,215 realistic Midjourney fakes with box-level artifact descriptions summarized via GPT-4V \cite{achiam2023gpt}, plus 5,000 real and 1,000 fake samples from GenImage to balance fine-tuning. Training proceeds in two stages with LoRA: Stage-1 jointly tunes CLIP-ViT and the MLLM on GenImage/ForenSynths; Stage-2 freezes CLIP-ViT and further instruction-tunes the MLLM on ForgReason to strengthen explanation quality without degrading recognition. ForenX achieves strong cross-source detection (e.g., 97.8\% mAcc on GenImage; 94.4\% on ForenSynths) and produces grounded, user-study–preferred explanations; it also generalizes to SD-v3 and FLUX images (97.7–97.8\% mAcc). 

\subsubsection{AIGI-Holmes: Towards Explainable and Generalizable AI-Generated Image Detection via MLLMs by Zhou et al., 2025}
AIGI-Holmes \cite{zhou2025aigi} introduces an explainable and generalizable AI-generated image (AIGI) detector that builds on LLaVA-1.6-Mistral-7B with a CLIP-ViT/L-14 visual encoder augmented by an NPR \cite{tan2024rethinking} low-level artifact expert. It constructs the Holmes-Set, a large-scale dataset comprising Holmes-SFTSet (65K images with textual explanations across semantic and artifact-level features) and Holmes-DPOSet (4K contrastive preference pairs refined through expert feedback). The dataset is annotated via a Multi-Expert Jury system combining four MLLMs (Qwen2VL-72B, InternVL2-76B, InternVL2.5-78B, and Pixtral-124B) with structured prompting and human preference refinement. The proposed Holmes Pipeline consists of three training stages: (1) Visual Expert Pre-training using CLIP-ViT and a low-level artifact detector NPR \cite{tan2024rethinking} to establish domain-specific perception; (2) Supervised Fine-Tuning (SFT) to align multimodal representations with textual explanations of real/fake reasoning; and (3) Direct Preference Optimization (DPO) to refine reasoning and align explanations with human judgment. During inference, a Collaborative Decoding strategy fuses predictions from the visual expert and MLLM for robust decision-making. Trained on LLaVA1.6-Mistral-7B \cite, AIGI-Holmes achieves 99.2\% mean accuracy and 99.9 A.P. on unseen diffusion and autoregressive models (e.g., VAR, FLUX), outperforming prior methods such as RINE, AIDE, and NPR. Its explanations also surpass GPT-4o and Pixtral-124B in BLEU (0.622) and ELO (11.42), demonstrating state-of-the-art interpretability and robustness against perturbations (JPEG, blur, downsampling).

\subsubsection{Fake-GPT: Detecting Fake Image via Large Language Model by Fan et al., 2024}
Fake-GPT \cite{fan2024fake} introduces a paradigm shift in fake image detection by employing a pure Large Language Model (LLM) for AI-generated image detection, without any vision encoders, position embeddings, or multimodal alignment. Instead of visual feature extraction, the authors reformulate the problem as a text-sequence prediction task, converting each RGB image into a serialized textual representation of pixel values (e.g., 32×32 RGB flattened as a string) and fine-tuning an LLM to classify it as real or fake. The proposed system uses Qwen-7B-Chat \cite{bai2025qwen2} as the base model and applies Low-Rank Adaptation (LoRA) fine-tuning to adapt the pre-trained LLM for this unconventional visual task. The model receives a prompt such as “You are a trained fake image detector. Given a string of RGB pixel values, determine if this image is fake.”—allowing the LLM’s sequence reasoning ability to distinguish subtle pixel-level patterns indicative of generative artifacts. For training, real samples were drawn from LSUN, ImageNet, CelebA, CelebA-HQ, and COCO datasets, while fake images were generated using ProGAN and evaluated across StyleGAN, StyleGAN2, BigGAN, CycleGAN, StarGAN, GauGAN, and DeepFake generators. All inputs were resized to 32×32 to fit within the token limit, showcasing the model’s robustness under low-resolution settings. Experiments demonstrated competitive accuracy—88.2\% mean across cross-model evaluations and up to 97.4\% on CycleGAN images—outperforming several CNN-based baselines like BiHPF, LGrad, and FrePGAN. When extended to diffusion-generated datasets, Fake-GPT maintained a mean accuracy of 83.8\%, confirming strong cross-model generalization.
In summary, Fake-GPT pioneers the application of pure LLMs for image authenticity detection, reframing fake image detection as a language modeling task. It eliminates vision-specific modules, simplifies training, and demonstrates that large-scale sequence models can serve as universal detectors for AI-generated imagery across both GAN and diffusion domains.
\paragraph{Comparative Analysis of Vision-Language Models}
The earlier generation of vision-language approaches focuses primarily on leveraging pretrained VLMs such as CLIP to improve the generalization of AI-generated image detection through robust visual–textual feature alignment. Ojha et al. \cite{ojha2023towards} initiated this direction by mapping images into CLIP’s semantic space to enable cross-model detection, while Wu et al. \cite{wu2023generalizable} extended it with language-guided textual prompts to refine semantic supervision. Subsequent works, including Zhu et al. \cite{zhu2023gendet}, Liu et al. \cite{liu2024mixture}, and Cao et al. \cite{cao2024hyperdet}, progressively enhanced adaptability through feature-level regularization, low-rank adapters, and hypernetwork-driven parameterization, thereby strengthening cross-generator robustness. FatFormer \cite{liu2024fatformer} advanced this paradigm by incorporating frequency-domain features and language-guided alignment, while Cozzolino et al. \cite{cozzolino2024raising} simplified CLIP-based detection through a linear SVM classifier to isolate higher-level semantics less sensitive to low-level perturbations. Finally, Koutlis and Papadopoulos \cite{koutlis2025leveraging} further optimized representation learning by aggregating intermediate CLIP features to capture both global and local semantics efficiently. Collectively, these studies form a coherent progression—from feature-space distance learning to parameter-efficient fine-tuning and adaptive multimodal integration—marking a shift from handcrafted forensic cues toward transferable, foundation-model-based visual-text alignment for synthetic image detection.

\paragraph{Comparative Analysis of MLLM Models}
All studies converge toward a shared goal of enhancing the interpretability, reliability, and generalizability of AI-generated image detection through language-based reasoning frameworks. The earliest efforts (from 2024 to 2025), such as Fake-GPT \cite{fan2024fake} and ForenX \cite{tan2025forenx}, establish two divergent yet complementary directions—Fake-GPT reformulates image forensics as a pure sequence modeling task using only a textual representation of pixels, while ForenX introduces the first multimodal prompting mechanism that explicitly injects forensic embeddings into an MLLM for explainable decisions. Ji et al. \cite{ji2025towards, ji2025interpretable} advance the field with Towards Explainable Detection and Grounded Reasoning, which evolve from handcrafted forensic prompts to structured, grounded reasoning guided by reinforcement learning and visual grounding, deepening interpretability and human alignment. ThinkFake \cite{huang2025thinkfake} continues this trajectory by explicitly modeling a reasoning chain-of-thought through GRPO optimization and multi-agent feedback, bridging MLLM interpretability with robustness. Finally, AIGI-Holmes \cite{bai2025qwen2} represents the culmination of these trends, integrating multi-expert jury annotation, large-scale instruction tuning, and collaborative decoding to achieve both explainability and cross-domain generalization. Collectively, these works demonstrate a clear methodological progression—from textual reformulation (Fake-GPT), to multimodal prompting (ForenX), to reasoning-driven and human-aligned explanation systems (Ji et al., ThinkFake), culminating in a comprehensive, expert-supervised reasoning pipeline (AIGI-Holmes). The principal differences thus lie in their modality design (pure LLM vs. MLLM), training paradigm (prompt engineering vs. SFT/RLHF/DPO), and explanation grounding (textual, visual, or collaborative), marking a coherent evolution toward explainable and reasoning-based forensic detection.

\begin{table*}
  \caption{Evaluation of Detection Models from Pretrained Vision-Language Methods (and/or Multi-modal Large Language Model) Category on Cross-Family Generators, Cross-Category, and Cross-Scene Generalization}
  \label{tab:cross_evaluation_visionlanguage}
  \begin{tabular}{lccc}
    \toprule
    \textbf{Models} & \textbf{Cross-Family Generators} & \textbf{Cross-Category} & \textbf{Cross-Scene} \\
    \midrule
    Ojha et al. \cite{ojha2023towards} & \cmark & \cmark & \xmark \\
    \midrule
    Wu et al. \cite{wu2023generalizable} & \cmark & \cmark & \xmark \\
    \midrule
    Zhu et al. \cite{zhu2023gendet} & \cmark & \cmark & \xmark \\
    \midrule
    Liu et al. \cite{liu2024mixture} & \cmark & \cmark & \xmark \\
    \midrule
    Cao et al. \cite{cao2024hyperdet} & \cmark & \cmark & \xmark \\
    \midrule
    Liu et al. \cite{liu2024fatformer} & \cmark & \cmark & \xmark \\
    \midrule
    Cozzolino et al. \cite{cozzolino2024raising} & \cmark & \cmark & \xmark \\
    \midrule
    Koutlis et al. \cite{koutlis2025leveraging} & \cmark & \cmark & \xmark \\
    \midrule
    Ji et al. \cite{ji2025towards} & \cmark & \cmark & \xmark \\
    \midrule
    Ji et al. \cite{ji2025interpretable} & \cmark & \cmark & \xmark \\
    \midrule
    Huang et al. \cite{huang2025thinkfake} & \cmark & \cmark & \xmark \\
    \midrule
    Tan et al. \cite{tan2025forenx} & \cmark & \cmark & \xmark \\
    \midrule
    Zhou et al. \cite{zhou2025aigi} & \cmark & \cmark & \cmark \\
    \midrule
    Fan et al. \cite{fan2024fake} & \cmark & \cmark & \xmark \\
    \bottomrule
  \end{tabular}
\end{table*}

\begin{table*}[ht]
\centering
\caption{Evaluation of multimodal vision-language methods on GAN and diffusion-generated images using the UnivFD dataset \cite{ojha2023towards}. Results are reported in classification accuracy (\%). Methods from \cite{cozzolino2024raising} and \cite{koutlis2025leveraging} were trained on only four classes of ProGAN-generated images, while others were trained on the full ProGAN dataset from \cite{wang2020cnn}.}
\label{tab:detection-results-visionlangugage}
\resizebox{\textwidth}{!}{
\begin{tabular}{lcccccccccccccccccccc}
\hline
Method           & \multicolumn{7}{c}{Generative adversarial networks} & \multicolumn{2}{c}{Low-level vision} & \multicolumn{2}{c}{Perceptual loss} & \multicolumn{8}{c}{Diffusion models} & Total \\ \cline{2-7} \cline{9-10} \cline{11-12}   \cline{13-20}
                           & Pro-    & Cycle-  & Big-    & Style-  & Gau-    & Star-   & Deep-   & SITD   & SAN    & CRN    & IMLE   & Guided  & \multicolumn{3}{c}{LDM}                   & \multicolumn{3}{c}{GLIDE}  & DALL-E \\  
                           & GAN     & GAN     & GAN     & GAN     & GAN     & GAN     & fakes   &        &        &        &        &         & 200    & 200s    & 100    & 100   &  50 & 100       \\  
                           &         &         &         &         &         &         &         &        &        &        &        &         & steps  & w/CFG  &    steps & 27        & 27        & 10 \\  
                           &         &         &         &         &         &         &         &        &        &        &        &         &         &         &         &          &         & \\ \hline

Ojha et al. \cite{ojha2023towards} & 100.0 & 98.50 & 94.50 & 82.00 & 99.50 & 97.00 & 66.60 & 63.00 & 57.50 & 59.5 & 72.00 & 70.03 & 94.19 & 73.76 & 94.36 & 79.07 & 79.85 & 78.14 & 86.78 & 81.38 \\
Zhu et al. \cite{zhu2023gendet} & 99.00 & 99.50 & 99.30 & 99.05 & 99.00 & 96.75 & 88.20 & 63.50 & 67.50 & 93.90 & 98.75 & 98.70 & 98.80 & 98.60 & 98.75 & 98.75 & 98.75 & 98.75 & 98.45 & 94.42 \\
Liu et al. \cite{liu2024mixture} & 100.0 & 99.33 & 99.67 & 99.46 & 99.83 & 97.07 & 77.53 & 81.11 & 65.50 & 82.32 & 96.79 & 90.70 & 98.30 & 95.90 & 98.75 & 92.40 & 93.95 & 93.00 & 94.90 & 92.45 \\
Cao et al. \cite{cao2024hyperdet} & 100.0 & 97.40 & 97.50 & 97.50 & 96.20 & 98.65 & 73.85 & 93.00 & 75.00 & 92.75 & 93.20 & 77.35 & 98.70 & 96.60 & 98.80 & 87.75 & 89.95 & 88.70 & 97.00 & 92.10 \\
Liu et al. \cite{liu2024fatformer} &99.90 & 99.30 & 99.50  & 97.20 &99.40  & 99.80 & 93.20 &\_ & \_ & \_ & \_  &76.10 & 98.60 & 94.90   & 98.70  & 94.40  & 94.70 & 94.20 & 98.80 & \_  \\
Koutlis et al. \cite{koutlis2025leveraging} & 100.0 & 99.30 & 99.60 & 88.90 & 99.80 & 99.50& 80.60 & 90.60 & 68.30 & 89.20 & 90.60 & 76.10 & 98.30 & 88.20 & 98.60 & 88.90 &92.60 &90.70 & 95.00 &  \_   \\

\hline
\end{tabular}
}
\end{table*}

\subsection{Training-Free Methods}
Training-Free Methods refer to detection approaches that do not require any supervised training on synthetic or real datasets. Instead of learning discriminative boundaries through model optimization, these methods rely on intrinsic analytical cues, statistical inconsistencies, or reconstruction behaviors inherent to the image or its representation. They typically exploit properties observable through pretrained encoders, perturbation responses, or entropy measures to distinguish genuine from generated content. By eliminating dependence on dataset-specific training, training-free methods emphasize generalization, interpretability, and robustness to unseen generative models.

\subsubsection{AEROBLADE: Training-Free Detection of Latent Diffusion Images Using Autoencoder Reconstruction Error by Ricker et al., 2024}
AEROBLADE~\cite{ricker2024aeroblade} presents a training-free detection framework for identifying images generated by latent diffusion models (LDMs) through their inherent autoencoder (AE) structure. The core idea is that an LDM’s AE reconstructs generated images with lower error than real ones because synthetic samples lie within the learned latent manifold, while real images fall slightly outside it. Consequently, the reconstruction error between an input image and its AE-reconstructed version serves as a reliable indicator of synthesis. This reconstruction-based approach is comparable to DIRE~\cite{wang2023dire} but differs fundamentally in that AEROBLADE requires no training or classifier tuning. Formally, AEROBLADE computes the LPIPS \cite{zhang2018unreasonable} distance between an image $x$ and its reconstruction $D(E(x))$. To generalize across generators, reconstruction errors are evaluated over multiple AEs, and the minimum reconstruction error is used as the detection criterion. The framework was evaluated on seven LDMs and achieved near-perfect detection performance (mean AP up to 0.999), generalizing effectively even to closed-source generators such as Midjourney~\cite{midjourney}. Furthermore, it enables qualitative forensics by visualizing reconstruction error maps that reveal inpainted or manipulated regions within real images. The approach remains robust to perturbations such as JPEG compression, cropping, blur, and noise.

\subsubsection{RIGID: A Training-Free and Model-Agnostic Framework for Robust AI-Generated Image Detection by He et al., 2024}
RIGID~\cite{he2024rigid} introduces a training-free and model-agnostic method for detecting AI-generated images by exploiting the sensitivity of image representations to small noise perturbations. The central observation is that real images exhibit greater stability under minor perturbations than AI-generated images in the feature space of large vision models. RIGID compares the cosine similarity between an image’s feature embedding and that of its noise-perturbed version using pretrained extractors such as DINOv2 \cite{oquab2023dinov2}. Images showing lower similarity (determined with the threshold)  under perturbation are classified as AI-generated. Unlike training-based detectors DIRE \cite{wang2023dire} or training-free detectors such as AEROBLADE~\cite{ricker2024aeroblade}, RIGID operates during inference, requiring no fine-tuning or prior knowledge of the generative model. Its detection score is derived as:
\[
S(x) = 1\{sim(f(x), f(x + \lambda \delta)) \leq \epsilon\}, \quad \delta \sim N(0, I),
\]
where $f$ is the feature extractor and $\lambda$ controls perturbation intensity. The framework generalizes across backbones, with DINOv2 achieving the best trade-off between global and local representation. Evaluated on diverse datasets including ImageNet, LSUN-Bedroom, and GenImage—covering diffusion, GAN, and VAE models—RIGID achieved mean AUC/AP improvements exceeding 25\% over AEROBLADE and frequently surpassed trained detectors such as DIRE \cite{wang2023dire} and Corvi et al. \cite{corvi2023detection}. 

\subsubsection{HFI: A Unified Framework for Training-Free Detection and Implicit Watermarking of Latent Diffusion Model Generated Images by Choi et al., 2024}
HFI~\cite{choi2024hfi} proposes a training-free approach that improves the robustness of autoencoder-based detection of latent diffusion model (LDM) images. It addresses the key limitation of AEROBLADE~\cite{ricker2024aeroblade}, which relies on reconstruction error but tends to overfit to background information, failing on images with simple or uniform backgrounds. HFI instead measures the aliasing effect—distortions of high-frequency components introduced by the LDM autoencoder by quantifying how much high-frequency information is lost during reconstruction. HFI computes the directional derivative of the reconstruction distance in the direction of high-frequency components filtered by a low-pass kernel, capturing subtle distortions beyond global reconstruction loss. A numerical approximation via first-order Taylor expansion allows efficient evaluation without training or model fine-tuning. When multiple autoencoders are available, HFI adopts an ensemble strategy that selects the minimum response across them, improving generalization to unknown generators. Compared to prior training-free methods, HFI achieves clear architectural advancement: unlike RIGID~\cite{he2024rigid}, which relies on feature-space perturbation stability, and AEROBLADE, which measures pixel-space reconstruction consistency, HFI directly exploits the frequency-space degradation intrinsic to LDM autoencoders. This design allows both detection and model attribution (implicit watermarking) by associating distinct aliasing signatures with specific diffusion models. Empirical evaluation across GenImage, SynthBuster, and DiffusionFace benchmarks shows that HFI consistently surpasses AEROBLADE and RIGID, achieving competitive performance to training-based methods.

\subsubsection{Zero-Shot Detection of AI-Generated Images by Cozzolino et al., 2024}
Zero-Shot Entropy-based Detector (ZED)~\cite{cozzolino2024zero} introduces a training-free and generator-independent method for detecting AI-generated images, similar in spirit to AEROBLADE~\cite{ricker2024aeroblade} and RIGID~\cite{he2024rigid}, but differing in that it requires training only on real images. Instead of relying on reconstruction or perturbation cues, ZED measures the surprise of an image under a probabilistic model of real imagery learned through a lossless image encoder. The key idea is that real images conform to the learned distribution, while synthetic ones deviate from it, resulting in a higher coding cost. Using the Super-Resolution based lossless Compressor (SReC)~\cite{cao2020lossless}, ZED predicts conditional probability distributions of pixels across multiple resolutions and computes the negative log-likelihood (NLL) and entropy maps. The difference between the actual and expected coding costs—termed the coding cost gap—serves as the detection statistic. Synthetic images exhibit larger gaps, especially at higher resolutions, reflecting inconsistencies with the statistical model of natural images. This approach generalizes across both GAN- and diffusion-generated images without dependence on any generator-specific artifacts.  

\paragraph{Comparative Analysis of Training-Free Methods}
Training-free detection methods share the common objective of identifying AI-generated content without requiring supervised learning, instead leveraging analytical properties inherent to pretrained models or statistical characteristics of natural images. Despite this shared foundation, their core mechanisms diverge in various perspectives. AEROBLADE~\cite{ricker2024aeroblade} pioneers reconstruction-based detection by exploiting the autoencoder behavior in latent diffusion models, demonstrating that generated images exhibit lower reconstruction error than real ones. RIGID~\cite{he2024rigid} extends this paradigm by evaluating the stability of image embeddings under random perturbations, introducing feature-space sensitivity as a discriminative signal independent of any generator. HFI~\cite{choi2024hfi} advances the idea further through frequency-domain analysis, identifying aliasing artifacts in high-frequency components of reconstructed images, thus unifying detection and implicit watermarking. ZED~\cite{cozzolino2024zero} reinterprets detection as a coding-cost analysis problem, employing a probabilistic model of real images to measure deviations in entropy and compression cost, offering a fully zero-shot and generator-agnostic solution.  
Collectively, these approaches trace a clear methodological evolution—from pixel and latent reconstruction analysis (AEROBLADE), to perturbation and representation stability (RIGID), to spectral and aliasing cues (HFI), and finally to probabilistic learning (ZED). Importantly, these methods converge toward efficient forensic detection without relying on extensive training.

\begin{table*}[t]
  \caption{Evaluation of Detection Models from Training-Free Methods on Cross-Family Generators, Cross-Category, and Cross-Scene Generalization.}
  \label{tab:cross_evaluation_trainingfree}
  \centering
  \begin{tabular}{lccc}
    \toprule
    \textbf{Models} & \textbf{Cross-Family Generators} & \textbf{Cross-Category} & \textbf{Cross-Scene} \\
    \midrule
    AEROBLADE~\cite{ricker2024aeroblade} & \cmark & \xmark & \xmark \\
    \midrule
    RIGID~\cite{he2024rigid} & \cmark & \cmark & \cmark \\
    \midrule
    HFI~\cite{choi2024hfi} & \cmark & \xmark & \xmark \\
    \midrule
    ZED~\cite{cozzolino2024zero} & \cmark & \xmark & \xmark \\
    \bottomrule
  \end{tabular}
\end{table*}

\begin{table}[h!]
\centering
\caption{Diffusion-Based Evaluation of Training-Free Methods and an MLLM Method (ForenX~\cite{tan2025forenx}) on the GenImage dataset~\cite{zhu2024genimage}. Accuracy (Acc) is reported.}
\resizebox{\textwidth}{!}{%
\begin{tabular}{lccccccccc}
\toprule
Method & \multicolumn{1}{c}{ADM} & \multicolumn{1}{c}{BigGAN} & \multicolumn{1}{c}{GLIDE} & \multicolumn{1}{c}{Midjourney} & \multicolumn{1}{c}{SD1.4} & \multicolumn{1}{c}{SD1.5} & \multicolumn{1}{c}{VQDM} & \multicolumn{1}{c}{wukong} & \multicolumn{1}{c}{Mean} \\
\midrule
RIGID~\cite{he2024rigid} & 0.790 & 0.976 & 0.964 & 0.797 & 0.698 & 0.699 & 0.860 & 0.708 & 0.812 \\
AEROBLADE~\cite{ricker2024aeroblade} & 0.838 & 0.986 & 0.990 & 0.988 & 0.983 & 0.984 & 0.723 & 0.984 & 0.935 \\
HFI~\cite{choi2024hfi} & 0.923 & 0.996 & 0.995 & 0.998 & 0.998 & 0.998 & 0.905 & 0.999 & 0.977 \\
ForenX~\cite{tan2025forenx} & 0.974 & 0.978 & 0.980 & 0.979 & 0.978  & 0.977 & 0.977 & 0.980 & 0.978\\
\bottomrule
\end{tabular}%
}
\vspace{0.35em}
\parbox{\textwidth}{\footnotesize
Notes: All methods are training-free diffusion image detectors evaluated on the GenImage dataset~\cite{zhu2024genimage}, with an additional method from the MLLM category, ForenX~\cite{tan2025forenx}. Unlike training-free approaches, ForenX was trained on SDv1.4-generated images and provides reasoning to explain why an image is classified as generated. The results on training-free methods are reported from the study presented by Choi et al.~\cite{choi2024hfi}.
}
\label{tab:diffusion_trainingfree}
\end{table}

\subsection{Frequency Domain Analysis Methods}
Frequency domain analysis transforms image data into the spectral domain, facilitating the detection of periodic artifacts, noise distributions, and variations in frequency components often associated with synthetic image generation. Techniques such as the Discrete Fourier Transform (DFT), Discrete Wavelet Transform (DWT), and Discrete Cosine Transform (DCT) are commonly used to extract these spectral features. The Fourier Transform, proposed by Fourier \cite{baron2003analytical}, decomposes signals into their constituent frequencies and is effective in identifying global periodic patterns, including checkerboard artifacts. The two-dimensional Discrete Fourier Transform (DFT) of an image \( f(x, y) \) of size \( M \times N \) is given by:

\begin{equation}
F(u, v) = \sum_{x=0}^{M-1} \sum_{y=0}^{N-1} f(x, y) e^{-j 2\pi \left( \frac{ux}{M} + \frac{vy}{N} \right)}
\label{eq:DFT}
\end{equation}

where \( F(u, v) \) represents the frequency domain coefficients, and the inverse DFT (IDFT) reconstructs the image as:

\begin{equation}
f(x, y) = \frac{1}{MN} \sum_{u=0}^{M-1} \sum_{v=0}^{N-1} F(u, v) e^{j 2\pi \left( \frac{ux}{M} + \frac{vy}{N} \right)}
\end{equation}

To improve computational efficiency, the Fast Fourier Transform (FFT), introduced by Cooley and Tukey \cite{cooley1965algorithm}, accelerates the DFT process, enabling the rapid detection of periodic patterns and aliasing artifacts commonly found in synthetic images. The Wavelet Transform, introduced by Haar \cite{haar1911theorie}, provides both spatial and frequency localization, making it suitable for detecting transient and localized artifacts. Unlike the Fourier Transform, which represents signals in terms of sinusoidal waves, the Discrete Wavelet Transform (DWT) uses wavelet basis functions to analyze variations in different frequency bands. The one-level decomposition of an image using DWT is given by:

\begin{equation}
f(x, y) = \sum_{m} \sum_{n} c_{m,n} \phi_{m,n}(x, y) + \sum_{m} \sum_{n} d_{m,n} \psi_{m,n}(x, y)
\label{eq:DWT}
\end{equation}

where \( \phi_{m,n}(x, y) \) represents the approximation coefficients (low-frequency components), and \( \psi_{m,n}(x, y) \) captures the detailed coefficients (high-frequency components) at different scales. The Discrete Cosine Transform (DCT), widely applied in image compression (e.g., JPEG), is particularly effective for analyzing energy distribution in smooth regions and identifying compression-induced anomalies \cite{ahmed1974discrete}. The 2D DCT of an image block \( f(x, y) \) of size \( N \times N \) is computed as:

\begin{equation}
F(u, v) = \alpha(u) \alpha(v) \sum_{x=0}^{N-1} \sum_{y=0}^{N-1} f(x, y) 
\cos \left[ \frac{(2x+1)u\pi}{2N} \right] 
\cos \left[ \frac{(2y+1)v\pi}{2N} \right]
\label{eq:DCT}
\end{equation}
where \( \alpha(u), \alpha(v) \) are normalization factors:

\begin{equation}
\alpha(u) =
\begin{cases}
\frac{1}{\sqrt{N}}, & \text{if } u = 0 \\
\frac{\sqrt{2}}{\sqrt{N}}, & \text{if } u > 0
\end{cases}
\end{equation}

The Inverse DCT (IDCT) reconstructs the image block from its frequency coefficients. Since DCT concentrates most of the signal energy in a few low-frequency components, it is useful for detecting compression artifacts and synthetic image inconsistencies.

\begin{figure}[!t]
    \centering
    \includegraphics[width=\linewidth]{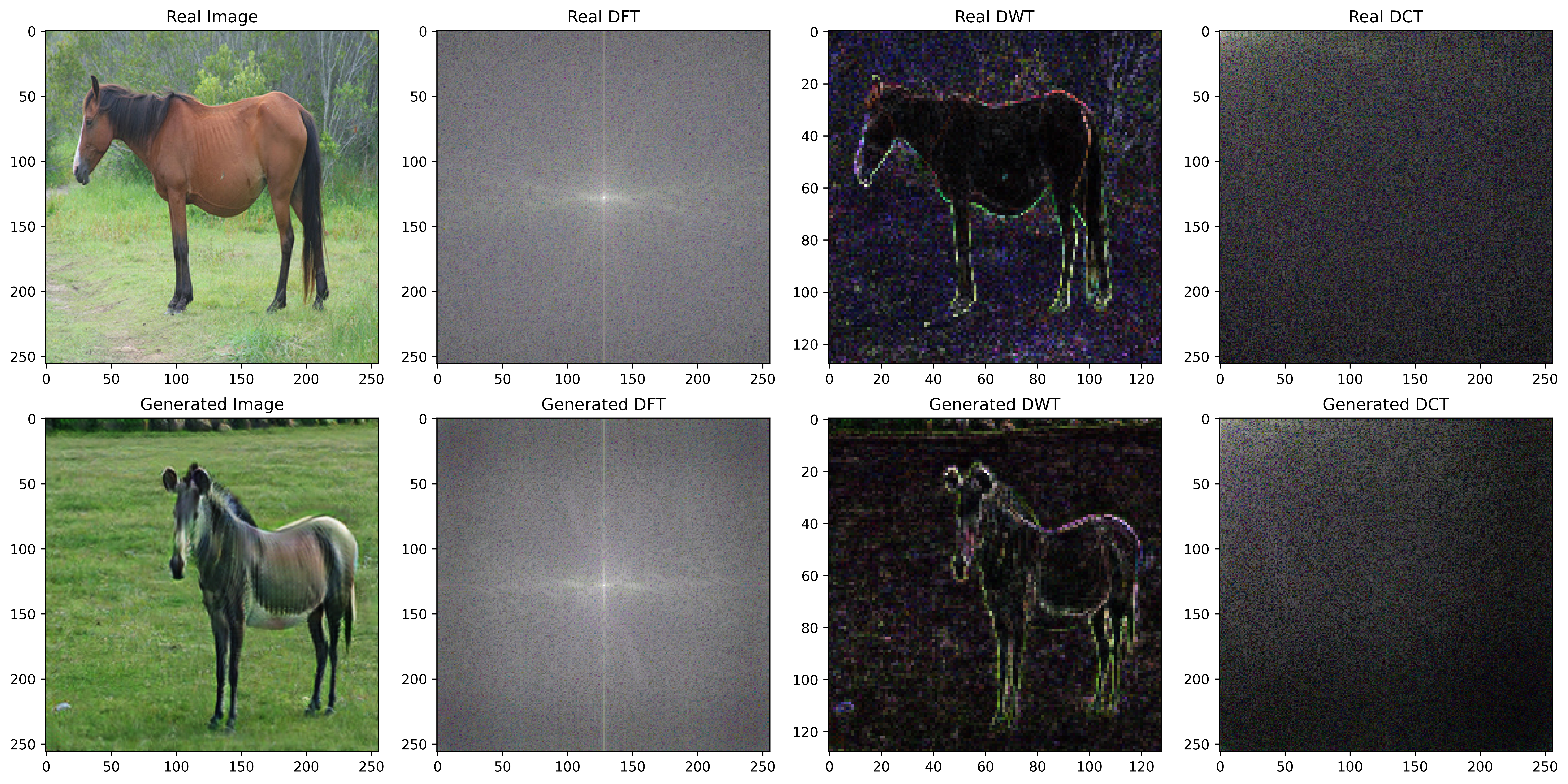}
    \caption{Comparison of frequency domain transformations for real and generated images. The first column presents the input images, followed by their respective DFT, DWT, and DCT representations. The images were obtained from ForenSynth \cite{wang2020cnn}, and the generated image was produced using CycleGAN \cite{zhu2017unpaired}. These images are reproduced under the Attribution-NonCommercial-ShareAlike 4.0 International license.}
    \label{fig:application_frequency_methods}
\end{figure}

\subsubsection{Fourier Spectrum Discrepancies: Dzanic 2020}
Dzanic et al. \cite{dzanic2020fourier} presented a method for analyzing high-frequency Fourier models to highlight the limitations of generative models, such as GANs and VAEs, in reconstructing high-frequency components. Their approach involves applying the Fourier Transform to images to obtain a reduced spectrum, which is then modeled using two decay parameters: \(b_1\), representing the magnitude of high-frequency content, and \(b_2\), representing the decay rate. These parameters were used to train a KNN classifier capable of distinguishing synthetic images from real ones, achieving 99.2\% accuracy on uncompressed high-resolution images. The process involves normalizing the Fourier transform by the DC gain, converting the data to normalized polar coordinates, binning and averaging Fourier coefficients to create a reduced spectrum, and fitting the decay parameters above a threshold wavenumber. This comprehensive method underscores the effectiveness of frequency domain features in identifying the discrepancies characteristic of synthetic image generation.

\subsubsection{Liu's Detection Method Derived from Analysis on Real Images: 2022}
Liu et al. \cite{liu2022detecting} proposed a novel approach to detecting synthetic images by focusing on the inherent noise patterns of real images, deviating from existing methods that analyze artifacts in generated images. They introduced the concept of Learned Noise Patterns (LNP), a high-dimensional spatial mapping derived from neural networks, to characterize the noise properties of real images. By comparing these learned patterns with the noise present in synthetic images, the method identifies discrepancies that indicate image generation. Leveraging both spatial and frequency domain representations, this approach demonstrated improved accuracy in detecting synthetic images across multiple domains.

\subsubsection{Two-Stream Convolutional Network for Fake Content Detection by Yousaf: 2022}
Yousaf et al. \cite{yousaf2022fake} proposed TwoStreamNet, a two-stream convolutional neural network designed to enhance the generalizability of fake visual content detection by jointly analyzing spatial and frequency features. The network comprises two main modules: the Spatial Stream and the Frequency Stream, which independently process input images and fuse their outputs at the classification stage to improve detection accuracy.

The Frequency Stream captures frequency domain artifacts by first converting images to the YCbCr color space to decorrelate color channels. Discrete Fourier Transform (DFT) and Discrete Wavelet Transform (DWT) (see equation~\eqref{eq:DWT}) are applied to each channel. DFT decomposes image signals into real and imaginary components, while DWT captures both low- and high-frequency sub-bands. The resulting frequency features, represented as \(H \times W \times 18\) feature maps, are processed using ResNet-50 \cite{he2016deep} to extract discriminative frequency patterns. The Spatial Stream focuses on spatial domain features by processing original RGB images augmented with Gaussian blur and JPEG compression, similar to the approach in \cite{wang2020cnn}. These augmented images are passed through a ResNet-50 network to extract spatial features. Finally, the outputs of the two streams are fused via probability averaging, ensuring equal contributions from spatial and frequency domains to the final decision. This combined framework highlights the importance of integrating frequency features for robust detection of synthesized visual content.

\subsubsection{Synthbuster by Bammey: 2023}
Bammey \cite{Synth10334046} introduced Synthbuster, a forensic method for detecting synthetic images generated by state-of-the-art diffusion models. The method begins by applying a cross-difference filter, originally defined by Chen et al. \cite{chen2008image}, to highlight periodic frequency artifacts in synthetic images. The filter acts as a high-pass filter, generating residual images using the operation:

\[
C(x, y) = |I(x, y) + I(x+1, y+1) - I(x+1, y) - I(x, y+1)|,
\]

where \( I(x, y) \) represents pixel intensity. 

The Fast Fourier Transform (FFT) is then applied to the residual image, extracting spectral components corresponding to periodicities \(0, 2, 4,\) and \(8\) in both vertical and horizontal directions. The resulting features form a 135-dimensional magnitude peak vector for the RGB channels. These features are classified using a histogram-based gradient boosting tree classifier (HBGB) \cite{ke2017lightgbm}, trained specifically to distinguish synthetic images from real ones.

Synthbuster adapts techniques traditionally used for detecting JPEG compression artifacts to address artifacts from diffusion models, highlighting potential overlaps between the two. The method was validated on JPEG-compressed images to mitigate misclassification risks. Despite its simplicity, Synthbuster outperformed state-of-the-art methods for diffusion image detection. Additionally, the authors have publicly released their dataset, as detailed in the datasets section.

\subsubsection{Meng's Artifact Feature Purification: 2024}
Meng et al. \cite{meng2024artifact} addressed two key limitations in existing detection methods: poor generalization across generative models and limited effectiveness on images from diverse large-scale datasets. To overcome these challenges, they proposed the Artifact Purification Network (APN), which extracts generalizable artifact features through explicit and implicit purification processes.
Explicit purification isolates artifact features in spatial and frequency domains by employing feature decomposition and frequency-band proposals to detect suspicious patterns. Implicit purification, guided by a classifier, further refines these features using mutual information estimation, enhancing the robustness of detection across various generators and datasets. This dual purification approach significantly improves generalization and detection accuracy.

\subsubsection{An Image Transformation Perspective by Li: 2024}
Li et al. \cite{li2024improving} identified biases in existing detection methods, including weakened and overfitted artifact features, which limit their generalization capability. To address these challenges, the authors proposed the SAFE (Simple Preserved and Augmented Features) framework, designed to improve generalizable artifact detection by integrating effective transformations with preprocessing and augmentations. The framework employs several key strategies: (1) replacing conventional down-sampling with a crop operator (RandomCrop for training and CenterCrop for inference) to preserve local correlations and prevent artifact distortion; (2) applying invariant augmentations, such as ColorJitter and RandomRotation, to mitigate color discrepancies and irrelevant rotation-related features; (3) using a patch-based random masking strategy to enhance sensitivity to local image regions and subtle artifacts; and (4) incorporating Discrete Wavelet Transform (DWT) to extract high-frequency features critical for distinguishing synthetic from real images. The processed data was used to train a ResNet model \cite{he2016deep} with 1.44M parameters, achieving strong detection accuracy during testing. This approach demonstrated the effectiveness of integrating diverse transformations to improve detection robustness and generalization.

\subsubsection{Yan's AIDE Method for Synthetic Image Detection (2024)}
Yan et al. \cite{yan2024sanity} introduced the Chameleon dataset, a collection of AI-generated images designed to closely resemble real-world scenes, highlighting the limitations of existing detectors, which frequently misclassify these images as real. To address these challenges, the authors proposed AIDE (AI-generated Image Detector with Hybrid Features), a method that combines low-level frequency-based features with high-level semantic embeddings from CLIP. AIDE processes input images by dividing them into patches and applying discrete cosine transform (DCT) (see equation~\eqref{eq:DCT}) to extract frequency-domain features. Patches are sorted based on computed scores, with the highest- and lowest-frequency patches selected, resized, and processed through SRM \cite{fridrich2012richSRM} to capture noise patterns. These frequency-based features are embedded using a ResNet-50 network, while high-level semantic features are extracted via ConvNeXt-based OpenCLIP. The outputs from both networks are concatenated and passed through a multi-layer perceptron (MLP) for final classification. This hybrid approach demonstrated better performance across benchmarks, including improvements on \textit{AIGCDetectBenchmark} and GenImage \cite{zhu2024genimage}, while achieving competitive results on the challenging Chameleon dataset.

\paragraph{Comparative Analysis of Frequency Domain Analysis Methods}
Frequency-domain detection methods analyze the spectral signatures and periodic inconsistencies introduced by generative models, offering an orthogonal perspective to spatial feature analysis. Early work by Dzanic et al.~\cite{dzanic2020fourier} demonstrated that GANs fail to reproduce high-frequency spectral content accurately, revealing that the Fourier spectrum decay parameters could effectively distinguish real from generated images. However, their approach was limited to GAN-generated data and relied on handcrafted frequency modeling. Liu et al.~\cite{liu2022detecting} shifted focus toward the intrinsic noise characteristics of real images, improving robustness by learning noise priors rather than relying solely on synthetic artifacts, thereby addressing over-dependence on generator-specific patterns. Yousaf et al.~\cite{yousaf2022fake} advanced this direction with TwoStreamNet, which fused spatial and frequency representations through parallel CNN streams, overcoming the modality isolation problem seen in prior methods that treated spatial and frequency cues independently. Bammey’s SynthBuster~\cite{Synth10334046} further extended frequency-based detection to diffusion models, addressing the prior gap in diffusion image forensics. By employing a cross-difference filter and spectral peak analysis, SynthBuster generalized frequency-based techniques beyond GANs and achieved robustness under compression. Building on these insights, Meng et al.~\cite{meng2024artifact} introduced the Artifact Purification Network (APN), which directly tackled two recurring challenges—poor cross-model generalization and dataset dependency—by purifying spatial and frequency artifacts through explicit and implicit feature separation. Li et al.~\cite{li2024improving} improved upon this with the SAFE framework, emphasizing preserved and augmented transformations, including DWT-based feature extraction, to strengthen generalization and mitigate overfitting observed in prior frequency-based CNNs. Yan et al.~\cite{yan2024sanity} then unified these developments by combining frequency and semantic embeddings in AIDE, bridging low-level artifact cues with high-level CLIP features, and achieving superior robustness on challenging datasets like Chameleon, where earlier models failed due to limited semantic understanding. Collectively, these methods demonstrate a clear evolution—from handcrafted spectral analysis (Dzanic) and real-noise modeling (Liu) to dual-domain fusion (Yousaf), diffusion-oriented adaptation (Bammey), feature purification and augmentation (Meng, Li), and finally hybrid semantic–frequency reasoning (Yan). This trajectory reflects a shift from model-specific artifact detection toward unified, semantically informed, and cross-domain generalization in frequency-domain image forensics.

\begin{table*}
  \caption{Evaluation of Detection Models from Frequency Domain Analysis Category on Cross-Family Generators, Cross-Category, and Cross-Scene Generalization}
  \label{tab:cross_evaluation_frequency}
  \begin{tabular}{lccc}
    \toprule
    \textbf{Models} & \textbf{Cross-Family Generators} & \textbf{Cross-Category} & \textbf{Cross-Scene} \\
    \midrule
    Dzanic et al. \cite{dzanic2020fourier} & \xmark & \xmark & \xmark \\
    \midrule
    Liu et al. \cite{liu2022detecting} & \cmark & \cmark & \cmark \\
    \midrule
    Yousaf et al. \cite{yousaf2022fake} & \cmark & \cmark & \xmark \\
    \midrule
    Bammey \cite{Synth10334046} & \cmark & \cmark & \xmark \\
    \midrule
    Meng et al. \cite{meng2024artifact} & \cmark & \cmark & \cmark \\
    \midrule
    Li et al. \cite{li2024improving} & \cmark & \cmark & \cmark \\
    \midrule
    Yan et al. \cite{yan2024sanity} & \cmark & \cmark & \cmark \\
    \bottomrule
  \end{tabular}
\end{table*}

\subsection{Fingerprint Analysis Methods}
Fingerprint analysis in digital forensics refers to techniques that detect unique, traceable features left behind by devices or processes during image generation. Traditional approaches focused on detecting handcrafted features, including device fingerprints and postprocessing fingerprints. Device fingerprints, such as the photo-response nonuniformity (PRNU) pattern \cite{lukas2006digital}, arise from manufacturing imperfections in imaging sensors, leaving a unique and stable mark on each captured image. Postprocessing fingerprints, on the other hand, originate from in-camera processing pipelines, including operations like demosaicking and compression, which embed specific patterns into images \cite{cozzolino2019noiseprint}. While fingerprint analysis often overlaps with frequency domain methods, its primary focus is on extracting distinct fingerprints specific to generative models, such as GANs and diffusion models, rather than general spectral artifacts. The following methods are categorized under fingerprint analysis, as they aim to identify the unique traces of generative techniques.

\subsubsection{Disentangling GAN Fingerprints by Yang et al. (2021)}
Yang et al. \cite{yang2021learning} proposed the GAN Fingerprint Disentangling Network (GFD-Net), advancing previous methods by focusing on disentangling content-irrelevant and GAN-specific fingerprints in synthetic images. Unlike earlier approaches that employ classification setups, GFD-Net integrates a generator, a discriminator, and an auxiliary classifier within an extended GAN framework to explicitly learn and separate these fingerprints. The generator adopts a U-Net architecture with an encoder-decoder structure. A classifier is added to the encoder's output to predict the GAN source, enhancing the generator's feature learning capability. The generator outputs a fingerprint \( f \), which is added to a real image \( x_{\text{real}} \) to produce a fingerprinted synthetic image \( x_{\text{fp}} \):
\[
x_{\text{fp}} = x_{\text{real}} + f.
\]

The discriminator utilizes a PatchGAN architecture to classify images by evaluating smaller patches and averaging their authenticity scores, encouraging the generator to focus on GAN-specific features. Consequently, a ResNet-50 classifier is employed to differentiate fingerprints from different GANs. The training process alternates between:
1. Training the generator (\(G\)) with fixed discriminator (\(D\)) and classifier (\(C\)).
2. Training the discriminator (\(D\)) and classifier (\(C\)) with a fixed generator (\(G\)). The generator is trained using a combination of loss functions:
\[
L_G = \omega_1 L_{z_G} + \omega_2 L_{\text{adv}_G} + \omega_3 L_{\text{cls}_G} + \omega_4 L_{\text{percept}_G},
\]
where \(L_{z_G}\) ensures the generator correctly predicts the image source, \(L_{\text{adv}_G}\) enforces realistic fingerprinted images, \(L_{\text{cls}_G}\) ensures the fingerprint represents the GAN source, and \(L_{\text{percept}_G}\) minimizes perceptual differences between \(x_{\text{fp}}\) and \(x_{\text{real}}\). The discriminator and classifier are trained with two loss functions:
\[
L_{\text{adv}_D} = \mathbb{E}[\log(1 - D(x_{\text{fp}}))] + \mathbb{E}[\log(D(x_{\text{real}}))],
\]
\[
L_{\text{cls}_C} = \text{L}_{\text{CE}}(C(x_{\text{real}}), y) + \text{L}_{\text{CE}}(C(x_{\text{fp}}), y),
\]
where \(L_{\text{adv}_D}\) ensures the discriminator distinguishes real and fingerprinted images, and \(L_{\text{cls}_C}\) trains the classifier for accurate source attribution. By explicitly disentangling GAN-specific fingerprints from image content, GFD-Net demonstrated improved robustness and generalizability compared to earlier methods. Its design builds on prior GAN fingerprinting approaches by incorporating both architectural and training innovations, addressing limitations in source attribution across diverse GANs.

\subsubsection{FingerprintNet: Synthesized Fingerprints for Generalized GAN Detection by Jeong: 2022}
Jeong et al. \cite{jeong2022fingerprintnet} proposed FingerprintNet, a framework aimed at generalizing GAN-generated image detection by synthesizing diverse GAN fingerprints. This approach addresses the challenge of detecting images from unseen GAN architectures without relying on GAN-specific training datasets. FingerprintNet employs an autoencoder-based fingerprint generator, incorporating random layer selection, multi-kernel deconvolution, and feature blending modules to create diverse and robust fingerprints. The generator is trained using a combination of reconstruction loss and similarity loss, ensuring that synthesized fingerprints accurately represent the characteristics of GAN-generated images.

For detection, FingerprintNet applies a Fast Fourier Transform (FFT) to the generated images to extract 2D spectra, where GAN fingerprints are more evident. A ResNet-50 \cite{he2016deep} classifier is then used to distinguish real from generated images based on these fingerprint-highlighted spectra. To address dataset imbalance, the generator creates three fake images for each real image, and a mixed-batch strategy is applied during training to maintain balanced mini-batches. By synthesizing fingerprints and avoiding reliance on specific GAN datasets, FingerprintNet demonstrates improved generalization for detecting images from unseen GAN architectures, advancing robustness in GAN detection tasks.

\subsubsection{Learning on Gradients by Tan: 2023}
Tan et al. \cite{tan2023learning} proposed a novel detection framework, Learning on Gradients (LGrad), which leverages gradient-based representations as generalized artifacts. The framework begins by employing a transformative model, \(T\), a pretrained CNN, to convert real and generated images into feature vectors. The gradient of \( \text{sum}(l) \) with respect to the input image is then computed to capture generalized artifacts:
\[
G = \frac{\partial \text{sum}(l)}{\partial I_i},
\]
where \( l \) represents the feature vector output by \(T\) for an input image \( I_i \).

Following this, a classification model, ResNet-50 \cite{he2016deep}, pretrained on the ImageNet dataset \cite{russakovsky2015imagenet}, is trained on the computed gradients to learn underlying artifacts. Notably, \(T\) remains fixed during gradient computation and is reused during inference to first extract gradients, which are then classified by the trained model. The authors highlight that training the classifier on computed gradients enables generalized learning of artifacts common across GAN models. The paper also compares the performance of various transformative models, including pretrained classifiers and GAN discriminators, showcasing their impact on detection performance. Readers are encouraged to refer to the main paper for a detailed analysis.

\subsubsection{Data Augmentation in Fingerprint Domain By Wang: 2023 (Scaling \& Mixup)}
Wang et al. \cite{wang2023fingerprintAug} proposed a framework to enhance the generalizability of GAN-generated image detectors by augmenting synthetic data in the fingerprint domain. The method involves two key contributions: (1) extraction of fingerprints from synthetic images across different scenes using an encoder trained with Mean Squared Error (MSE) and adversarial losses, and (2) improved cross-GAN generalization through fingerprint perturbation. The framework utilizes an autoencoder trained on real images to extract residual fingerprints from GAN-generated images. These fingerprints, which represent GAN-specific artifacts, are made generalizable across scenes by incorporating a Gradient Reversal Layer (GRL) alongside MSE loss.

To address architecture dependency and simulate fingerprints of unseen GANs, two augmentation strategies were introduced: scaling, which applies a random scaling factor \(\alpha\) to modify fingerprint intensity, and Mixup, which combines fingerprints from multiple samples using weighted sums to generate diverse synthetic fingerprints. After augmentation, perturbed fingerprints are added back to the reconstructed images to create augmented fake images. These augmented samples are used to train a binary classifier with cross-entropy loss, improving detection accuracy across unseen GAN architectures. This approach demonstrates the effectiveness of fingerprint augmentation in enhancing the generalization of GAN detectors, addressing the variability of GAN fingerprints across architectures and scenes.

\subsubsection{Corvi's Analysis on Trending Detection Methods to Detect Generated Images by Diffusion Models: 2023}
Corvi et al. \cite{corvi2023detection} examined the effectiveness of current forensic detection methods in identifying synthetic images generated by diffusion models. Building on the methodology of Marra et al. \cite{marra2019gans}, they employed a denoising filter \cite{zhang2017beyond} to isolate noise residuals by subtracting the scene content from images. Averaging the residuals across multiple images produced a synthetic fingerprint, capturing the artifacts specific to the generation process. Additionally, Corvi et al. conducted spectral analysis by applying the Fourier transform to averaged residuals from 1,000 images. Their findings revealed distinct spectral patterns for models like Stable Diffusion and Latent Diffusion, whereas weaker artifacts were observed in ADM and DALL·E 2, posing challenges for detection. 
The study highlighted the limitations of current detectors, particularly in handling resized and compressed images, and emphasized the need for robust methods tailored to diffusion models. Their findings contribute to understanding the unique challenges of detecting diffusion-generated images, with datasets and code available for further research.

\subsubsection{MaskSim (Li, 2024)}
Li et al. \cite{li2024masksim} proposed MaskSim, a forensic framework for detecting synthetic images generated by diffusion models by focusing on extraction of artifacts. The method leverages traceable artifacts in the Fourier Transformed Spectrum, selectively amplifies these artifacts, and uses a simple linear classifier to achieve competitive detection performance. The framework begins by preprocessing input images with a DnCNN denoiser \cite{zhang2017beyond} to suppress textures and enhance residual synthesis artifacts building on approaches introduced in \cite{marra2019gans} and later adapted in \cite{corvi2023detection} for diffusion-generated images. The residual image undergoes DFT to compute the logarithmic magnitude spectrum, which is then refined using a trainable \(1 \times 1\) convolutional layer and an element-wise masking procedure. A trainable mask is applied to the spectrum, followed by Batch Normalization, and a normalized reference spectrum is computed for comparison.
Detection is performed by computing the cosine similarity between the masked spectrum and the reference spectrum. For synthetic images, cosine similarity values are maximized, while for real images, the absolute cosine values are minimized. A logistic regression classifier, trained using cross-entropy loss, predicts the probability of an image being synthetic. During testing, only regular cosine similarity is used, ensuring robustness and avoiding overfitting. The framework was validated on a dataset of diffusion-generated images \cite{Synth10334046}, achieving strong detection performance and demonstrating its effectiveness in leveraging frequency artifacts for synthetic image detection.

\paragraph{Comparative Analysis of Fingerprint Analysis Methods}
Fingerprint-based methods aim to capture intrinsic traces left by generative models, evolving from GAN-specific fingerprints to diffusion-aware and hybrid artifact extraction techniques. Yang et al.~\cite{yang2021learning} initiated this line of research with GFD-Net, which disentangled content-irrelevant and GAN-specific fingerprints, improving robustness over earlier classifier-based approaches. However, its dependence on model-specific training limited scalability to unseen architectures. Jeong et al.~\cite{jeong2022fingerprintnet} addressed this constraint through FingerprintNet, which synthesized diverse fingerprints via an autoencoder-based generator, enhancing generalization to unseen GANs without retraining. Tan et al.~\cite{tan2023learning} advanced the paradigm with LGrad, which replaced explicit fingerprint synthesis with gradient-based representations, capturing universal generation artifacts across models and reducing reliance on handcrafted fingerprint features. Wang et al.~\cite{wang2023fingerprintAug} further improved generalization by augmenting fingerprints through scaling and Mixup in the latent fingerprint domain, mitigating variability across architectures and scenes. Corvi et al.~\cite{corvi2023detection} shifted focus toward diffusion models, demonstrating that residual-based fingerprints could capture distinctive frequency-domain signatures while highlighting limitations under compression and resizing. Li et al.~\cite{li2024masksim} proposed MaskSim, which unified prior residual-based and spectral fingerprinting techniques using a masked similarity approach in the Fourier domain, achieving robust detection of diffusion-generated imagery without extensive retraining. Collectively, these studies outline a coherent progression—from explicit fingerprint disentanglement and synthesis to generalized and gradient-based learning, and finally to diffusion-oriented spectral fingerprinting. This trajectory reflects the transition from architecture-specific fingerprint extraction to universal, spectrum-informed, and model-agnostic forensic analysis.

\begin{table*}
  \caption{Fingerprint Analysis Methods on Cross-Family Generators, Cross-Category, and Cross-Scene Generalization}
  \label{tab:fingerprint_analysis}
  \begin{tabular}{lccc}
    \toprule
    \textbf{Models} & \textbf{Cross-Family Generators} & \textbf{Cross-Category} & \textbf{Cross-Scene} \\
    \midrule
    Yang et al. \cite{yang2021learning} & \xmark & \cmark & \cmark \\
    \midrule
    Jeong et al. \cite{jeong2022fingerprintnet} & \cmark & \cmark & \xmark \\
    \midrule
    Corvi et al. \cite{corvi2023detection} & \cmark & \xmark & \xmark \\ 
    \midrule
    Tan et al. \cite{tan2023learning} & \xmark & \cmark & \cmark \\ 
    \midrule
    Wang et al. \cite{wang2023fingerprintAug} & \xmark & \cmark & \xmark \\
    \midrule
    Li et al. \cite{li2024masksim} & \xmark & \cmark & \cmark \\
    \bottomrule
  \end{tabular}
\end{table*}

\begin{table}[h!]
\centering
\caption{Evaluation of Fingerprint Analysis Methods on GAN and Diffusion-Generated Images.}
\resizebox{\textwidth}{!}{%
\begin{tabular}{lcccccccccccccccc}
\toprule
\multicolumn{17}{c}{\textbf{GAN-Based Evaluations using Forensynths dataset \cite{wang2020cnn})}. Accuracy (Acc) and Average Precision (AP) metrics are reported.} \\
\multicolumn{17}{c}{\cite{tan2023learning} was trained on class Bedroom from ProGAN-generated image, while remaining were trained on class Horse.} \\
\midrule
Method & \multicolumn{2}{c}{StyleGAN} & \multicolumn{2}{c}{StyleGAN2} & \multicolumn{2}{c}{BigGAN} & \multicolumn{2}{c}{CycleGAN} & \multicolumn{2}{c}{StarGAN} & \multicolumn{2}{c}{GauGAN} & \multicolumn{2}{c}{Mean} \\
\cmidrule(lr){2-3} \cmidrule(lr){4-5} \cmidrule(lr){6-7} \cmidrule(lr){8-9} \cmidrule(lr){10-11} \cmidrule(lr){12-13} \cmidrule(lr){14-15}
 & Acc & AP & Acc & AP & Acc & AP & Acc & AP & Acc & AP & Acc & AP & Acc & AP \\
\midrule
Jeong \cite{jeong2022fingerprintnet} & 74.1 & 85.3 & 89.5 & 96.1 & 85.0 & 94.8 & 71.2 & 96.9 & 99.9 & 100.0 & 75.9 & 90.9 & 82.6 & 94.0 \\
Tan et al. \cite{tan2023learning} & 82.60 & 95.60 & 83.30 & 98.40 & 76.20 & 81.80 & 82.30 & 90.60 & 99.70 & 100.0 & 71.70 & 75.00 & 80.90 & 87.40 \\
Wang et al. (Scaling) \cite{wang2023fingerprintAug} & 85.7 & 98.6 & 83.8 & 98.2 & 81.2 & 85.3 & 83.3 & 93.9 & 99.1 & 100.0 & 75.1 & 81.3 & 84.7 & 92.9 \\
Wang et al. (Mixup) \cite{wang2023fingerprintAug} & 82.2 & 98.7 & 78.0 & 98.1 & 79.1 & 84.8 & 86.4 & 95.6 & 98.8 & 100.0 & 83.4 & 90.3 & 84.7 & 94.6 \\
\midrule
\multicolumn{17}{c}{\textbf{Diffusion-Based Evaluations on the dataset as mentioned in (\cite{li2024masksim})}. Area Under the Curve (AUC) and Accuracy (Acc) metrics are reported.} \\
\midrule
Method & \multicolumn{2}{c}{SD-1} & \multicolumn{2}{c}{SD-2} & \multicolumn{2}{c}{SD-XL} & \multicolumn{2}{c}{DALL·E 2} & \multicolumn{2}{c}{DALL·E 3} & \multicolumn{2}{c}{Midjourney} & \multicolumn{2}{c}{Firefly} & \multicolumn{2}{c}{Mean} \\
\cmidrule(lr){2-3} \cmidrule(lr){4-5} \cmidrule(lr){6-7} \cmidrule(lr){8-9} \cmidrule(lr){10-11} \cmidrule(lr){12-13} \cmidrule(lr){14-15} \cmidrule(lr){16-17}
 & AUC & ACC & AUC & ACC & AUC & ACC & AUC & ACC & AUC & ACC & AUC & ACC & AUC & ACC & AUC & ACC \\
\midrule
Corvi et al. \cite{corvi2023detection} & 100.0 & 99.6 & 99.5 & 97.2 & 98.9 & 80.4 & 48.8 & 49.9 & 54.9 & 49.7 & 99.8 & 95.0 & 86.2 & 52.4 & 84.0 & 74.9 \\
Li et al. \cite{li2024masksim} & 89.4 & 75.5 & 99.1 & 95.9 & 96.6 & 90.0 & 68.2 & 55.4 & 90.2 & 75.3 & 96.4 & 90.9 & 76.0 & 64.0 & 88.3 & 79.4 \\
\bottomrule
\end{tabular}%
}
\label{tab:combined_evaluations_fingerprint}
\end{table}

\subsection{Patch-Based Analysis Methods}
Patch-Based Analysis methods focus on identifying synthetic images by analyzing localized patches rather than the entire image. These methods exploit inconsistencies or artifacts that may appear within smaller regions, enabling the detection of subtle generative patterns. By dividing an image into patches and examining features such as texture, edge coherence, or pixel-level anomalies, these approaches enhance detection granularity. Patch-based strategies are particularly effective in scenarios where global analysis may miss fine-grained artifacts introduced by generative models.

\subsubsection{Patch-Based Classification by Chai et al.: 2020}
Chai et al. \cite{chai2020makes} proposed a patch-based classification framework using truncated ResNet and Xception backbones to classify localized patches of an image as real or fake. By limiting the model's receptive field, the framework focuses on local artifacts rather than global image structure, enhancing generalization across datasets and generative model types. Each patch is processed independently using a 1×1 convolution layer appended to the truncated backbone, and cross-entropy loss is applied. The final output is obtained by averaging the softmax predictions across patches. This increases the data-to-parameter ratio, improving generalization to unseen data. The framework includes a visualization mechanism that generates heatmaps, highlighting regions contributing to the classifier’s predictions. These heatmaps reveal that complex textures, such as hair and facial boundaries, are key to distinguishing real from generated images.

\subsubsection{Orthogonal Training in Detecting GAN-Generated Images: 2022}
Mandelli et al. \cite{mandelli2022detecting} proposed a compact detection framework designed to generalize to unseen GAN architectures. This method employs orthogonal training, where multiple CNNs with EfficientNet-B4 \cite{tan2019efficientnet} backbones are trained on datasets differing in semantic content, GAN models, and post-processing operations. The framework divides input images into 128×128 RGB patches, which each CNN analyzes independently. These patch-level scores are aggregated to classify the entire image. Synthetic images receive the highest patch scores, while real images receive the lowest. The final decision is derived from the average of all CNNs' image-level scores. Experimental results confirmed the method's strong performance in detecting StyleGAN3-generated images without prior exposure during training

\subsubsection{Fusing Global and Local Information by Ju: 2022}
Ju et al. \cite{ju2022fusing} proposed a detection method that combines global and local features to improve generalization for synthetic image detection. The framework uses a ResNet-50 backbone to extract global feature maps from input images. These global features are complemented by a Patch Selection Module (PSM), which identifies and processes the most informative patches to capture subtle, localized artifacts. The PSM selects patches by sliding windows of sizes $3 \times 3$ and $2 \times 2$ over the activation maps, scoring each patch based on its aggregated activation. The top patches (\( k = 6 \)) are mapped back to the original image and reprocessed through ResNet-50 for local feature extraction. Subsequently, an Attention-based Feature Fusion Module (AFFM) combines the global and local features through multi-head attention, generating a unified representation for classification. The authors evaluated their method using a dataset synthesized by 19 different models, including GANs and autoencoder-based DeepFakes. Experimental results showed improved performance over prior methods, particularly under diverse post-processing conditions such as Gaussian blur and JPEG compression.

\subsubsection{Zhong's PatchCraft: 2024}
Zhong et al. \cite{zhong2024patchcraft} introduced PatchCraft, a novel framework emphasizing texture patches over global semantic features for detecting synthetic images. The key innovation is a preprocessing technique, Smash and Reconstruction, which segments an image into patches, ranks them by texture diversity—measured as the sum of pixel differences in horizontal, vertical, and diagonal directions—and reconstructs two images: one enriched with rich-texture patches and another with poor-texture patches. Both reconstructed images are processed through Spatial Rich Model (SRM) filters \cite{fridrich2012richSRM} to extract high-frequency noise patterns, which are further analyzed using a learnable convolutional block. The residual noise patterns between rich and poor texture regions capture inter-pixel correlations, forming a fingerprint for generative models. This fingerprint is leveraged by a convolutional neural network (CNN) classifier, trained with cross-entropy loss, to distinguish between real and synthetic images. During inference, the same pipeline extracts the fingerprint, enabling an accurate classification of input images across diverse generative models.

\subsubsection{Chen's Single Patch Method for AI-Generated Image Detection (2024)}
Chen et al. \cite{chen2024single} proposed the Single Simple Patch (SSP) network, designed to identify whether an image is real or AI-generated by analyzing a single low-texture patch. The selected patch, with dimensions \( M \times M \), is determined by having the lowest texture diversity among all image patches, calculated using the method in \cite{zhong2024patchcraft}. The selected patch is resized to match the original image dimensions and processed using the SRM \cite{fridrich2012richSRM} to extract high-frequency noise patterns. These noise patterns are then fed into a ResNet-50 \cite{he2016deep} classifier, which is trained with binary cross-entropy loss to distinguish real images from generated ones.
To address performance challenges with low-quality images (e.g., those affected by blur or compression artifacts), Chen et al. introduced two additional modules: an enhancement module and a perception module. The perception module, a lightweight three-class classifier, identifies whether a patch is blurry, compressed, or intact. Its predictions guide the enhancement module, which uses a U-Net architecture to improve the patch quality by performing deblurring, decompression, or reconstruction. The improved patch is then processed by the SSP network. 

\subsubsection{Inter-Patch Dependencies for AI-Generated Image Detection by Chen: 2024 (IPD-Net)}
Chen et al. \cite{chen2024ipd} introduced IPD-Net, a detection framework that leverages inter-patch dependencies to improve generalization for AI-generated image detection. This work builds on previous research by Zhong et al. \cite{zhong2024patchcraft}, which demonstrated that inconsistencies in interpixel relations between rich and poor texture regions can serve as key features for detection. IPD-Net extends this by modeling dependencies between patches using a self-attention mechanism. The framework consists of two main modules: Inter-Patch Dependencies Extraction and Inter-Patch Dependencies Classification. In the extraction module, input images are preprocessed with operations such as Gaussian blur and JPEG compression before being passed through SRM filters \cite{fridrich2012richSRM} to extract noise patterns. A non-trained ResNet-50 backbone generates feature maps, and patch dependencies are computed using dot-product similarity across all patches. In the classification module, the dependency matrix undergoes two-dimensional average pooling to reduce dimensionality. A linear classification layer with sigmoid activation then predicts whether an image is real or synthetic. The model is trained using binary cross-entropy loss, and the architecture supports end-to-end training.
Experimental evaluations on the Forensynths \cite{wang2020cnn} and GenImage \cite{zhu2024genimage} datasets demonstrated that IPD-Net outperforms state-of-the-art baseline models in both in-dataset and cross-dataset evaluations, showcasing strong generalization capabilities.

\paragraph{Comparative Analysis of Patch-Based Analysis Methods}
Patch-based detection approaches aim to enhance sensitivity to localized generative artifacts by focusing on smaller image regions rather than global context. Chai et al.~\cite{chai2020makes} demonstrated that limiting the receptive field to local patches enables the network to detect fine-grained generative inconsistencies, improving generalization across GAN architectures. Mandelli et al.~\cite{mandelli2022detecting} extended the concept by employing orthogonal training across semantically diverse datasets and generative sources, enhancing robustness to unseen models and post-processing operations. Ju et al.~\cite{ju2022fusing} further improved representation quality by integrating both global and local cues through a patch selection and feature fusion mechanism, achieving balanced detection performance under varied degradation scenarios. Zhong et al.~\cite{zhong2024patchcraft} emphasized texture-based evidence using PatchCraft, which distinguished between rich- and poor-texture regions to extract residual fingerprints indicative of generative artifacts. Chen et al.~\cite{chen2024single} refined this perspective with the Single Simple Patch (SSP) network, showing that even a single low-texture region can encode sufficient forensic information when enhanced for degradations such as blur or compression. Finally, Chen et al.~\cite{chen2024ipd} proposed IPD-Net, which modeled inter-patch dependencies through self-attention, capturing relational inconsistencies across image regions and achieving strong generalization across datasets. Together, these works trace a clear methodological evolution—from independent local patch analysis to integrated global–local reasoning and inter-patch dependency modeling—demonstrating how localized feature extraction has matured into a robust and generalizable paradigm for synthetic image forensics.

\begin{table*}
  \caption{Evaluation of Patch-Based Analysis Methods on Cross-Family Generators, Cross-Category, and Cross-Scene Generalization}
  \label{tab:patch_based_analysis}
  \begin{tabular}{lccc}
    \toprule
    \textbf{Models} & \textbf{Cross-Family Generators} & \textbf{Cross-Category} & \textbf{Cross-Scene} \\
    \midrule
    Chai et al. \cite{chai2020makes} & \xmark & \xmark & \xmark \\
    \midrule
    Mandelli et al. \cite{mandelli2022detecting} & \xmark & \cmark & \xmark \\
    \midrule
    Ju et al. \cite{ju2022fusing} & \cmark & \cmark & \xmark \\
    \midrule
    Zhong et al. \cite{zhong2024patchcraft} & \cmark & \cmark & \xmark \\
    \midrule
    Chen et al. \cite{chen2024single} & \cmark & \cmark & \xmark \\
    \midrule
    Chen et al. \cite{chen2024ipd} & \cmark & \cmark & \xmark \\
    \bottomrule
  \end{tabular}
\end{table*}

\begin{table}[h!]
\centering
\caption{Evaluation of Patch-Based Analysis Methods on GAN and Diffusion-Generated Images.}
\resizebox{\textwidth}{!}{%
\begin{tabular}{lcccccccccccccccc}
\toprule
\multicolumn{17}{c}{\textbf{GAN-Based Evaluations using Forensynths dataset \cite{wang2020cnn}}. Accuracy (Acc) is reported.} \\
\midrule
Method & \multicolumn{1}{c}{ProGAN} & \multicolumn{1}{c}{StyleGAN} & \multicolumn{1}{c}{StyleGAN2} & \multicolumn{1}{c}{BigGAN} & \multicolumn{1}{c}{CycleGAN} & \multicolumn{1}{c}{StarGAN} & \multicolumn{1}{c}{GauGAN} \\
\midrule
Chai et al. \cite{chai2020makes} & 75.03 & 79.16& \_ & \_ & \_ & \_ & \_ \\
Ju et al. \cite{ju2022fusing} & 100.0 & 85.20 & 83.30 & 77.40 & 87.00 & 97.00 & 77.00 \\
Zhong et al. \cite{zhong2024patchcraft}& 100.0 & 92.77 & 89.55 & 95.8& 70.17 & 99.97 & 71.58 \\
Chen et al. \cite{chen2024single} & 97.05 & 96.05 & \_ & 68.65 & 83.25 & 95.00 & 57.85 \\
 Chen et al. \cite{chen2024ipd} & 99.98 & 95.19 & \_ & 81.02 & 86.57 & 99.08 & 68.67 \\
\midrule
\multicolumn{17}{c}{\textbf{Diffusion-Based Evaluations on the GenImage dataset \cite{zhu2024genimage}}. Accuracy (Acc) metrics are reported.} \\
\multicolumn{17}{c}{Chen et al. \cite{chen2024single} and Chai et al. \cite{chai2020makes} were trained on SD V1.4, while all other methods were trained on ProGAN from the ForenSynths dataset.}\\
\multicolumn{17}{c}{The performance metrics for Chai et al. were recorded from the work of Meng et al. \cite{meng2024artifact}} \\

\midrule
Method & \multicolumn{1}{c}{SDv1.4} & \multicolumn{1}{c}{SD-1.5} & \multicolumn{1}{c}{ADM} & \multicolumn{1}{c}{Glide} & \multicolumn{1}{c}{Midjourney} & \multicolumn{1}{c}{VQDM} & \multicolumn{1}{c}{wukong} & \multicolumn{1}{c}{DALLE2} & \multicolumn{1}{c}{SDXL} \\
\midrule
Chai et al. \cite{chai2020makes}& 99.70 & 99.40 & 51.00& 54.10 & 66.20 & 54.10 & 96.70 & \_ & \_ \\
Ju et al. \cite{ju2022fusing}& 51.00 & 51.40 & 49.00 & 57.20 & 52.20 & 55.10 & 51.70 & 52.80 & 55.60 \\
Zhong et al. \cite{zhong2024patchcraft}& 95.38 & 95.30 & 82.17 & 83.79 & 90.12 & 88.91 & 91.07 & 96.60 & 98.43\\
Chen et al. \cite{chen2024single}  & 99.20 & 99.30 & 78.90 & 88.90 & 82.60 & 96.00 & 98.60 & \_ & \_ \\
Chen et al. \cite{chen2024ipd} & 80.03 & 79.70 & 84.38 & 95.05 & 72.19 & 79.37 & 77.19 & \_ & \_ \\
\bottomrule
\end{tabular}%
}
\label{tab:combined_evaluations_patchbased}
\end{table}

\begin{table*}[ht]
\centering
\caption{Comprehensive Evaluation of methods on GAN and diffusion-generated images using the UnivFD dataset \cite{ojha2023towards}. Results are reported in classification accuracy (\%).}
\label{tab:comprehensive-accuracy-detection-results}
\resizebox{\textwidth}{!}{
\begin{tabular}{lcccccccccccccccccccc}
\hline
Method           & \multicolumn{7}{c}{Generative adversarial networks} & \multicolumn{2}{c}{Low-level vision} & \multicolumn{2}{c}{Perceptual loss} & \multicolumn{8}{c}{Diffusion models} & avg. \\ \cline{2-7} \cline{9-10} \cline{11-12}   \cline{13-20}
                           & Pro-    & Cycle-  & Big-    & Style-  & Gau-    & Star-   & Deep-   & SITD   & SAN    & CRN    & IMLE   & Guided  & \multicolumn{3}{c}{LDM}                   & \multicolumn{3}{c}{GLIDE}  & DALL-E \\  
                           & GAN     & GAN     & GAN     & GAN     & GAN     & GAN     & fakes   &        &        &        &        &         & 200    & 200s    & 100    & 100   &  50 & 100       \\  
                           &         &         &         &         &         &         &         &        &        &        &        &         & steps  & w/CFG  &    steps & 27        & 27        & 10 \\  
                           &         &         &         &         &         &         &         &        &        &        &        &         &         &         &         &          &         & \\ \hline
Wang et al. \cite{wang2020cnn} & 100.0 & 80.49 & 55.77 & 64.14 & 82.23 & 80.97 & 50.66 & 56.11 & 50.00 & 87.73 & 92.85 & 52.30 & 51.20 & 52.20 & 51.40 & 53.45 & 55.35 & 54.30 & 52.60 & 64.41 \\
Chai et al. \cite{chai2020makes} & 68.81 & 53.02 & 55.76 & 59.24 & 52.64 & 77.49 & 55.78 & 59.65 & 48.80 & 65.57 & 61.69 & 52.26 & 58.53 & 60.72 & 58.21 & 55.78 & 56.58 & 55.05 & 61.24 & 58.78 \\
Ojha et al. \cite{ojha2023towards} & 100.0 & 98.50 & 94.50 & 82.00 & 99.50 & 97.00 & 66.60 & 63.00 & 57.50 & 59.5 & 72.00 & 70.03 & 94.19 & 73.76 & 94.36 & 79.07 & 79.85 & 78.14 & 86.78 & 81.38 \\
Wang et al. \cite{wang2023dire} &100.0 & 67.73 & 64.78 & 83.08 & 65.30 & 100.0 & 94.75 & 57.62 & 60.96 & 62.36 & 62.31 & 83.20 & 82.70 & 84.05 & 84.25 & 87.10 & 90.80 & 90.25 & 58.75 & 77.89\\
Tan et al. \cite{tan2023learning}&99.90 & 85.10 & 83.00& 94.80& 72.50 & 99.60& 56.40& 47.80 & 41.10 & 50.60 & 50.70 & 74.20 & 94.20 & 95.90& 95.00& 87.20 & 90.80 & 89.80 & 88.40 & \_\\
Corvi et al. \cite{corvi2023detection}&100.00 & 92.00 & 96.90 & 99.40& 94.80 & 99.50& 54.10& 90.60 & 55.50 & 100.00 & 100.00 & 53.90 & 58.00 & 61.10 & 57.50& 56.90 & 59.60 & 58.80 & 71.70 & \_\\
Zhu et al. \cite{zhu2023gendet} & 99.00 & 99.50 & 99.30 & 99.05 & 99.00 & 96.75 & 88.20 & 63.50 & 67.50 & 93.90 & 98.75 & 98.70 & 98.80 & 98.60 & 98.75 & 98.75 & 98.75 & 98.75 & 98.45 & 94.42 \\
Liu et al. \cite{liu2024mixture} & 100.0 & 99.33 & 99.67 & 99.46 & 99.83 & 97.07 & 77.53 & 81.11 & 65.50 & 82.32 & 96.79 & 90.70 & 98.30 & 95.90 & 98.75 & 92.40 & 93.95 & 93.00 & 94.90 & 92.45 \\
Cao et al. \cite{cao2024hyperdet} & 100.0 & 97.40 & 97.50 & 97.50 & 96.20 & 98.65 & 73.85 & 93.00 & 75.00 & 92.75 & 93.20 & 77.35 & 98.70 & 96.60 & 98.80 & 87.75 & 89.95 & 88.70 & 97.00 & 92.10 \\
Liu et al. \cite{liu2024fatformer} &99.90 & 99.30 & 99.50  & 97.20 &99.40  & 99.80 & 93.20 &\_ & \_ & \_ & \_  &76.10 & 98.60 & 94.90   & 98.70  & 94.40  & 94.70 & 94.20 & 98.80 & \_  \\
Koutlis et al. \cite{koutlis2025leveraging} & 100.0 & 99.30 & 99.60 & 88.90 & 99.80 & 99.50& 80.60 & 90.60 & 68.30 & 89.20 & 90.60 & 76.10 & 98.30 & 88.20 & 98.60 & 88.90 &92.60 &90.70 & 95.00 &  \_   \\
\hline
\end{tabular}
}
\end{table*}

\begin{table*}[ht]
\centering
\caption{Comprehensive Evaluation of methods on GAN and diffusion-generated images using the UnivFD dataset \cite{ojha2023towards}. Results are reported in average precision (\%).}
\label{tab:comprehensive-averageprecision-detection-results}
\resizebox{\textwidth}{!}{
\begin{tabular}{lcccccccccccccccccccc}
\hline
Method           & \multicolumn{7}{c}{Generative adversarial networks} & \multicolumn{2}{c}{Low-level vision} & \multicolumn{2}{c}{Perceptual loss} & \multicolumn{8}{c}{Diffusion models} & mAP \\ \cline{2-7} \cline{9-10} \cline{11-12}   \cline{13-20}
                           & Pro-    & Cycle-  & Big-    & Style-  & Gau-    & Star-   & Deep-   & SITD   & SAN    & CRN    & IMLE   & Guided  & \multicolumn{3}{c}{LDM}                   & \multicolumn{3}{c}{GLIDE}  & DALL-E \\  
                           & GAN     & GAN     & GAN     & GAN     & GAN     & GAN     & fakes   &        &        &        &        &         & 200    & 200s    & 100    & 100   &  50 & 100       \\  
                           &         &         &         &         &         &         &         &        &        &        &        &         & steps  & w/CFG  &    steps & 27        & 27        & 10 \\  
                           &         &         &         &         &         &         &         &        &        &        &        &         &         &         &         &          &         & \\ \hline
Wang et al. \cite{wang2020cnn} &100.0 & 96.36 & 85.34 & 98.10 & 98.48 & 96.97 & 60.33 & 82.95 & 54.22 & 99.61 & 99.81 & 69.93 & 66.17 & 67.68 & 66.13 & 71.18 & 76.37 & 72.13 & 67.66 & 80.50\\
Chai et al. \cite{chai2020makes} & 68.44 & 55.59 & 64.37 & 64.10 & 58.74 & 84.48 & 59.92 & 72.08 & 47.63 & 73.05 & 68.38 & 58.98 & 77.05 & 76.87 & 76.35 & 75.97 & 77.41 & 74.68 & 71.91 & 68.74\\
Ojha et al. \cite{ojha2023towards} & 100.0 & 99.46 & 99.59 & 97.24 & 99.98 & 99.60 & 82.45 & 61.32 & 79.02 & 96.72 & 99.00 & 87.77 & 99.14 & 92.15 & 99.17 & 94.74 & 95.34 & 94.57 & 97.15 & 93.38 \\ 
Wang et al. \cite{wang2023dire} &100.0 & 76.73 & 72.80 & 97.06 & 68.44 & 100.0 & 98.55 & 54.51 & 65.62 & 97.10 & 93.74 & 94.29 & 95.17 & 95.43 & 95.77 & 96.18 & 97.30 & 97.53 & 68.73 & 87.63\\

Corvi et al. \cite{corvi2023detection} &100.00 & 98.60 & 99.80 & 100.0& 99.80 & 100.0& 94.70& 99.80& 87.70 & 100.00 & 100.00 & 73.00 & 86.80& 89.40   & 87.30& 86.50 & 89.90 & 89.00 & 96.10 & \_\\ 
Zhu et al. \cite{zhu2023gendet} & 99.95 & 99.95 & 99.92 & 99.92 & 99.92 & 99.25 & 91.38 & 61.23 & 72.66 & 97.90 & 98.88 & 99.30 & 99.85 & 99.51 & 99.85 & 99.50 & 99.46 & 99.19 & 99.47 & 95.64\\
Liu et al. \cite{liu2024mixture} & 100.0 & 99.85 & 99.88 & 99.69 & 100.0 & 99.68 & 87.38 & 88.26 & 84.48 & 98.82 & 99.84 & 93.39 & 99.81 & 96.80 & 99.88 & 98.71 & 98.84 & 98.60 & 98.81 & 96.99\\
Cao et al. \cite{cao2024hyperdet} & 100.0 & 99.96 & 99.89 & 99.73 & 99.93 & 100.0 & 88.38 & 97.12 & 89.22 & 98.82 & 99.98 & 95.31 & 99.86 & 99.14 & 99.90 & 97.20 & 97.99 & 98.02 & 99.65 & 97.90\\
Tan et al. \cite{tan2023learning} &100.0& 94.00 & 90.70 & 99.90& 79.30 & 100.0& 67.90& \_ & \_ & \_ & \_  & 100.0 & 99.10 & 99.20 & 99.20& 93.20 & 95.10 & 94.90 & 97.30 & \_\\
Liu et al. \cite{liu2024fatformer}&100.0& 100.0 & 100.0 & 99.80& 100.0 & 100.0& 98.00& \_ & \_ & \_ & \_  & 92.00 & 99.80 & 99.10 & 99.90& 99.10 & 99.40 & 99.20 & 99.80 & \_\\
Koutlis et al. \cite{koutlis2025leveraging} & 100.0 & 100.00 & 99.90 & 99.40 & 100.0& 100.0& 97.90 & 97.20 & 94.90 & 97.30 & 99.70 & 96.40 & 99.80 & 98.30 & 99.90 & 98.80 &99.30 &98.90 & 99.30 &  \_   \\ 
\hline
\end{tabular}
}
\end{table*}

\subsection{Commercial Detection Methods}
Watermarking by DeepMind \cite{synthid2024deepmind} is a commercial technology that introduce watermarking to the generated image invisible to human eyes, that work as basis for the detection of the synthesized image properly.

\section{Discussion and Limitation}
Each category of forensic detection contributes distinct insights into identifying AI-generated imagery, forming a complementary ecosystem of methodologies. Spatial-domain and frequency-domain analyses remain the foundation of pixel- and signal-level forensic understanding. Spatial detectors such as DIRE~\cite{wang2023dire} and SelfCon~\cite{boychev2024imaginet} have moved beyond early CNN-based designs by integrating reconstruction and contrastive learning for improved generalization, whereas frequency-based frameworks like SAFE~\cite{li2024improving} and AIDE~\cite{yan2024sanity} extend robustness by coupling spectral cues with semantic embeddings. Together, these advances demonstrate that spatial–frequency fusion provides a more holistic view of generation artifacts, though both remain limited under heavy compression and geometric transformations. Fingerprint and patch-based methods address these weaknesses from orthogonal perspectives. Fingerprint-based detectors such as MaskSim~\cite{li2024masksim} and LGrad~\cite{tan2023learning} mitigate model dependence through spectral and gradient-informed fingerprints, achieving cross-generator consistency absent in earlier handcrafted designs. Patch-based frameworks, including IPD-Net~\cite{chen2024ipd} and PatchCraft~\cite{zhong2024patchcraft}, enhance sensitivity to localized texture inconsistencies while maintaining contextual awareness through inter-patch relationships. These directions collectively narrow the gap between local artifact discovery and global scene understanding. Successful yet, these methods require extensive training or finetuning, which is computationally expensive, as well as limiting in scope since they require retraining when new and advanced generative models enter the market. In recent years, these limitations have been addressed by training-free approaches such as RIGID~\cite{he2024rigid}, HFI~\cite{choi2024hfi}, and ZED~\cite{cozzolino2024zero}, which depart from traditional learning paradigms by requiring no training on AI-generated images, instead leveraging statistical stability, frequency aliasing, or entropy deviation. While these methods excel in scalability and generator-agnostic detection with low computational cost, they still lack interpretability and resilience to adversarial perturbations.

Even more recently, vision-language and reasoning-based multimodal models—spanning CLIP-based frameworks like HyperDet~\cite{cao2024hyperdet} to reasoning-driven systems such as ForenX~\cite{tan2025forenx} and ThinkFake~\cite{huang2025thinkfake}—push the field toward semantically aligned and explainable forensics with reasoning. These approaches overcome the semantic blindness of classical detectors, offering text-grounded rationales for model decisions, though they rely heavily on computationally expensive foundation backbones and curated instruction data. Importantly, Fan et al.~\cite{fan2024fake} present detection that is fully dependent on LLMs without integrating understanding between text–vision data and still provide comparable results. While this is a great perspective on the usage of LLMs, it might suffer in generalizability or from hallucination, as LLMs may provide contextual reasoning with incorrect information during detection.

Finally, commercial detection frameworks unify several of these perspectives by prioritizing scalability and usability over transparency. They often integrate pretrained multimodal backbones and frequency cues for enterprise-level deployment, but their proprietary nature limits academic reproducibility and benchmarking.

For clarity, we present a comparative analysis of methods across six distinct categories, following a general review of their reported performances in the literature. The results are detailed in Table \ref{tab:detection-results-visionlangugage}, Table \ref{tab:combined_evaluations_fingerprint}, Table \ref{tab:combined_evaluations_patchbased}, and Table \ref{tab:diffusion_trainingfree}. Additionally, a comprehensive performance comparison across all six categories using the UnivFD dataset \cite{ojha2023towards} is provided in Table \ref{tab:comprehensive-accuracy-detection-results} for classification accuracy and Table \ref{tab:comprehensive-averageprecision-detection-results} for average precision. The evaluation results on the UnivFD dataset \cite{ojha2023towards} are selected due to its inclusion of state-of-the-art detection methods, particularly those utilizing vision-language approaches, which have demonstrated strong forensic capabilities in synthetic image detection. The recent perspective of reasoning provided by Multimodal Large Language Models (MLLMs) helps in moving from a black box to a grey box understanding—explaining why content is generated for the general public. A limitation we faced for these methods' quantitative comparative analysis is that they are recent and present new datasets, often compared with general MLLM models that were not designed for forensic detection; such comparisons differ from the scope of this study. Therefore, we only include the quantitative evaluation of ForenX~\cite{tan2025forenx} by merging it with training-free methods in Table \ref{tab:diffusion_trainingfree}.

While previous methods from spatial-domain, frequency-domain, fingerprint-based, and patch-based categories provide good detection results on different datasets, they may suffer from generalizability issues and computational expense due to dependence on the training of generative methods. Training-free methods could provide generalizability and computational efficiency, yet they may fail when the generated images exhibit very high fidelity. MLLM (VLM) methods could offer strong detection performance with reasoning—aligning with recent trends in the AI field—but they may be limited by extensive computational requirements and may suffer from the inherent issue of LLM hallucination. A hybrid structure leveraging the cross-category functions of MLLM and training-free methods may provide a new perspective on robustness, reasoning, generalizability, and efficiency.

\section{Datasets}
\subsubsection{ForenSynths by Wang: 2020}
The dataset used by Wang et al. \cite{wang2020cnn} is publicly available on their GitHub page, as referenced in the main paper. The training set consists of 724,000 images, including 362,000 real images and 362,000 fake images generated by ProGAN \cite{karras2017progressive}, trained on the LSUN dataset \cite{yu2015lsun} across 20 different object categories. The testing dataset, which includes images generated by various other GAN models, is also accessible on the GitHub page.

\subsubsection{Artifact Dataset by Rahman: 2023}
Rahman et al. \cite{rahman2023artifact} introduced the Artifact dataset, containing 2,496,738 images, including 964,989 real images and 1,531,749 synthetic images generated using 25 different models (both GANs and diffusion models). The dataset includes diverse object categories representative of general real-world content.

\subsubsection{SynthBuster by Bammey: 2023}
Bammey et al. \cite{Synth10334046} introduced the SynthBuster dataset, consisting of 9,000 synthetic images, with 1,000 images generated by each of nine different diffusion models, namely: DALL·E 2 \cite{ramesh2022hierarchical}, DALL·E 3 \cite{ramesh2022hierarchical}, Adobe Firefly \cite{adobe_firefly}, Midjourney v5 \cite{midjourney}, Stable Diffusion \cite{rombach2022high} 1.3, 1.4, 2, and XL, and GLIDE \cite{nichol2021glide}. The dataset is publicly available, with access details provided in the main paper. For real images, the authors recommend using the RAISE-1k dataset \cite{dang2015raise}, accessible via the same portal.

\subsubsection{DiffusionForensics Dataset by Wang et al.: 2023}
Wang et al. \cite{wang2023dire} introduced the DiffusionForensics dataset, specifically designed to evaluate forensic detectors for diffusion-generated images. The dataset includes images sourced from LSUN-Bedroom \cite{yu2015lsun}, ImageNet \cite{russakovsky2015imagenet}, and CelebA-HQ \cite{karras2017progressive}, with the following distributions:  
- 42,000 synthetic images generated from LSUN-Bedroom,  
- 50,000 synthetic images generated from ImageNet, and  
- 42,000 synthetic face images generated using Stable Diffusion V2 (SD-V2), paired with 42,000 real images from CelebA-HQ.

\subsubsection{UnivFD Dataset by Ojha et al.: 2023}
Ojha et al. \cite{ojha2023towards} expanded the ForenSynths dataset by integrating images generated from various models, including ProGAN \cite{karras2017progressive}, CycleGAN \cite{zhu2017unpaired}, BigGAN \cite{brock2018large}, StyleGAN \cite{karras2019style}, GauGAN \cite{park2019semantic}, StarGAN \cite{choi2018stargan}, Deepfakes \cite{rossler2019faceforensics++}, SITD \cite{chen2018learning}, SAN \cite{dai2019second}, CRN \cite{chen2017photographic}, and IMLE \cite{li2019diverse}. In addition, diffusion-based models such as the Guided diffusion model (ADM) \cite{dhariwal2021diffusion}, LDM \cite{rombach2022high}, GLIDE \cite{nichol2021glide}, and DALL-E \cite{ramesh2021zero} were included, increasing its diversity for forensic detection studies.

\subsubsection{Community Forensics by Park: 2024}
Many existing datasets contain images generated by diverse generative models; however, they often lack diversity and generalization capability for many-to-many scenarios. These scenarios include testing images generated from multiple diffusion models against those from various GANs, as well as incorporating generative models such as VAEs. Recognizing the need for a large, diverse dataset capable of detecting synthetic images from unseen generative models, Park and Owens \cite{park2024community} introduced the Community Forensics Dataset. This dataset comprises 2.4 million images, generated by 4,803 distinct generative models, alongside an equal number of real images, making it more diverse than previous datasets. To evaluate its effectiveness, the authors trained and tested state-of-the-art models on this dataset, utilizing a binary classification setup with pre-trained models ViT-S \cite{dosovitskiy2020image} and ConvNeXt-S \cite{liu2022convnet} as backbone architectures. Unlike previous methods that freeze backbone layers, the authors fine-tuned the entire backbone end-to-end, achieving better performance compared to existing state-of-the-art methods on both the Community Forensics dataset and other publicly available benchmarks.

\subsubsection{GenImage by Zhu: 2024}
Zhu et al. \cite{zhu2024genimage} introduced a large-scale dataset containing synthetic images generated by advanced state-of-the-art GANs and diffusion models (Wukong \cite{Wukong} and VQDM \cite{gu2022vector}, including commercial models, alongside real images. The dataset comprises approximately 1.3 million synthetic images and 1.3 million real images, covering diverse general image content. The authors also propose two real-world analysis factors to assess detection performance:  
(a) Cross-Generator Image Classification: training detectors on images generated by one model and evaluating them on images from different generators, and (b) Degraded Image Classification: testing detectors on degraded images affected by factors such as low resolution, JPEG compression, and Gaussian blur.

\subsubsection{CIFAKE Dataset by Bird and Lotfi: 2024}
Bird and Lotfi \cite{bird2024cifake} released a dataset comprising 120,000 images, equally divided between 60,000 real images and 60,000 synthetic images. The real images originate from CIFAR-10 \cite{krizhevsky2009learning}, which includes ten object categories: airplane, automobile, bird, cat, deer, dog, frog, horse, ship, and truck. The synthetic images were generated using diffusion models, including commercially available ones.

\subsubsection{AIGCDetection Benchmark Dataset by Zhong et al.: 2024}
Zhong et al. \cite{zhong2024patchcraft} introduced a curated benchmark dataset incorporating images generated by 17 different generative models, including GANs and diffusion models. The dataset aggregates samples from ForenSynths \cite{wang2020cnn} and GenImage \cite{zhu2024genimage}, and includes additional images from recent diffusion models, offering a comprehensive benchmark for generative image forensics.

\subsubsection{ImagiNet Dataset by Boychev: 2024}
Boychev et al. \cite{boychev2024imaginet} presented the ImagiNet dataset, comprising 200,000 images equally split between real and synthetic categories. The dataset further categorizes images into photos, paintings, faces, and uncategorized types. The synthetic images are generated using GANs, diffusion models, and proprietary generative models. For evaluation, the dataset is divided into 160,000 training images and 40,000 test images.

\subsubsection{Chameleon Dataset by Yan et al.: 2024}
Yan et al. \cite{yan2024sanity} introduced the Chameleon dataset, which focuses on realistic AI-generated image detection. The dataset consists of 150,000 synthetic images covering content categories such as humans, animals, scenes, and objects, generated using GANs and diffusion models. In addition, it includes 20,000 real images to facilitate forensic evaluation.

\subsection{Metrics}
The performance of synthetic image detection is primarily evaluated using classification accuracy (Acc) and average precision (AP), which are mathematically expressed as follows:

\begin{equation}
    \text{Acc} = \frac{TP + TN}{TP + TN + FP + FN}
\end{equation}

\begin{equation}
    \text{AP} = \sum_n (R_n - R_{n-1}) P_n
\end{equation}

where \( TP \), \( TN \), \( FP \), and \( FN \) represent true positives, true negatives, false positives, and false negatives, respectively. The AP is computed as the weighted sum of precision \( P_n \) at different recall levels \( R_n \).

\section{Conclusion}  
The advancement of generative AI in image synthesis and forensic detection has been an ongoing and increasingly active research area, driven by both the successes of AI and the ethical concerns it introduces. From a practical standpoint, forensic detection must remain ahead to mitigate potential human-related harm. While several reviews exist in the current literature, they do not comprehensively cover detection methods capable of identifying images generated by state-of-the-art generative models, nor do they adequately address the role of multimodal approaches in detection. To bridge this gap, we have conducted a comprehensive survey of detection methodologies. First, we categorize these methods into seven primary groups based on their underlying approaches. We then describe each method in detail, followed by an analysis of their comparative performance on publicly available datasets, and assess whether they satisfy three key criteria for evaluating their generalizability. Our findings indicate that detection methods leveraging multimodal frameworks tend to exhibit greater robustness and adaptability across different generative models.  Building upon this comparative analysis, we observe that spatial- and frequency-domain methods form the foundational layer of forensic understanding by identifying low-level pixel and spectral inconsistencies, yet they often struggle under compression or strong generative realism. Fingerprint- and patch-based methods address these weaknesses through architecture-agnostic learning and localized texture analysis, improving sensitivity and cross-generator robustness. Training-free approaches further enhance scalability by eliminating retraining needs, although interpretability and adversarial resilience remain open challenges. Finally, multimodal and reasoning-based models, particularly those integrating vision-language and large language models, bring explainability and semantic grounding but at the expense of heavy computational requirements.  
Looking ahead, the convergence of these paradigms offers a promising pathway. Hybrid frameworks that combine the reasoning and semantic grounding of multimodal models with the lightweight efficiency of training-free detection could lead to robust, interpretable, and real-time forensic systems. Future research should also emphasize cross-modal benchmarking and unified evaluation standards. Through these directions, the field can advance toward trustworthy and transparent generative AI systems capable of safeguarding authenticity in the digital era.

\printcredits

\bibliographystyle{cas-model2-names}


\begin{thebibliography}{108}


\ifx \showCODEN    \undefined \def \showCODEN     #1{\unskip}     \fi
\ifx \showDOI      \undefined \def \showDOI       #1{#1}\fi
\ifx \showISBNx    \undefined \def \showISBNx     #1{\unskip}     \fi
\ifx \showISBNxiii \undefined \def \showISBNxiii  #1{\unskip}     \fi
\ifx \showISSN     \undefined \def \showISSN      #1{\unskip}     \fi
\ifx \showLCCN     \undefined \def \showLCCN      #1{\unskip}     \fi
\ifx \shownote     \undefined \def \shownote      #1{#1}          \fi
\ifx \showarticletitle \undefined \def \showarticletitle #1{#1}   \fi
\ifx \showURL      \undefined \def \showURL       {\relax}        \fi
\providecommand\bibfield[2]{#2}
\providecommand\bibinfo[2]{#2}
\providecommand\natexlab[1]{#1}
\providecommand\showeprint[2][]{arXiv:#2}

\bibitem[{Adobe Inc.}(2024)]%
        {adobe_firefly}
\bibfield{author}{\bibinfo{person}{{Adobe Inc.}}} \bibinfo{year}{2024}\natexlab{}.
\newblock \bibinfo{title}{Adobe Firefly}.
\newblock \bibinfo{howpublished}{\url{https://www.adobe.com/products/firefly}}.
\newblock
\newblock
\shownote{Accessed: 2024-12-08}.


\bibitem[Ahmed et~al\mbox{.}(1974)]%
        {ahmed1974discrete}
\bibfield{author}{\bibinfo{person}{Nasir Ahmed}, \bibinfo{person}{T\_ Natarajan}, {and} \bibinfo{person}{Kamisetty~R Rao}.} \bibinfo{year}{1974}\natexlab{}.
\newblock \showarticletitle{Discrete cosine transform}.
\newblock \bibinfo{journal}{\emph{IEEE Trans. Comput.}} \bibinfo{volume}{100}, \bibinfo{number}{1} (\bibinfo{year}{1974}), \bibinfo{pages}{90--93}.
\newblock


\bibitem[Bammey(2024)]%
        {Synth10334046}
\bibfield{author}{\bibinfo{person}{Quentin Bammey}.} \bibinfo{year}{2024}\natexlab{}.
\newblock \showarticletitle{Synthbuster: Towards Detection of Diffusion Model Generated Images}.
\newblock \bibinfo{journal}{\emph{IEEE Open Journal of Signal Processing}}  \bibinfo{volume}{5} (\bibinfo{year}{2024}), \bibinfo{pages}{1--9}.
\newblock
\urldef\tempurl%
\url{https://doi.org/10.1109/OJSP.2023.3337714}
\showDOI{\tempurl}


\bibitem[Baraheem and Nguyen(2023)]%
        {baraheem2023ai}
\bibfield{author}{\bibinfo{person}{Samah~S Baraheem} {and} \bibinfo{person}{Tam~V Nguyen}.} \bibinfo{year}{2023}\natexlab{}.
\newblock \showarticletitle{AI vs. AI: Can AI Detect AI-Generated Images?}
\newblock \bibinfo{journal}{\emph{Journal of Imaging}} \bibinfo{volume}{9}, \bibinfo{number}{10} (\bibinfo{year}{2023}), \bibinfo{pages}{199}.
\newblock


\bibitem[Baron~Fourier et~al\mbox{.}(2003)]%
        {baron2003analytical}
\bibfield{author}{\bibinfo{person}{Jean Baptiste~Joseph Baron~Fourier} {et~al\mbox{.}}} \bibinfo{year}{2003}\natexlab{}.
\newblock \bibinfo{booktitle}{\emph{The analytical theory of heat}}.
\newblock \bibinfo{publisher}{Courier Corporation}.
\newblock


\bibitem[Bird and Lotfi(2024)]%
        {bird2024cifake}
\bibfield{author}{\bibinfo{person}{Jordan~J Bird} {and} \bibinfo{person}{Ahmad Lotfi}.} \bibinfo{year}{2024}\natexlab{}.
\newblock \showarticletitle{Cifake: Image classification and explainable identification of ai-generated synthetic images}.
\newblock \bibinfo{journal}{\emph{IEEE Access}} (\bibinfo{year}{2024}).
\newblock


\bibitem[Boychev and Cholakov(2024)]%
        {boychev2024imaginet}
\bibfield{author}{\bibinfo{person}{Delyan Boychev} {and} \bibinfo{person}{Radostin Cholakov}.} \bibinfo{year}{2024}\natexlab{}.
\newblock \showarticletitle{ImagiNet: A Multi-Content Dataset for Generalizable Synthetic Image Detection via Contrastive Learning}.
\newblock \bibinfo{journal}{\emph{arXiv preprint arXiv:2407.20020}} (\bibinfo{year}{2024}).
\newblock


\bibitem[Brock(2018a)]%
        {brock2018big}
\bibfield{author}{\bibinfo{person}{Andrew Brock}.} \bibinfo{year}{2018}\natexlab{a}.
\newblock \showarticletitle{Large Scale GAN Training for High Fidelity Natural Image Synthesis}.
\newblock \bibinfo{journal}{\emph{arXiv preprint arXiv:1809.11096}} (\bibinfo{year}{2018}).
\newblock


\bibitem[Brock(2018b)]%
        {brock2018large}
\bibfield{author}{\bibinfo{person}{Andrew Brock}.} \bibinfo{year}{2018}\natexlab{b}.
\newblock \showarticletitle{Large Scale GAN Training for High Fidelity Natural Image Synthesis}.
\newblock \bibinfo{journal}{\emph{arXiv preprint arXiv:1809.11096}} (\bibinfo{year}{2018}).
\newblock


\bibitem[Cao et~al\mbox{.}(2024)]%
        {cao2024hyperdet}
\bibfield{author}{\bibinfo{person}{Huangsen Cao}, \bibinfo{person}{Yongwei Wang}, \bibinfo{person}{Yinfeng Liu}, \bibinfo{person}{Sixian Zheng}, \bibinfo{person}{Kangtao Lv}, \bibinfo{person}{Zhimeng Zhang}, \bibinfo{person}{Bo Zhang}, \bibinfo{person}{Xin Ding}, {and} \bibinfo{person}{Fei Wu}.} \bibinfo{year}{2024}\natexlab{}.
\newblock \showarticletitle{HyperDet: Generalizable Detection of Synthesized Images by Generating and Merging A Mixture of Hyper LoRAs}.
\newblock \bibinfo{journal}{\emph{arXiv preprint arXiv:2410.06044}} (\bibinfo{year}{2024}).
\newblock


\bibitem[Chai et~al\mbox{.}(2020)]%
        {chai2020makes}
\bibfield{author}{\bibinfo{person}{Lucy Chai}, \bibinfo{person}{David Bau}, \bibinfo{person}{Ser-Nam Lim}, {and} \bibinfo{person}{Phillip Isola}.} \bibinfo{year}{2020}\natexlab{}.
\newblock \showarticletitle{What makes fake images detectable? Understanding properties that generalize}. In \bibinfo{booktitle}{\emph{Computer Vision--ECCV 2020: 16th European Conference, Glasgow, UK, August 23--28, 2020, Proceedings, Part XXVI 16}}. Springer, \bibinfo{pages}{103--120}.
\newblock


\bibitem[Chen et~al\mbox{.}(2018)]%
        {chen2018learning}
\bibfield{author}{\bibinfo{person}{Chen Chen}, \bibinfo{person}{Qifeng Chen}, \bibinfo{person}{Jia Xu}, {and} \bibinfo{person}{Vladlen Koltun}.} \bibinfo{year}{2018}\natexlab{}.
\newblock \showarticletitle{Learning to see in the dark}. In \bibinfo{booktitle}{\emph{Proceedings of the IEEE conference on computer vision and pattern recognition}}. \bibinfo{pages}{3291--3300}.
\newblock


\bibitem[Chen et~al\mbox{.}(2024a)]%
        {chen2024ipd}
\bibfield{author}{\bibinfo{person}{Jiahan Chen}, \bibinfo{person}{Mengtin Lo}, \bibinfo{person}{Hailiang Liao}, {and} \bibinfo{person}{Tianlin Huang}.} \bibinfo{year}{2024}\natexlab{a}.
\newblock \showarticletitle{IPD-Net: Detecting AI-Generated Images via Inter-Patch Dependencies}.
\newblock \bibinfo{journal}{\emph{International Journal of Advanced Computer Science \& Applications}} \bibinfo{volume}{15}, \bibinfo{number}{7} (\bibinfo{year}{2024}).
\newblock


\bibitem[Chen et~al\mbox{.}(2024b)]%
        {chen2024single}
\bibfield{author}{\bibinfo{person}{Jiaxuan Chen}, \bibinfo{person}{Jieteng Yao}, {and} \bibinfo{person}{Li Niu}.} \bibinfo{year}{2024}\natexlab{b}.
\newblock \showarticletitle{A single simple patch is all you need for ai-generated image detection}.
\newblock \bibinfo{journal}{\emph{arXiv preprint arXiv:2402.01123}} (\bibinfo{year}{2024}).
\newblock


\bibitem[Chen and Koltun(2017)]%
        {chen2017photographic}
\bibfield{author}{\bibinfo{person}{Qifeng Chen} {and} \bibinfo{person}{Vladlen Koltun}.} \bibinfo{year}{2017}\natexlab{}.
\newblock \showarticletitle{Photographic image synthesis with cascaded refinement networks}. In \bibinfo{booktitle}{\emph{Proceedings of the IEEE international conference on computer vision}}. \bibinfo{pages}{1511--1520}.
\newblock


\bibitem[Chen et~al\mbox{.}(2020)]%
        {chen2020simplecontrastive}
\bibfield{author}{\bibinfo{person}{Ting Chen}, \bibinfo{person}{Simon Kornblith}, \bibinfo{person}{Mohammad Norouzi}, {and} \bibinfo{person}{Geoffrey Hinton}.} \bibinfo{year}{2020}\natexlab{}.
\newblock \showarticletitle{A simple framework for contrastive learning of visual representations}. In \bibinfo{booktitle}{\emph{International conference on machine learning}}. PMLR, \bibinfo{pages}{1597--1607}.
\newblock


\bibitem[Chen and Hsu(2008)]%
        {chen2008image}
\bibfield{author}{\bibinfo{person}{Yi-Lei Chen} {and} \bibinfo{person}{Chiou-Ting Hsu}.} \bibinfo{year}{2008}\natexlab{}.
\newblock \showarticletitle{Image tampering detection by blocking periodicity analysis in JPEG compressed images}. In \bibinfo{booktitle}{\emph{2008 IEEE 10th Workshop on Multimedia Signal Processing}}. IEEE, \bibinfo{pages}{803--808}.
\newblock


\bibitem[Choi et~al\mbox{.}(2018)]%
        {choi2018stargan}
\bibfield{author}{\bibinfo{person}{Yunjey Choi}, \bibinfo{person}{Minje Choi}, \bibinfo{person}{Munyoung Kim}, \bibinfo{person}{Jung-Woo Ha}, \bibinfo{person}{Sunghun Kim}, {and} \bibinfo{person}{Jaegul Choo}.} \bibinfo{year}{2018}\natexlab{}.
\newblock \showarticletitle{Stargan: Unified generative adversarial networks for multi-domain image-to-image translation}. In \bibinfo{booktitle}{\emph{Proceedings of the IEEE conference on computer vision and pattern recognition}}. \bibinfo{pages}{8789--8797}.
\newblock


\bibitem[Chollet(2017)]%
        {chollet2017xception}
\bibfield{author}{\bibinfo{person}{Fran{\c{c}}ois Chollet}.} \bibinfo{year}{2017}\natexlab{}.
\newblock \showarticletitle{Xception: Deep learning with depthwise separable convolutions}. In \bibinfo{booktitle}{\emph{Proceedings of the IEEE conference on computer vision and pattern recognition}}. \bibinfo{pages}{1251--1258}.
\newblock


\bibitem[Cooley and Tukey(1965)]%
        {cooley1965algorithm}
\bibfield{author}{\bibinfo{person}{James~W Cooley} {and} \bibinfo{person}{John~W Tukey}.} \bibinfo{year}{1965}\natexlab{}.
\newblock \showarticletitle{An algorithm for the machine calculation of complex Fourier series}.
\newblock \bibinfo{journal}{\emph{Math. Comp.}} \bibinfo{volume}{19}, \bibinfo{number}{90} (\bibinfo{year}{1965}), \bibinfo{pages}{297--301}.
\newblock


\bibitem[Corvi et~al\mbox{.}(2023)]%
        {corvi2023detection}
\bibfield{author}{\bibinfo{person}{Riccardo Corvi}, \bibinfo{person}{Davide Cozzolino}, \bibinfo{person}{Giada Zingarini}, \bibinfo{person}{Giovanni Poggi}, \bibinfo{person}{Koki Nagano}, {and} \bibinfo{person}{Luisa Verdoliva}.} \bibinfo{year}{2023}\natexlab{}.
\newblock \showarticletitle{On the detection of synthetic images generated by diffusion models}. In \bibinfo{booktitle}{\emph{ICASSP 2023-2023 IEEE International Conference on Acoustics, Speech and Signal Processing (ICASSP)}}. IEEE, \bibinfo{pages}{1--5}.
\newblock


\bibitem[Cozzolino et~al\mbox{.}(2024)]%
        {cozzolino2024raising}
\bibfield{author}{\bibinfo{person}{Davide Cozzolino}, \bibinfo{person}{Giovanni Poggi}, \bibinfo{person}{Riccardo Corvi}, \bibinfo{person}{Matthias Nie{\ss}ner}, {and} \bibinfo{person}{Luisa Verdoliva}.} \bibinfo{year}{2024}\natexlab{}.
\newblock \showarticletitle{Raising the Bar of AI-generated Image Detection with CLIP}. In \bibinfo{booktitle}{\emph{Proceedings of the IEEE/CVF Conference on Computer Vision and Pattern Recognition}}. \bibinfo{pages}{4356--4366}.
\newblock


\bibitem[Cozzolino and Verdoliva(2019)]%
        {cozzolino2019noiseprint}
\bibfield{author}{\bibinfo{person}{Davide Cozzolino} {and} \bibinfo{person}{Luisa Verdoliva}.} \bibinfo{year}{2019}\natexlab{}.
\newblock \showarticletitle{Noiseprint: A CNN-based camera model fingerprint}.
\newblock \bibinfo{journal}{\emph{IEEE Transactions on Information Forensics and Security}}  \bibinfo{volume}{15} (\bibinfo{year}{2019}), \bibinfo{pages}{144--159}.
\newblock


\bibitem[Dai et~al\mbox{.}(2019)]%
        {dai2019second}
\bibfield{author}{\bibinfo{person}{Tao Dai}, \bibinfo{person}{Jianrui Cai}, \bibinfo{person}{Yongbing Zhang}, \bibinfo{person}{Shu-Tao Xia}, {and} \bibinfo{person}{Lei Zhang}.} \bibinfo{year}{2019}\natexlab{}.
\newblock \showarticletitle{Second-order attention network for single image super-resolution}. In \bibinfo{booktitle}{\emph{Proceedings of the IEEE/CVF conference on computer vision and pattern recognition}}. \bibinfo{pages}{11065--11074}.
\newblock


\bibitem[Dang-Nguyen et~al\mbox{.}(2015)]%
        {dang2015raise}
\bibfield{author}{\bibinfo{person}{Duc-Tien Dang-Nguyen}, \bibinfo{person}{Cecilia Pasquini}, \bibinfo{person}{Valentina Conotter}, {and} \bibinfo{person}{Giulia Boato}.} \bibinfo{year}{2015}\natexlab{}.
\newblock \showarticletitle{Raise: A raw images dataset for digital image forensics}. In \bibinfo{booktitle}{\emph{Proceedings of the 6th ACM multimedia systems conference}}. \bibinfo{pages}{219--224}.
\newblock


\bibitem[DeepMind(2024)]%
        {synthid2024deepmind}
\bibfield{author}{\bibinfo{person}{DeepMind}.} \bibinfo{year}{2024}\natexlab{}.
\newblock \bibinfo{title}{Identifying AI-Generated Images with SynthID}.
\newblock
\newblock
\urldef\tempurl%
\url{https://deepmind.google/technologies/synthid/}
\showURL{%
\tempurl}
\newblock
\shownote{Accessed: 2024-12-02}.


\bibitem[Dhariwal and Nichol(2021)]%
        {dhariwal2021diffusion}
\bibfield{author}{\bibinfo{person}{Prafulla Dhariwal} {and} \bibinfo{person}{Alexander Nichol}.} \bibinfo{year}{2021}\natexlab{}.
\newblock \showarticletitle{Diffusion models beat gans on image synthesis}.
\newblock \bibinfo{journal}{\emph{Advances in Neural Information Processing Systems}}  \bibinfo{volume}{34} (\bibinfo{year}{2021}), \bibinfo{pages}{8780--8794}.
\newblock


\bibitem[Dosovitskiy(2020)]%
        {dosovitskiy2020image}
\bibfield{author}{\bibinfo{person}{Alexey Dosovitskiy}.} \bibinfo{year}{2020}\natexlab{}.
\newblock \showarticletitle{An image is worth 16x16 words: Transformers for image recognition at scale}.
\newblock \bibinfo{journal}{\emph{arXiv preprint arXiv:2010.11929}} (\bibinfo{year}{2020}).
\newblock


\bibitem[Dzanic et~al\mbox{.}(2020)]%
        {dzanic2020fourier}
\bibfield{author}{\bibinfo{person}{Tarik Dzanic}, \bibinfo{person}{Karan Shah}, {and} \bibinfo{person}{Freddie Witherden}.} \bibinfo{year}{2020}\natexlab{}.
\newblock \showarticletitle{Fourier spectrum discrepancies in deep network generated images}.
\newblock \bibinfo{journal}{\emph{Advances in Neural Information Processing Systems}}  \bibinfo{volume}{33} (\bibinfo{year}{2020}), \bibinfo{pages}{3022--3032}.
\newblock


\bibitem[Fridrich and Kodovsky(2012)]%
        {fridrich2012richSRM}
\bibfield{author}{\bibinfo{person}{Jessica Fridrich} {and} \bibinfo{person}{Jan Kodovsky}.} \bibinfo{year}{2012}\natexlab{}.
\newblock \showarticletitle{Rich models for steganalysis of digital images}.
\newblock \bibinfo{journal}{\emph{IEEE Transactions on Information Forensics and Security}} \bibinfo{volume}{7}, \bibinfo{number}{3} (\bibinfo{year}{2012}), \bibinfo{pages}{868--882}.
\newblock


\bibitem[Goebel et~al\mbox{.}(2020)]%
        {goebel2020detection}
\bibfield{author}{\bibinfo{person}{Michael Goebel}, \bibinfo{person}{Lakshmanan Nataraj}, \bibinfo{person}{Tejaswi Nanjundaswamy}, \bibinfo{person}{Tajuddin~Manhar Mohammed}, \bibinfo{person}{Shivkumar Chandrasekaran}, {and} \bibinfo{person}{BS Manjunath}.} \bibinfo{year}{2020}\natexlab{}.
\newblock \showarticletitle{Detection, attribution and localization of gan generated images}.
\newblock \bibinfo{journal}{\emph{arXiv preprint arXiv:2007.10466}} (\bibinfo{year}{2020}).
\newblock


\bibitem[Goodfellow et~al\mbox{.}(2014)]%
        {goodfellow2014generative}
\bibfield{author}{\bibinfo{person}{Ian Goodfellow}, \bibinfo{person}{Jean Pouget-Abadie}, \bibinfo{person}{Mehdi Mirza}, \bibinfo{person}{Bing Xu}, \bibinfo{person}{David Warde-Farley}, \bibinfo{person}{Sherjil Ozair}, \bibinfo{person}{Aaron Courville}, {and} \bibinfo{person}{Yoshua Bengio}.} \bibinfo{year}{2014}\natexlab{}.
\newblock \showarticletitle{Generative adversarial nets}.
\newblock \bibinfo{journal}{\emph{Advances in Neural Information Processing Systems}}  \bibinfo{volume}{27} (\bibinfo{year}{2014}).
\newblock


\bibitem[{Google DeepMind}(2024)]%
        {imagen3}
\bibfield{author}{\bibinfo{person}{{Google DeepMind}}.} \bibinfo{year}{2024}\natexlab{}.
\newblock \bibinfo{title}{Imagen 3}.
\newblock \bibinfo{howpublished}{\url{https://deepmind.google/technologies/imagen-3}}.
\newblock
\newblock
\shownote{Accessed: 2024-12-08}.


\bibitem[Gu et~al\mbox{.}(2022)]%
        {gu2022vector}
\bibfield{author}{\bibinfo{person}{Shuyang Gu}, \bibinfo{person}{Dong Chen}, \bibinfo{person}{Jianmin Bao}, \bibinfo{person}{Fang Wen}, \bibinfo{person}{Bo Zhang}, \bibinfo{person}{Dongdong Chen}, \bibinfo{person}{Lu Yuan}, {and} \bibinfo{person}{Baining Guo}.} \bibinfo{year}{2022}\natexlab{}.
\newblock \showarticletitle{Vector quantized diffusion model for text-to-image synthesis}. In \bibinfo{booktitle}{\emph{Proceedings of the IEEE/CVF conference on computer vision and pattern recognition}}. \bibinfo{pages}{10696--10706}.
\newblock


\bibitem[Haar(1911)]%
        {haar1911theorie}
\bibfield{author}{\bibinfo{person}{Alfred Haar}.} \bibinfo{year}{1911}\natexlab{}.
\newblock \showarticletitle{Zur theorie der orthogonalen funktionensysteme}.
\newblock \bibinfo{journal}{\emph{Math. Ann.}} \bibinfo{volume}{71}, \bibinfo{number}{1} (\bibinfo{year}{1911}), \bibinfo{pages}{38--53}.
\newblock


\bibitem[He et~al\mbox{.}(2016)]%
        {he2016deep}
\bibfield{author}{\bibinfo{person}{Kaiming He}, \bibinfo{person}{Xiangyu Zhang}, \bibinfo{person}{Shaoqing Ren}, {and} \bibinfo{person}{Jian Sun}.} \bibinfo{year}{2016}\natexlab{}.
\newblock \showarticletitle{Deep residual learning for image recognition}. In \bibinfo{booktitle}{\emph{Proceedings of the IEEE conference on computer vision and pattern recognition}}. \bibinfo{pages}{770--778}.
\newblock


\bibitem[Ho et~al\mbox{.}(2020)]%
        {ho2020denoising}
\bibfield{author}{\bibinfo{person}{Jonathan Ho}, \bibinfo{person}{Ajay Jain}, {and} \bibinfo{person}{Pieter Abbeel}.} \bibinfo{year}{2020}\natexlab{}.
\newblock \showarticletitle{Denoising diffusion probabilistic models}.
\newblock \bibinfo{journal}{\emph{Advances in Neural Information Processing Systems}}  \bibinfo{volume}{33} (\bibinfo{year}{2020}), \bibinfo{pages}{6840--6851}.
\newblock


\bibitem[Jeong et~al\mbox{.}(2022)]%
        {jeong2022fingerprintnet}
\bibfield{author}{\bibinfo{person}{Yonghyun Jeong}, \bibinfo{person}{Doyeon Kim}, \bibinfo{person}{Youngmin Ro}, \bibinfo{person}{Pyounggeon Kim}, {and} \bibinfo{person}{Jongwon Choi}.} \bibinfo{year}{2022}\natexlab{}.
\newblock \showarticletitle{Fingerprintnet: Synthesized fingerprints for generated image detection}. In \bibinfo{booktitle}{\emph{European Conference on Computer Vision}}. Springer, \bibinfo{pages}{76--94}.
\newblock


\bibitem[Jiang et~al\mbox{.}(2021)]%
        {jiang2021layercam}
\bibfield{author}{\bibinfo{person}{Peng-Tao Jiang}, \bibinfo{person}{Chang-Bin Zhang}, \bibinfo{person}{Qibin Hou}, \bibinfo{person}{Ming-Ming Cheng}, {and} \bibinfo{person}{Yunchao Wei}.} \bibinfo{year}{2021}\natexlab{}.
\newblock \showarticletitle{Layercam: Exploring hierarchical class activation maps for localization}.
\newblock \bibinfo{journal}{\emph{IEEE Transactions on Image Processing}}  \bibinfo{volume}{30} (\bibinfo{year}{2021}), \bibinfo{pages}{5875--5888}.
\newblock


\bibitem[Ju et~al\mbox{.}(2022)]%
        {ju2022fusing}
\bibfield{author}{\bibinfo{person}{Yan Ju}, \bibinfo{person}{Shan Jia}, \bibinfo{person}{Lipeng Ke}, \bibinfo{person}{Hongfei Xue}, \bibinfo{person}{Koki Nagano}, {and} \bibinfo{person}{Siwei Lyu}.} \bibinfo{year}{2022}\natexlab{}.
\newblock \showarticletitle{Fusing global and local features for generalized ai-synthesized image detection}. In \bibinfo{booktitle}{\emph{2022 IEEE International Conference on Image Processing (ICIP)}}. IEEE, \bibinfo{pages}{3465--3469}.
\newblock


\bibitem[Karras(2017)]%
        {karras2017progressive}
\bibfield{author}{\bibinfo{person}{Tero Karras}.} \bibinfo{year}{2017}\natexlab{}.
\newblock \showarticletitle{Progressive Growing of GANs for Improved Quality, Stability, and Variation}.
\newblock \bibinfo{journal}{\emph{arXiv preprint arXiv:1710.10196}} (\bibinfo{year}{2017}).
\newblock


\bibitem[Karras et~al\mbox{.}(2019)]%
        {karras2019style}
\bibfield{author}{\bibinfo{person}{Tero Karras}, \bibinfo{person}{Samuli Laine}, {and} \bibinfo{person}{Timo Aila}.} \bibinfo{year}{2019}\natexlab{}.
\newblock \showarticletitle{A style-based generator architecture for generative adversarial networks}. In \bibinfo{booktitle}{\emph{Proceedings of the IEEE/CVF conference on computer vision and pattern recognition}}. \bibinfo{pages}{4401--4410}.
\newblock


\bibitem[Ke et~al\mbox{.}(2017)]%
        {ke2017lightgbm}
\bibfield{author}{\bibinfo{person}{Guolin Ke}, \bibinfo{person}{Qi Meng}, \bibinfo{person}{Thomas Finley}, \bibinfo{person}{Taifeng Wang}, \bibinfo{person}{Wei Chen}, \bibinfo{person}{Weidong Ma}, \bibinfo{person}{Qiwei Ye}, {and} \bibinfo{person}{Tie-Yan Liu}.} \bibinfo{year}{2017}\natexlab{}.
\newblock \showarticletitle{Lightgbm: A highly efficient gradient boosting decision tree}.
\newblock \bibinfo{journal}{\emph{Advances in Neural Information Processing Systems}}  \bibinfo{volume}{30} (\bibinfo{year}{2017}).
\newblock


\bibitem[Kingma(2013)]%
        {kingma2013auto}
\bibfield{author}{\bibinfo{person}{Diederik~P Kingma}.} \bibinfo{year}{2013}\natexlab{}.
\newblock \showarticletitle{Auto-encoding variational bayes}.
\newblock \bibinfo{journal}{\emph{arXiv preprint arXiv:1312.6114}} (\bibinfo{year}{2013}).
\newblock


\bibitem[Koutlis and Papadopoulos(2025)]%
        {koutlis2025leveraging}
\bibfield{author}{\bibinfo{person}{Christos Koutlis} {and} \bibinfo{person}{Symeon Papadopoulos}.} \bibinfo{year}{2025}\natexlab{}.
\newblock \showarticletitle{Leveraging representations from intermediate encoder-blocks for synthetic image detection}. In \bibinfo{booktitle}{\emph{European Conference on Computer Vision}}. Springer, \bibinfo{pages}{394--411}.
\newblock


\bibitem[Krizhevsky et~al\mbox{.}(2009)]%
        {krizhevsky2009learning}
\bibfield{author}{\bibinfo{person}{Alex Krizhevsky}, \bibinfo{person}{Geoffrey Hinton}, {et~al\mbox{.}}} \bibinfo{year}{2009}\natexlab{}.
\newblock \showarticletitle{Learning multiple layers of features from tiny images}.
\newblock  (\bibinfo{year}{2009}).
\newblock


\bibitem[Li et~al\mbox{.}(2019)]%
        {li2019diverse}
\bibfield{author}{\bibinfo{person}{Ke Li}, \bibinfo{person}{Tianhao Zhang}, {and} \bibinfo{person}{Jitendra Malik}.} \bibinfo{year}{2019}\natexlab{}.
\newblock \showarticletitle{Diverse image synthesis from semantic layouts via conditional IMLE}. In \bibinfo{booktitle}{\emph{Proceedings of the IEEE/CVF International Conference on Computer Vision}}. \bibinfo{pages}{4220--4229}.
\newblock


\bibitem[Li et~al\mbox{.}(2024b)]%
        {li2024improving}
\bibfield{author}{\bibinfo{person}{Ouxiang Li}, \bibinfo{person}{Jiayin Cai}, \bibinfo{person}{Yanbin Hao}, \bibinfo{person}{Xiaolong Jiang}, \bibinfo{person}{Yao Hu}, {and} \bibinfo{person}{Fuli Feng}.} \bibinfo{year}{2024}\natexlab{b}.
\newblock \showarticletitle{Improving Synthetic Image Detection Towards Generalization: An Image Transformation Perspective}.
\newblock \bibinfo{journal}{\emph{arXiv preprint arXiv:2408.06741}} (\bibinfo{year}{2024}).
\newblock


\bibitem[Li et~al\mbox{.}(2022)]%
        {li2022ArtifactSimilarity}
\bibfield{author}{\bibinfo{person}{Weichuang Li}, \bibinfo{person}{Peisong He}, \bibinfo{person}{Haoliang Li}, \bibinfo{person}{Hongxia Wang}, {and} \bibinfo{person}{Ruimei Zhang}.} \bibinfo{year}{2022}\natexlab{}.
\newblock \showarticletitle{Detection of GAN-Generated Images by Estimating Artifact Similarity}.
\newblock \bibinfo{journal}{\emph{IEEE Signal Processing Letters}}  \bibinfo{volume}{29} (\bibinfo{year}{2022}), \bibinfo{pages}{862--866}.
\newblock
\urldef\tempurl%
\url{https://doi.org/10.1109/LSP.2021.3130525}
\showDOI{\tempurl}


\bibitem[Li et~al\mbox{.}(2024a)]%
        {li2024masksim}
\bibfield{author}{\bibinfo{person}{Yanhao Li}, \bibinfo{person}{Quentin Bammey}, \bibinfo{person}{Marina Gardella}, \bibinfo{person}{Tina Nikoukhah}, \bibinfo{person}{Jean-Michel Morel}, \bibinfo{person}{Miguel Colom}, {and} \bibinfo{person}{Rafael~Grompone Von~Gioi}.} \bibinfo{year}{2024}\natexlab{a}.
\newblock \showarticletitle{MaskSim: Detection of Synthetic Images by Masked Spectrum Similarity Analysis}. In \bibinfo{booktitle}{\emph{Proceedings of the IEEE/CVF Conference on Computer Vision and Pattern Recognition}}. \bibinfo{pages}{3855--3865}.
\newblock


\bibitem[Liu et~al\mbox{.}(2022b)]%
        {liu2022detecting}
\bibfield{author}{\bibinfo{person}{Bo Liu}, \bibinfo{person}{Fan Yang}, \bibinfo{person}{Xiuli Bi}, \bibinfo{person}{Bin Xiao}, \bibinfo{person}{Weisheng Li}, {and} \bibinfo{person}{Xinbo Gao}.} \bibinfo{year}{2022}\natexlab{b}.
\newblock \showarticletitle{Detecting generated images by real images}. In \bibinfo{booktitle}{\emph{European Conference on Computer Vision}}. Springer, \bibinfo{pages}{95--110}.
\newblock


\bibitem[Liu et~al\mbox{.}(2024a)]%
        {liu2024fatformer}
\bibfield{author}{\bibinfo{person}{Huan Liu}, \bibinfo{person}{Zichang Tan}, \bibinfo{person}{Chuangchuang Tan}, \bibinfo{person}{Yunchao Wei}, \bibinfo{person}{Jingdong Wang}, {and} \bibinfo{person}{Yao Zhao}.} \bibinfo{year}{2024}\natexlab{a}.
\newblock \showarticletitle{Forgery-aware adaptive transformer for generalizable synthetic image detection}. In \bibinfo{booktitle}{\emph{Proceedings of the IEEE/CVF Conference on Computer Vision and Pattern Recognition}}. \bibinfo{pages}{10770--10780}.
\newblock


\bibitem[Liu et~al\mbox{.}(2022a)]%
        {liu2022convnet}
\bibfield{author}{\bibinfo{person}{Zhuang Liu}, \bibinfo{person}{Hanzi Mao}, \bibinfo{person}{Chao-Yuan Wu}, \bibinfo{person}{Christoph Feichtenhofer}, \bibinfo{person}{Trevor Darrell}, {and} \bibinfo{person}{Saining Xie}.} \bibinfo{year}{2022}\natexlab{a}.
\newblock \showarticletitle{A convnet for the 2020s}. In \bibinfo{booktitle}{\emph{Proceedings of the IEEE/CVF conference on computer vision and pattern recognition}}. \bibinfo{pages}{11976--11986}.
\newblock


\bibitem[Liu et~al\mbox{.}(2024b)]%
        {liu2024mixture}
\bibfield{author}{\bibinfo{person}{Zihan Liu}, \bibinfo{person}{Hanyi Wang}, \bibinfo{person}{Yaoyu Kang}, {and} \bibinfo{person}{Shilin Wang}.} \bibinfo{year}{2024}\natexlab{b}.
\newblock \showarticletitle{Mixture of Low-rank Experts for Transferable AI-Generated Image Detection}.
\newblock \bibinfo{journal}{\emph{arXiv preprint arXiv:2404.04883}} (\bibinfo{year}{2024}).
\newblock


\bibitem[Lorenz et~al\mbox{.}(2023)]%
        {lorenz2023detecting}
\bibfield{author}{\bibinfo{person}{Peter Lorenz}, \bibinfo{person}{Ricard~L Durall}, {and} \bibinfo{person}{Janis Keuper}.} \bibinfo{year}{2023}\natexlab{}.
\newblock \showarticletitle{Detecting images generated by deep diffusion models using their local intrinsic dimensionality}. In \bibinfo{booktitle}{\emph{Proceedings of the IEEE/CVF International Conference on Computer Vision}}. \bibinfo{pages}{448--459}.
\newblock


\bibitem[Lorenz et~al\mbox{.}(2022)]%
        {lorenz2022unfolding}
\bibfield{author}{\bibinfo{person}{Peter Lorenz}, \bibinfo{person}{Margret Keuper}, {and} \bibinfo{person}{Janis Keuper}.} \bibinfo{year}{2022}\natexlab{}.
\newblock \showarticletitle{Unfolding local growth rate estimates for (almost) perfect adversarial detection}.
\newblock \bibinfo{journal}{\emph{arXiv preprint arXiv:2212.06776}} (\bibinfo{year}{2022}).
\newblock


\bibitem[Lukas et~al\mbox{.}(2006)]%
        {lukas2006digital}
\bibfield{author}{\bibinfo{person}{Jan Lukas}, \bibinfo{person}{Jessica Fridrich}, {and} \bibinfo{person}{Miroslav Goljan}.} \bibinfo{year}{2006}\natexlab{}.
\newblock \showarticletitle{Digital camera identification from sensor pattern noise}.
\newblock \bibinfo{journal}{\emph{IEEE Transactions on Information Forensics and Security}} \bibinfo{volume}{1}, \bibinfo{number}{2} (\bibinfo{year}{2006}), \bibinfo{pages}{205--214}.
\newblock


\bibitem[Mandelli et~al\mbox{.}(2022)]%
        {mandelli2022detecting}
\bibfield{author}{\bibinfo{person}{Sara Mandelli}, \bibinfo{person}{Nicol{\`o} Bonettini}, \bibinfo{person}{Paolo Bestagini}, {and} \bibinfo{person}{Stefano Tubaro}.} \bibinfo{year}{2022}\natexlab{}.
\newblock \showarticletitle{Detecting gan-generated images by orthogonal training of multiple cnns}. In \bibinfo{booktitle}{\emph{2022 IEEE International Conference on Image Processing (ICIP)}}. IEEE, \bibinfo{pages}{3091--3095}.
\newblock


\bibitem[Marra et~al\mbox{.}(2019)]%
        {marra2019gans}
\bibfield{author}{\bibinfo{person}{Francesco Marra}, \bibinfo{person}{Diego Gragnaniello}, \bibinfo{person}{Luisa Verdoliva}, {and} \bibinfo{person}{Giovanni Poggi}.} \bibinfo{year}{2019}\natexlab{}.
\newblock \showarticletitle{Do GANs leave artificial fingerprints?}. In \bibinfo{booktitle}{\emph{2019 IEEE conference on multimedia information processing and retrieval (MIPR)}}. IEEE, \bibinfo{pages}{506--511}.
\newblock


\bibitem[Meng et~al\mbox{.}(2024)]%
        {meng2024artifact}
\bibfield{author}{\bibinfo{person}{Zheling Meng}, \bibinfo{person}{Bo Peng}, \bibinfo{person}{Jing Dong}, \bibinfo{person}{Tieniu Tan}, {and} \bibinfo{person}{Haonan Cheng}.} \bibinfo{year}{2024}\natexlab{}.
\newblock \showarticletitle{Artifact feature purification for cross-domain detection of AI-generated images}.
\newblock \bibinfo{journal}{\emph{Computer Vision and Image Understanding}}  \bibinfo{volume}{247} (\bibinfo{year}{2024}), \bibinfo{pages}{104078}.
\newblock


\bibitem[{MidJourney}(2024)]%
        {midjourney}
\bibfield{author}{\bibinfo{person}{{MidJourney}}.} \bibinfo{year}{2024}\natexlab{}.
\newblock \bibinfo{title}{MidJourney}.
\newblock \bibinfo{howpublished}{\url{https://www.midjourney.com/}}.
\newblock
\newblock
\shownote{Accessed: 2024-12-08}.


\bibitem[Mylrea(2025)]%
        {mylrea2025generative}
\bibfield{author}{\bibinfo{person}{Michael Mylrea}.} \bibinfo{year}{2025}\natexlab{}.
\newblock \showarticletitle{The generative AI weapon of mass destruction: Evolving disinformation threats, vulnerabilities, and mitigation frameworks}.
\newblock In \bibinfo{booktitle}{\emph{Interdependent Human-Machine Teams}}. \bibinfo{publisher}{Elsevier}, \bibinfo{pages}{315--347}.
\newblock


\bibitem[Nataraj et~al\mbox{.}(2019)]%
        {nataraj2019detecting}
\bibfield{author}{\bibinfo{person}{Lakshmanan Nataraj}, \bibinfo{person}{Tajuddin~Manhar Mohammed}, \bibinfo{person}{Shivkumar Chandrasekaran}, \bibinfo{person}{Arjuna Flenner}, \bibinfo{person}{Jawadul~H Bappy}, \bibinfo{person}{Amit~K Roy-Chowdhury}, {and} \bibinfo{person}{BS Manjunath}.} \bibinfo{year}{2019}\natexlab{}.
\newblock \showarticletitle{Detecting GAN generated fake images using co-occurrence matrices}.
\newblock \bibinfo{journal}{\emph{arXiv preprint arXiv:1903.06836}} (\bibinfo{year}{2019}).
\newblock


\bibitem[Nichol et~al\mbox{.}(2021)]%
        {nichol2021glide}
\bibfield{author}{\bibinfo{person}{Alex Nichol}, \bibinfo{person}{Prafulla Dhariwal}, \bibinfo{person}{Aditya Ramesh}, \bibinfo{person}{Pranav Shyam}, \bibinfo{person}{Pamela Mishkin}, \bibinfo{person}{Bob McGrew}, \bibinfo{person}{Ilya Sutskever}, {and} \bibinfo{person}{Mark Chen}.} \bibinfo{year}{2021}\natexlab{}.
\newblock \showarticletitle{Glide: Towards photorealistic image generation and editing with text-guided diffusion models}.
\newblock \bibinfo{journal}{\emph{arXiv preprint arXiv:2112.10741}} (\bibinfo{year}{2021}).
\newblock


\bibitem[Ojha et~al\mbox{.}(2023)]%
        {ojha2023towards}
\bibfield{author}{\bibinfo{person}{Utkarsh Ojha}, \bibinfo{person}{Yuheng Li}, {and} \bibinfo{person}{Yong~Jae Lee}.} \bibinfo{year}{2023}\natexlab{}.
\newblock \showarticletitle{Towards universal fake image detectors that generalize across generative models}. In \bibinfo{booktitle}{\emph{Proceedings of the IEEE/CVF Conference on Computer Vision and Pattern Recognition}}. \bibinfo{pages}{24480--24489}.
\newblock


\bibitem[{OpenAI}(2024)]%
        {dalle3}
\bibfield{author}{\bibinfo{person}{{OpenAI}}.} \bibinfo{year}{2024}\natexlab{}.
\newblock \bibinfo{title}{DALL·E 3}.
\newblock \bibinfo{howpublished}{\url{https://openai.com/index/dall-e-3/}}.
\newblock
\newblock
\shownote{Accessed: 2024-12-08}.


\bibitem[Park and Owens(2024)]%
        {park2024community}
\bibfield{author}{\bibinfo{person}{Jeongsoo Park} {and} \bibinfo{person}{Andrew Owens}.} \bibinfo{year}{2024}\natexlab{}.
\newblock \showarticletitle{Community Forensics: Using Thousands of Generators to Train Fake Image Detectors}.
\newblock \bibinfo{journal}{\emph{arXiv preprint arXiv:2411.04125}} (\bibinfo{year}{2024}).
\newblock


\bibitem[Park et~al\mbox{.}(2019)]%
        {park2019semantic}
\bibfield{author}{\bibinfo{person}{Taesung Park}, \bibinfo{person}{Ming-Yu Liu}, \bibinfo{person}{Ting-Chun Wang}, {and} \bibinfo{person}{Jun-Yan Zhu}.} \bibinfo{year}{2019}\natexlab{}.
\newblock \showarticletitle{Semantic image synthesis with spatially-adaptive normalization}. In \bibinfo{booktitle}{\emph{Proceedings of the IEEE/CVF conference on computer vision and pattern recognition}}. \bibinfo{pages}{2337--2346}.
\newblock


\bibitem[Parmar et~al\mbox{.}(2018)]%
        {parmar2018image}
\bibfield{author}{\bibinfo{person}{Niki Parmar}, \bibinfo{person}{Ashish Vaswani}, \bibinfo{person}{Jakob Uszkoreit}, \bibinfo{person}{Lukasz Kaiser}, \bibinfo{person}{Noam Shazeer}, \bibinfo{person}{Alexander Ku}, {and} \bibinfo{person}{Dustin Tran}.} \bibinfo{year}{2018}\natexlab{}.
\newblock \showarticletitle{Image transformer}. In \bibinfo{booktitle}{\emph{International conference on machine learning}}. PMLR, \bibinfo{pages}{4055--4064}.
\newblock


\bibitem[Radford et~al\mbox{.}(2021)]%
        {radford2021learning}
\bibfield{author}{\bibinfo{person}{Alec Radford}, \bibinfo{person}{Jong~Wook Kim}, \bibinfo{person}{Chris Hallacy}, \bibinfo{person}{Aditya Ramesh}, \bibinfo{person}{Gabriel Goh}, \bibinfo{person}{Sandhini Agarwal}, \bibinfo{person}{Girish Sastry}, \bibinfo{person}{Amanda Askell}, \bibinfo{person}{Pamela Mishkin}, \bibinfo{person}{Jack Clark}, {et~al\mbox{.}}} \bibinfo{year}{2021}\natexlab{}.
\newblock \showarticletitle{Learning transferable visual models from natural language supervision}. In \bibinfo{booktitle}{\emph{International conference on machine learning}}. PMLR, \bibinfo{pages}{8748--8763}.
\newblock


\bibitem[Rahman et~al\mbox{.}(2023)]%
        {rahman2023artifact}
\bibfield{author}{\bibinfo{person}{Md~Awsafur Rahman}, \bibinfo{person}{Bishmoy Paul}, \bibinfo{person}{Najibul~Haque Sarker}, \bibinfo{person}{Zaber Ibn~Abdul Hakim}, {and} \bibinfo{person}{Shaikh~Anowarul Fattah}.} \bibinfo{year}{2023}\natexlab{}.
\newblock \showarticletitle{Artifact: A large-scale dataset with artificial and factual images for generalizable and robust synthetic image detection}. In \bibinfo{booktitle}{\emph{2023 IEEE International Conference on Image Processing (ICIP)}}. IEEE, \bibinfo{pages}{2200--2204}.
\newblock


\bibitem[Ramaswamy et~al\mbox{.}(2020)]%
        {ramaswamy2020ablationcam}
\bibfield{author}{\bibinfo{person}{Harish~Guruprasad Ramaswamy} {et~al\mbox{.}}} \bibinfo{year}{2020}\natexlab{}.
\newblock \showarticletitle{Ablation-cam: Visual explanations for deep convolutional network via gradient-free localization}. In \bibinfo{booktitle}{\emph{Proceedings of the IEEE/CVF winter conference on applications of computer vision}}. \bibinfo{pages}{983--991}.
\newblock


\bibitem[Ramesh et~al\mbox{.}(2022)]%
        {ramesh2022hierarchical}
\bibfield{author}{\bibinfo{person}{Aditya Ramesh}, \bibinfo{person}{Prafulla Dhariwal}, \bibinfo{person}{Alex Nichol}, \bibinfo{person}{Casey Chu}, {and} \bibinfo{person}{Mark Chen}.} \bibinfo{year}{2022}\natexlab{}.
\newblock \showarticletitle{Hierarchical text-conditional image generation with clip latents}.
\newblock \bibinfo{journal}{\emph{arXiv preprint arXiv:2204.06125}} \bibinfo{volume}{1}, \bibinfo{number}{2} (\bibinfo{year}{2022}), \bibinfo{pages}{3}.
\newblock


\bibitem[Ramesh et~al\mbox{.}(2021)]%
        {ramesh2021zero}
\bibfield{author}{\bibinfo{person}{Aditya Ramesh}, \bibinfo{person}{Mikhail Pavlov}, \bibinfo{person}{Gabriel Goh}, \bibinfo{person}{Scott Gray}, \bibinfo{person}{Chelsea Voss}, \bibinfo{person}{Alec Radford}, \bibinfo{person}{Mark Chen}, {and} \bibinfo{person}{Ilya Sutskever}.} \bibinfo{year}{2021}\natexlab{}.
\newblock \showarticletitle{Zero-shot text-to-image generation}. In \bibinfo{booktitle}{\emph{International Conference on Machine Learning}}. Pmlr, \bibinfo{pages}{8821--8831}.
\newblock


\bibitem[Rezende and Mohamed(2015)]%
        {rezende2015variational}
\bibfield{author}{\bibinfo{person}{Danilo Rezende} {and} \bibinfo{person}{Shakir Mohamed}.} \bibinfo{year}{2015}\natexlab{}.
\newblock \showarticletitle{Variational inference with normalizing flows}. In \bibinfo{booktitle}{\emph{International conference on machine learning}}. PMLR, \bibinfo{pages}{1530--1538}.
\newblock


\bibitem[Rombach et~al\mbox{.}(2022)]%
        {rombach2022high}
\bibfield{author}{\bibinfo{person}{Robin Rombach}, \bibinfo{person}{Andreas Blattmann}, \bibinfo{person}{Dominik Lorenz}, \bibinfo{person}{Patrick Esser}, {and} \bibinfo{person}{Bj{\"o}rn Ommer}.} \bibinfo{year}{2022}\natexlab{}.
\newblock \showarticletitle{High-resolution image synthesis with latent diffusion models}. In \bibinfo{booktitle}{\emph{Proceedings of the IEEE/CVF conference on computer vision and pattern recognition}}. \bibinfo{pages}{10684--10695}.
\newblock


\bibitem[Rossler et~al\mbox{.}(2019)]%
        {rossler2019faceforensics++}
\bibfield{author}{\bibinfo{person}{Andreas Rossler}, \bibinfo{person}{Davide Cozzolino}, \bibinfo{person}{Luisa Verdoliva}, \bibinfo{person}{Christian Riess}, \bibinfo{person}{Justus Thies}, {and} \bibinfo{person}{Matthias Nie{\ss}ner}.} \bibinfo{year}{2019}\natexlab{}.
\newblock \showarticletitle{Faceforensics++: Learning to detect manipulated facial images}. In \bibinfo{booktitle}{\emph{Proceedings of the IEEE/CVF international conference on computer vision}}. \bibinfo{pages}{1--11}.
\newblock


\bibitem[Russakovsky et~al\mbox{.}(2015)]%
        {russakovsky2015imagenet}
\bibfield{author}{\bibinfo{person}{Olga Russakovsky}, \bibinfo{person}{Jia Deng}, \bibinfo{person}{Hao Su}, \bibinfo{person}{Jonathan Krause}, \bibinfo{person}{Sanjeev Satheesh}, \bibinfo{person}{Sean Ma}, \bibinfo{person}{Zhiheng Huang}, \bibinfo{person}{Andrej Karpathy}, \bibinfo{person}{Aditya Khosla}, \bibinfo{person}{Michael Bernstein}, {et~al\mbox{.}}} \bibinfo{year}{2015}\natexlab{}.
\newblock \showarticletitle{Imagenet large scale visual recognition challenge}.
\newblock \bibinfo{journal}{\emph{International Journal of Computer Vision}}  \bibinfo{volume}{115} (\bibinfo{year}{2015}), \bibinfo{pages}{211--252}.
\newblock


\bibitem[Sandrini and Somogyi(2023)]%
        {sandrini2023generative}
\bibfield{author}{\bibinfo{person}{Luca Sandrini} {and} \bibinfo{person}{Robert Somogyi}.} \bibinfo{year}{2023}\natexlab{}.
\newblock \showarticletitle{Generative AI and deceptive news consumption}.
\newblock \bibinfo{journal}{\emph{Economics Letters}}  \bibinfo{volume}{232} (\bibinfo{year}{2023}), \bibinfo{pages}{111317}.
\newblock


\bibitem[Selvaraju et~al\mbox{.}(2017)]%
        {selvaraju2017gradcam}
\bibfield{author}{\bibinfo{person}{Ramprasaath~R Selvaraju}, \bibinfo{person}{Michael Cogswell}, \bibinfo{person}{Abhishek Das}, \bibinfo{person}{Ramakrishna Vedantam}, \bibinfo{person}{Devi Parikh}, {and} \bibinfo{person}{Dhruv Batra}.} \bibinfo{year}{2017}\natexlab{}.
\newblock \showarticletitle{Grad-cam: Visual explanations from deep networks via gradient-based localization}. In \bibinfo{booktitle}{\emph{Proceedings of the IEEE international conference on computer vision}}. \bibinfo{pages}{618--626}.
\newblock


\bibitem[Sencar and Memon(2013)]%
        {sencar2013digital}
\bibfield{author}{\bibinfo{person}{Husrev~T Sencar} {and} \bibinfo{person}{Nasir Memon}.} \bibinfo{year}{2013}\natexlab{}.
\newblock \bibinfo{booktitle}{\emph{Digital image forensics}}.
\newblock \bibinfo{publisher}{Springer}.
\newblock


\bibitem[Sohl-Dickstein et~al\mbox{.}(2015)]%
        {sohl2015deep}
\bibfield{author}{\bibinfo{person}{Jascha Sohl-Dickstein}, \bibinfo{person}{Eric Weiss}, \bibinfo{person}{Niru Maheswaranathan}, {and} \bibinfo{person}{Surya Ganguli}.} \bibinfo{year}{2015}\natexlab{}.
\newblock \showarticletitle{Deep unsupervised learning using nonequilibrium thermodynamics}. In \bibinfo{booktitle}{\emph{International conference on machine learning}}. PMLR, \bibinfo{pages}{2256--2265}.
\newblock


\bibitem[Song et~al\mbox{.}(2020)]%
        {song2020denoising}
\bibfield{author}{\bibinfo{person}{Jiaming Song}, \bibinfo{person}{Chenlin Meng}, {and} \bibinfo{person}{Stefano Ermon}.} \bibinfo{year}{2020}\natexlab{}.
\newblock \showarticletitle{Denoising diffusion implicit models}.
\newblock \bibinfo{journal}{\emph{arXiv preprint arXiv:2010.02502}} (\bibinfo{year}{2020}).
\newblock


\bibitem[Tan et~al\mbox{.}(2023)]%
        {tan2023learning}
\bibfield{author}{\bibinfo{person}{Chuangchuang Tan}, \bibinfo{person}{Yao Zhao}, \bibinfo{person}{Shikui Wei}, \bibinfo{person}{Guanghua Gu}, {and} \bibinfo{person}{Yunchao Wei}.} \bibinfo{year}{2023}\natexlab{}.
\newblock \showarticletitle{Learning on gradients: Generalized artifacts representation for gan-generated images detection}. In \bibinfo{booktitle}{\emph{Proceedings of the IEEE/CVF Conference on Computer Vision and Pattern Recognition}}. \bibinfo{pages}{12105--12114}.
\newblock


\bibitem[Tan and Le(2019)]%
        {tan2019efficientnet}
\bibfield{author}{\bibinfo{person}{Mingxing Tan} {and} \bibinfo{person}{Quoc Le}.} \bibinfo{year}{2019}\natexlab{}.
\newblock \showarticletitle{Efficientnet: Rethinking model scaling for convolutional neural networks}. In \bibinfo{booktitle}{\emph{International conference on machine learning}}. PMLR, \bibinfo{pages}{6105--6114}.
\newblock


\bibitem[Van~den Oord et~al\mbox{.}(2016)]%
        {van2016conditional}
\bibfield{author}{\bibinfo{person}{Aaron Van~den Oord}, \bibinfo{person}{Nal Kalchbrenner}, \bibinfo{person}{Lasse Espeholt}, \bibinfo{person}{Oriol Vinyals}, \bibinfo{person}{Alex Graves}, {et~al\mbox{.}}} \bibinfo{year}{2016}\natexlab{}.
\newblock \showarticletitle{Conditional image generation with PixelCNN decoders}.
\newblock \bibinfo{journal}{\emph{Advances in Neural Information Processing Systems}}  \bibinfo{volume}{29} (\bibinfo{year}{2016}).
\newblock


\bibitem[Van Den~Oord et~al\mbox{.}(2016)]%
        {van2016pixel}
\bibfield{author}{\bibinfo{person}{A{\"a}ron Van Den~Oord}, \bibinfo{person}{Nal Kalchbrenner}, {and} \bibinfo{person}{Koray Kavukcuoglu}.} \bibinfo{year}{2016}\natexlab{}.
\newblock \showarticletitle{Pixel recurrent neural networks}. In \bibinfo{booktitle}{\emph{International conference on machine learning}}. PMLR, \bibinfo{pages}{1747--1756}.
\newblock


\bibitem[Van~der Maaten and Hinton(2008)]%
        {van2008visualizing}
\bibfield{author}{\bibinfo{person}{Laurens Van~der Maaten} {and} \bibinfo{person}{Geoffrey Hinton}.} \bibinfo{year}{2008}\natexlab{}.
\newblock \showarticletitle{Visualizing data using t-SNE.}
\newblock \bibinfo{journal}{\emph{Journal of Machine Learning Research}} \bibinfo{volume}{9}, \bibinfo{number}{11} (\bibinfo{year}{2008}).
\newblock


\bibitem[Wang et~al\mbox{.}(2023b)]%
        {wang2023fingerprintAug}
\bibfield{author}{\bibinfo{person}{Huaming Wang}, \bibinfo{person}{Jianwei Fei}, \bibinfo{person}{Yunshu Dai}, \bibinfo{person}{Lingyun Leng}, {and} \bibinfo{person}{Zhihua Xia}.} \bibinfo{year}{2023}\natexlab{b}.
\newblock \showarticletitle{General GAN-generated image detection by data augmentation in fingerprint domain}. In \bibinfo{booktitle}{\emph{2023 IEEE International Conference on Multimedia and Expo (ICME)}}. IEEE, \bibinfo{pages}{1187--1192}.
\newblock


\bibitem[Wang et~al\mbox{.}(2020a)]%
        {wang2020scorecam}
\bibfield{author}{\bibinfo{person}{Haofan Wang}, \bibinfo{person}{Zifan Wang}, \bibinfo{person}{Mengnan Du}, \bibinfo{person}{Fan Yang}, \bibinfo{person}{Zijian Zhang}, \bibinfo{person}{Sirui Ding}, \bibinfo{person}{Piotr Mardziel}, {and} \bibinfo{person}{Xia Hu}.} \bibinfo{year}{2020}\natexlab{a}.
\newblock \showarticletitle{Score-CAM: Score-weighted visual explanations for convolutional neural networks}. In \bibinfo{booktitle}{\emph{Proceedings of the IEEE/CVF conference on computer vision and pattern recognition workshops}}. \bibinfo{pages}{24--25}.
\newblock


\bibitem[Wang et~al\mbox{.}(2020b)]%
        {wang2020cnn}
\bibfield{author}{\bibinfo{person}{Sheng-Yu Wang}, \bibinfo{person}{Oliver Wang}, \bibinfo{person}{Richard Zhang}, \bibinfo{person}{Andrew Owens}, {and} \bibinfo{person}{Alexei~A Efros}.} \bibinfo{year}{2020}\natexlab{b}.
\newblock \showarticletitle{CNN-generated images are surprisingly easy to spot... for now}. In \bibinfo{booktitle}{\emph{Proceedings of the IEEE/CVF conference on computer vision and pattern recognition}}. \bibinfo{pages}{8695--8704}.
\newblock


\bibitem[Wang et~al\mbox{.}(2024)]%
        {wang2024deepfake}
\bibfield{author}{\bibinfo{person}{Tianyi Wang}, \bibinfo{person}{Xin Liao}, \bibinfo{person}{Kam~Pui Chow}, \bibinfo{person}{Xiaodong Lin}, {and} \bibinfo{person}{Yinglong Wang}.} \bibinfo{year}{2024}\natexlab{}.
\newblock \showarticletitle{Deepfake detection: A comprehensive survey from the reliability perspective}.
\newblock \bibinfo{journal}{\emph{Comput. Surveys}} \bibinfo{volume}{57}, \bibinfo{number}{3} (\bibinfo{year}{2024}), \bibinfo{pages}{1--35}.
\newblock


\bibitem[Wang et~al\mbox{.}(2023a)]%
        {wang2023dire}
\bibfield{author}{\bibinfo{person}{Zhendong Wang}, \bibinfo{person}{Jianmin Bao}, \bibinfo{person}{Wengang Zhou}, \bibinfo{person}{Weilun Wang}, \bibinfo{person}{Hezhen Hu}, \bibinfo{person}{Hong Chen}, {and} \bibinfo{person}{Houqiang Li}.} \bibinfo{year}{2023}\natexlab{a}.
\newblock \showarticletitle{Dire for diffusion-generated image detection}. In \bibinfo{booktitle}{\emph{Proceedings of the IEEE/CVF International Conference on Computer Vision}}. \bibinfo{pages}{22445--22455}.
\newblock


\bibitem[Wu et~al\mbox{.}(2023)]%
        {wu2023generalizable}
\bibfield{author}{\bibinfo{person}{Haiwei Wu}, \bibinfo{person}{Jiantao Zhou}, {and} \bibinfo{person}{Shile Zhang}.} \bibinfo{year}{2023}\natexlab{}.
\newblock \showarticletitle{Generalizable synthetic image detection via language-guided contrastive learning}.
\newblock \bibinfo{journal}{\emph{arXiv preprint arXiv:2305.13800}} (\bibinfo{year}{2023}).
\newblock


\bibitem[{Wukong}(2022)]%
        {Wukong}
\bibfield{author}{\bibinfo{person}{{Wukong}}.} \bibinfo{year}{2022}\natexlab{}.
\newblock \bibinfo{title}{{Wukong}}.
\newblock
\newblock
\urldef\tempurl%
\url{https://xihe.mindspore.cn/modelzoo/wukong}
\showURL{%
\tempurl}


\bibitem[Yan et~al\mbox{.}(2024)]%
        {yan2024sanity}
\bibfield{author}{\bibinfo{person}{Shilin Yan}, \bibinfo{person}{Ouxiang Li}, \bibinfo{person}{Jiayin Cai}, \bibinfo{person}{Yanbin Hao}, \bibinfo{person}{Xiaolong Jiang}, \bibinfo{person}{Yao Hu}, {and} \bibinfo{person}{Weidi Xie}.} \bibinfo{year}{2024}\natexlab{}.
\newblock \showarticletitle{A sanity check for ai-generated image detection}.
\newblock \bibinfo{journal}{\emph{arXiv preprint arXiv:2406.19435}} (\bibinfo{year}{2024}).
\newblock


\bibitem[Yang et~al\mbox{.}(2021)]%
        {yang2021learning}
\bibfield{author}{\bibinfo{person}{Tianyun Yang}, \bibinfo{person}{Juan Cao}, \bibinfo{person}{Qiang Sheng}, \bibinfo{person}{Lei Li}, \bibinfo{person}{Jiaqi Ji}, \bibinfo{person}{Xirong Li}, {and} \bibinfo{person}{Sheng Tang}.} \bibinfo{year}{2021}\natexlab{}.
\newblock \showarticletitle{Learning to disentangle gan fingerprint for fake image attribution}.
\newblock \bibinfo{journal}{\emph{arXiv preprint arXiv:2106.08749}} (\bibinfo{year}{2021}).
\newblock


\bibitem[Yousaf et~al\mbox{.}(2022)]%
        {yousaf2022fake}
\bibfield{author}{\bibinfo{person}{Bilal Yousaf}, \bibinfo{person}{Muhammad Usama}, \bibinfo{person}{Waqas Sultani}, \bibinfo{person}{Arif Mahmood}, {and} \bibinfo{person}{Junaid Qadir}.} \bibinfo{year}{2022}\natexlab{}.
\newblock \showarticletitle{Fake visual content detection using two-stream convolutional neural networks}.
\newblock \bibinfo{journal}{\emph{Neural Computing and Applications}} \bibinfo{volume}{34}, \bibinfo{number}{10} (\bibinfo{year}{2022}), \bibinfo{pages}{7991--8004}.
\newblock


\bibitem[Yu et~al\mbox{.}(2015)]%
        {yu2015lsun}
\bibfield{author}{\bibinfo{person}{Fisher Yu}, \bibinfo{person}{Ari Seff}, \bibinfo{person}{Yinda Zhang}, \bibinfo{person}{Shuran Song}, \bibinfo{person}{Thomas Funkhouser}, {and} \bibinfo{person}{Jianxiong Xiao}.} \bibinfo{year}{2015}\natexlab{}.
\newblock \showarticletitle{Lsun: Construction of a large-scale image dataset using deep learning with humans in the loop}.
\newblock \bibinfo{journal}{\emph{arXiv preprint arXiv:1506.03365}} (\bibinfo{year}{2015}).
\newblock


\bibitem[Yu et~al\mbox{.}(2019)]%
        {yu2019attributing}
\bibfield{author}{\bibinfo{person}{Ning Yu}, \bibinfo{person}{Larry~S Davis}, {and} \bibinfo{person}{Mario Fritz}.} \bibinfo{year}{2019}\natexlab{}.
\newblock \showarticletitle{Attributing fake images to gans: Learning and analyzing gan fingerprints}. In \bibinfo{booktitle}{\emph{Proceedings of the IEEE/CVF international conference on computer vision}}. \bibinfo{pages}{7556--7566}.
\newblock


\bibitem[Zhang et~al\mbox{.}(2025)]%
        {zhang2025ai}
\bibfield{author}{\bibinfo{person}{Chuo~Jun Zhang}, \bibinfo{person}{Asif~Q Gill}, \bibinfo{person}{Bo Liu}, {and} \bibinfo{person}{Memoona~J Anwar}.} \bibinfo{year}{2025}\natexlab{}.
\newblock \showarticletitle{AI-based Identity Fraud Detection: A Systematic Review}.
\newblock \bibinfo{journal}{\emph{arXiv preprint arXiv:2501.09239}} (\bibinfo{year}{2025}).
\newblock


\bibitem[Zhang et~al\mbox{.}(2017)]%
        {zhang2017beyond}
\bibfield{author}{\bibinfo{person}{Kai Zhang}, \bibinfo{person}{Wangmeng Zuo}, \bibinfo{person}{Yunjin Chen}, \bibinfo{person}{Deyu Meng}, {and} \bibinfo{person}{Lei Zhang}.} \bibinfo{year}{2017}\natexlab{}.
\newblock \showarticletitle{Beyond a Gaussian denoiser: Residual learning of deep CNN for image denoising}.
\newblock \bibinfo{journal}{\emph{IEEE Transactions on Image Processing}} \bibinfo{volume}{26}, \bibinfo{number}{7} (\bibinfo{year}{2017}), \bibinfo{pages}{3142--3155}.
\newblock


\bibitem[Zhang et~al\mbox{.}(2019)]%
        {zhang2019detecting}
\bibfield{author}{\bibinfo{person}{Xu Zhang}, \bibinfo{person}{Svebor Karaman}, {and} \bibinfo{person}{Shih-Fu Chang}.} \bibinfo{year}{2019}\natexlab{}.
\newblock \showarticletitle{Detecting and simulating artifacts in gan fake images}. In \bibinfo{booktitle}{\emph{2019 IEEE international workshop on information forensics and security (WIFS)}}. IEEE, \bibinfo{pages}{1--6}.
\newblock


\bibitem[Zheng et~al\mbox{.}(2019)]%
        {zheng2019survey}
\bibfield{author}{\bibinfo{person}{Lilei Zheng}, \bibinfo{person}{Ying Zhang}, {and} \bibinfo{person}{Vrizlynn~LL Thing}.} \bibinfo{year}{2019}\natexlab{}.
\newblock \showarticletitle{A survey on image tampering and its detection in real-world photos}.
\newblock \bibinfo{journal}{\emph{Journal of Visual Communication and Image Representation}}  \bibinfo{volume}{58} (\bibinfo{year}{2019}), \bibinfo{pages}{380--399}.
\newblock


\bibitem[Zhong et~al\mbox{.}(2024)]%
        {zhong2024patchcraft}
\bibfield{author}{\bibinfo{person}{Nan Zhong}, \bibinfo{person}{Yiran Xu}, \bibinfo{person}{Sheng Li}, \bibinfo{person}{Zhenxing Qian}, {and} \bibinfo{person}{Xinpeng Zhang}.} \bibinfo{year}{2024}\natexlab{}.
\newblock \showarticletitle{Patchcraft: Exploring texture patch for efficient ai-generated image detection}.
\newblock \bibinfo{journal}{\emph{arXiv preprint arXiv:2311.12397}} (\bibinfo{year}{2024}), \bibinfo{pages}{1--18}.
\newblock


\bibitem[Zhu et~al\mbox{.}(2017)]%
        {zhu2017unpaired}
\bibfield{author}{\bibinfo{person}{Jun-Yan Zhu}, \bibinfo{person}{Taesung Park}, \bibinfo{person}{Phillip Isola}, {and} \bibinfo{person}{Alexei~A Efros}.} \bibinfo{year}{2017}\natexlab{}.
\newblock \showarticletitle{Unpaired image-to-image translation using cycle-consistent adversarial networks}. In \bibinfo{booktitle}{\emph{Proceedings of the IEEE international conference on computer vision}}. \bibinfo{pages}{2223--2232}.
\newblock


\bibitem[Zhu et~al\mbox{.}(2023)]%
        {zhu2023gendet}
\bibfield{author}{\bibinfo{person}{Mingjian Zhu}, \bibinfo{person}{Hanting Chen}, \bibinfo{person}{Mouxiao Huang}, \bibinfo{person}{Wei Li}, \bibinfo{person}{Hailin Hu}, \bibinfo{person}{Jie Hu}, {and} \bibinfo{person}{Yunhe Wang}.} \bibinfo{year}{2023}\natexlab{}.
\newblock \showarticletitle{Gendet: Towards good generalizations for ai-generated image detection}.
\newblock \bibinfo{journal}{\emph{arXiv preprint arXiv:2312.08880}} (\bibinfo{year}{2023}).
\newblock


\bibitem[Zhu et~al\mbox{.}(2024)]%
        {zhu2024genimage}
\bibfield{author}{\bibinfo{person}{Mingjian Zhu}, \bibinfo{person}{Hanting Chen}, \bibinfo{person}{Qiangyu Yan}, \bibinfo{person}{Xudong Huang}, \bibinfo{person}{Guanyu Lin}, \bibinfo{person}{Wei Li}, \bibinfo{person}{Zhijun Tu}, \bibinfo{person}{Hailin Hu}, \bibinfo{person}{Jie Hu}, {and} \bibinfo{person}{Yunhe Wang}.} \bibinfo{year}{2024}\natexlab{}.
\newblock \showarticletitle{Genimage: A million-scale benchmark for detecting ai-generated image}.
\newblock \bibinfo{journal}{\emph{Advances in Neural Information Processing Systems}}  \bibinfo{volume}{36} (\bibinfo{year}{2024}).


\bibitem[{Achiam et~al.(2023)Achiam, Adler, Agarwal, Ahmad, Akkaya, Aleman, Almeida, Altenschmidt, Altman, Anadkat et~al.}]{achiam2023gpt}
\bibinfo{author}{Achiam, J.}, \bibinfo{author}{Adler, S.}, \bibinfo{author}{Agarwal, S.}, \bibinfo{author}{Ahmad, L.}, \bibinfo{author}{Akkaya, I.}, \bibinfo{author}{Aleman, F.L.}, \bibinfo{author}{Almeida, D.}, \bibinfo{author}{Altenschmidt, J.}, \bibinfo{author}{Altman, S.}, \bibinfo{author}{Anadkat, S.}, et~al., \bibinfo{year}{2023}.
\newblock \bibinfo{title}{Gpt-4 technical report}.
\newblock \bibinfo{journal}{arXiv preprint arXiv:2303.08774} .
\bibitem[{Bai et~al.(2025)Bai, Chen, Liu, Wang, Ge, Song, Dang, Wang, Wang, Tang et~al.}]{bai2025qwen2}
\bibinfo{author}{Bai, S.}, \bibinfo{author}{Chen, K.}, \bibinfo{author}{Liu, X.}, \bibinfo{author}{Wang, J.}, \bibinfo{author}{Ge, W.}, \bibinfo{author}{Song, S.}, \bibinfo{author}{Dang, K.}, \bibinfo{author}{Wang, P.}, \bibinfo{author}{Wang, S.}, \bibinfo{author}{Tang, J.}, et~al., \bibinfo{year}{2025}.
\newblock \bibinfo{title}{Qwen2. 5-vl technical report}.
\newblock \bibinfo{journal}{arXiv preprint arXiv:2502.13923} .
\bibitem[{Cao et~al.(2020)Cao, Wu and Kr{\"a}henb{\"u}hl}]{cao2020lossless}
\bibinfo{author}{Cao, S.}, \bibinfo{author}{Wu, C.Y.}, \bibinfo{author}{Kr{\"a}henb{\"u}hl, P.}, \bibinfo{year}{2020}.
\newblock \bibinfo{title}{Lossless image compression through super-resolution}.
\newblock \bibinfo{journal}{arXiv preprint arXiv:2004.02872} .
\bibitem[{Choi et~al.(2024)Choi, Park, Lee, Kim, Choi and Lee}]{choi2024hfi}
\bibinfo{author}{Choi, S.}, \bibinfo{author}{Park, S.}, \bibinfo{author}{Lee, J.}, \bibinfo{author}{Kim, S.}, \bibinfo{author}{Choi, S.J.}, \bibinfo{author}{Lee, M.}, \bibinfo{year}{2024}.
\newblock \bibinfo{title}{Hfi: A unified framework for training-free detection and implicit watermarking of latent diffusion model generated images}.
\newblock \bibinfo{journal}{arXiv preprint arXiv:2412.20704} .
\bibitem[{Cozzolino et~al.(2024)Cozzolino, Poggi, Nie{\ss}ner and Verdoliva}]{cozzolino2024zero}
\bibinfo{author}{Cozzolino, D.}, \bibinfo{author}{Poggi, G.}, \bibinfo{author}{Nie{\ss}ner, M.}, \bibinfo{author}{Verdoliva, L.}, \bibinfo{year}{2024}.
\newblock \bibinfo{title}{Zero-shot detection of ai-generated images}, in: \bibinfo{booktitle}{European Conference on Computer Vision}, \bibinfo{organization}{Springer}. pp. \bibinfo{pages}{54--72}.
\bibitem[{Fan et~al.(2024)Fan, Yang, Zhang, Yang and Zou}]{fan2024fake}
\bibinfo{author}{Fan, Y.}, \bibinfo{author}{Yang, D.}, \bibinfo{author}{Zhang, J.}, \bibinfo{author}{Yang, B.}, \bibinfo{author}{Zou, Y.}, \bibinfo{year}{2024}.
\newblock \bibinfo{title}{Fake-gpt: Detecting fake image via large language model}, in: \bibinfo{booktitle}{Chinese Conference on Pattern Recognition and Computer Vision (PRCV)}, \bibinfo{organization}{Springer}. pp. \bibinfo{pages}{122--136}.
\bibitem[{He et~al.(2024)He, Chen and Ho}]{he2024rigid}
\bibinfo{author}{He, Z.}, \bibinfo{author}{Chen, P.Y.}, \bibinfo{author}{Ho, T.Y.}, \bibinfo{year}{2024}.
\newblock \bibinfo{title}{Rigid: A training-free and model-agnostic framework for robust ai-generated image detection}.
\newblock \bibinfo{journal}{arXiv preprint arXiv:2405.20112} .
\bibitem[{Huang et~al.(2025)Huang, Lin, Hua, Cheng, Yamagishi and Chen}]{huang2025thinkfake}
\bibinfo{author}{Huang, T.M.}, \bibinfo{author}{Lin, W.T.}, \bibinfo{author}{Hua, K.L.}, \bibinfo{author}{Cheng, W.H.}, \bibinfo{author}{Yamagishi, J.}, \bibinfo{author}{Chen, J.C.}, \bibinfo{year}{2025}.
\newblock \bibinfo{title}{Thinkfake: Reasoning in multimodal large language models for ai-generated image detection}.
\newblock \bibinfo{journal}{arXiv preprint arXiv:2509.19841} .
\bibitem[{Ji et~al.(2025a)Ji, Hong, Zhan, Chen, Zhu, Wang, Zhang, Zhang et~al.}]{ji2025towards}
\bibinfo{author}{Ji, Y.}, \bibinfo{author}{Hong, Y.}, \bibinfo{author}{Zhan, J.}, \bibinfo{author}{Chen, H.}, \bibinfo{author}{Zhu, H.}, \bibinfo{author}{Wang, W.}, \bibinfo{author}{Zhang, L.}, \bibinfo{author}{Zhang, J.}, et~al., \bibinfo{year}{2025}a.
\newblock \bibinfo{title}{Towards explainable fake image detection with multi-modal large language models}.
\newblock \bibinfo{journal}{arXiv preprint arXiv:2504.14245} .
\bibitem[{Ji et~al.(2025b)Ji, Yan, Lan, Zhu, Wang, Fan, Zhang and Zhang}]{ji2025interpretable}
\bibinfo{author}{Ji, Y.}, \bibinfo{author}{Yan, H.}, \bibinfo{author}{Lan, J.}, \bibinfo{author}{Zhu, H.}, \bibinfo{author}{Wang, W.}, \bibinfo{author}{Fan, Q.}, \bibinfo{author}{Zhang, L.}, \bibinfo{author}{Zhang, J.}, \bibinfo{year}{2025}b.
\newblock \bibinfo{title}{Interpretable and reliable detection of ai-generated images via grounded reasoning in mllms}.
\newblock \bibinfo{journal}{arXiv preprint arXiv:2506.07045} .
\bibitem[{Liu et~al.(2023)Liu, Li, Wu and Lee}]{liu2023visual}
\bibinfo{author}{Liu, H.}, \bibinfo{author}{Li, C.}, \bibinfo{author}{Wu, Q.}, \bibinfo{author}{Lee, Y.J.}, \bibinfo{year}{2023}.
\newblock \bibinfo{title}{Visual instruction tuning}.
\newblock \bibinfo{journal}{Advances in neural information processing systems} \bibinfo{volume}{36}, \bibinfo{pages}{34892--34916}.
\bibitem[{Oquab et~al.(2023)Oquab, Darcet, Moutakanni, Vo, Szafraniec, Khalidov, Fernandez, Haziza, Massa, El-Nouby et~al.}]{oquab2023dinov2}
\bibinfo{author}{Oquab, M.}, \bibinfo{author}{Darcet, T.}, \bibinfo{author}{Moutakanni, T.}, \bibinfo{author}{Vo, H.}, \bibinfo{author}{Szafraniec, M.}, \bibinfo{author}{Khalidov, V.}, \bibinfo{author}{Fernandez, P.}, \bibinfo{author}{Haziza, D.}, \bibinfo{author}{Massa, F.}, \bibinfo{author}{El-Nouby, A.}, et~al., \bibinfo{year}{2023}.
\newblock \bibinfo{title}{Dinov2: Learning robust visual features without supervision}.
\newblock \bibinfo{journal}{arXiv preprint arXiv:2304.07193} .
\bibitem[{Ricker et~al.(2024)Ricker, Lukovnikov and Fischer}]{ricker2024aeroblade}
\bibinfo{author}{Ricker, J.}, \bibinfo{author}{Lukovnikov, D.}, \bibinfo{author}{Fischer, A.}, \bibinfo{year}{2024}.
\newblock \bibinfo{title}{Aeroblade: Training-free detection of latent diffusion images using autoencoder reconstruction error}, in: \bibinfo{booktitle}{Proceedings of the IEEE/CVF Conference on Computer Vision and Pattern Recognition}, pp. \bibinfo{pages}{9130--9140}.
\bibitem[{Tan et~al.(2025)Tan, Wang, Ming, Tao, Wei, Zhao and Lu}]{tan2025forenx}
\bibinfo{author}{Tan, C.}, \bibinfo{author}{Wang, J.}, \bibinfo{author}{Ming, X.}, \bibinfo{author}{Tao, R.}, \bibinfo{author}{Wei, Y.}, \bibinfo{author}{Zhao, Y.}, \bibinfo{author}{Lu, Y.}, \bibinfo{year}{2025}.
\newblock \bibinfo{title}{Forenx: Towards explainable ai-generated image detection with multimodal large language models}.
\newblock \bibinfo{journal}{arXiv preprint arXiv:2508.01402} .
\bibitem[{Zhang et~al.(2018)Zhang, Isola, Efros, Shechtman and Wang}]{zhang2018unreasonable}
\bibinfo{author}{Zhang, R.}, \bibinfo{author}{Isola, P.}, \bibinfo{author}{Efros, A.A.}, \bibinfo{author}{Shechtman, E.}, \bibinfo{author}{Wang, O.}, \bibinfo{year}{2018}.
\newblock \bibinfo{title}{The unreasonable effectiveness of deep features as a perceptual metric}, in: \bibinfo{booktitle}{Proceedings of the IEEE conference on computer vision and pattern recognition}, pp. \bibinfo{pages}{586--595}.
\bibitem[{Zhou et~al.(2025)Zhou, Luo, Wu, Sun, Ji, Yan, Ding, Sun, Wu and Ji}]{zhou2025aigi}
\bibinfo{author}{Zhou, Z.}, \bibinfo{author}{Luo, Y.}, \bibinfo{author}{Wu, Y.}, \bibinfo{author}{Sun, K.}, \bibinfo{author}{Ji, J.}, \bibinfo{author}{Yan, K.}, \bibinfo{author}{Ding, S.}, \bibinfo{author}{Sun, X.}, \bibinfo{author}{Wu, Y.}, \bibinfo{author}{Ji, R.}, \bibinfo{year}{2025}.
\newblock \bibinfo{title}{Aigi-holmes: Towards explainable and generalizable ai-generated image detection via multimodal large language models}.
\newblock \bibinfo{journal}{arXiv preprint arXiv:2507.02664} .

\bibitem[{Tan et~al.(2024)Tan, Zhao, Wei, Gu, Liu and Wei}]{tan2024rethinking}
\bibinfo{author}{Tan, C.}, \bibinfo{author}{Zhao, Y.}, \bibinfo{author}{Wei, S.}, \bibinfo{author}{Gu, G.}, \bibinfo{author}{Liu, P.}, \bibinfo{author}{Wei, Y.}, \bibinfo{year}{2024}.
\newblock \bibinfo{title}{Rethinking the up-sampling operations in cnn-based generative network for generalizable deepfake detection}, in: \bibinfo{booktitle}{Proceedings of the IEEE/CVF Conference on Computer Vision and Pattern Recognition}, pp. \bibinfo{pages}{28130--28139}.


\newblock


\end{thebibliography}


\end{document}